\documentclass[a4paper,11pt]{article}
\pdfoutput=1

\usepackage{latexsym}
\usepackage{graphicx}
\usepackage{mathptmx}
\usepackage{caption}
\usepackage{subcaption}
\usepackage{amsmath}
\usepackage{amsfonts}
\usepackage{amssymb}
\usepackage{amsbsy}
\usepackage{amsthm, authblk, makecell}

\usepackage[pdftex,colorlinks=true,urlcolor=blue,citecolor=black,anchorcolor=black,linkcolor=black]{hyperref}

\usepackage{geometry}
\begin{document}
\title{A HIGH-FIDELITY, MACHINE-LEARNING ENHANCED QUEUEING NETWORK SIMULATION MODEL FOR HOSPITAL ULTRASOUND OPERATIONS}

\author[2]{Yihan Pan*}
\author[1]{Zhenghang Xu\footnote{Co-first authors}}
\author[1]{Jin Guang}
\author[8]{Jingjing Sun}
\author[7]{Chengwenjian Wang}
\author[7]{Xuanming Zhang}
\author[1]{Xinyun Chen}
\author[1,3]{J.G. Dai}
\author[5]{Yichuan Ding}
\author[4]{Pengyi Shi}
\author[6]{Hongxin Pan}
\author[6]{Kai Yang}
\author[6]{Song Wu}

\affil[1]{School of Data Science, Shenzhen Research Institute of Big Data, The Chinese University of Hong Kong, Shenzhen, Guangdong 518172, PRC}
\affil[2]{School of Management and Economics, The Chinese University of Hong Kong, Shenzhen, Guangdong 518172, PRC}
\affil[3]{School of Operations Research and Information Engineering, Cornell University, Ithaca, NY, 14853, USA}
\affil[4]{Krannert School of Management, Purdue University, West Lafayette, Indiana 47907, USA}
\affil[5]{Desautels Faculty of Management, McGill University, Montreal, QC H3A 1G5, CAN}
\affil[6]{Luohu Hospital System, Shenzhen, Guangdong 518172, PRC}
\affil[7]{School of Mathematical Sciences, Fudan University, Shanghai 200433, PRC}
\affil[8]{School of Information Management and Engineering, Shanghai University of Finance and Economics, Shanghai 200433, PRC}

\maketitle

\section*{ABSTRACT}
We collaborate with a large teaching hospital in Shenzhen, China and build a high-fidelity simulation model for its ultrasound center to predict key performance metrics, including the distributions of queue length, waiting time and sojourn time, with high accuracy. The key challenge to build an accurate simulation model is to understanding the complicated patient routing at the ultrasound center. To address the issue, we propose a novel two-level routing component to the queueing network model. We apply machine learning tools to calibrate the key components of the queueing model from data with enhanced accuracy.
\section{INTRODUCTION}


In this research, we collaborate with a large teaching hospital in Shenzhen, China, \textit{Luohu Hospital}, to develop a high-fidelity simulation model
to predict key performance metrics in its ultrasound center. Founded in April 1957, Luohu Hospital is a modern medical treatment center that integrates preventive health care, rehabilitation, research, and teaching. 
Like many hospitals in Asia, the hospital owns a large campus with several buildings that provide primary care, medical examinations, emergency care, and inpatient care. This mode is similar to the integrated practice unit (IPU) adopted by several major US hospital systems, including Mayo Clinic and Cleveland Clinic \cite{swensen2009quality,mueller2009incorporating,knoer2013highlights,patrnchak2016implementing}. In an IPU, most patients' outpatient \textit{itinerary}, including the primary visit and examinations, can be finished in one day given that primary doctors and examination services are co-located in the same place.

Among the examination services that Luohu Hospital provides, the ultrasound examination center is of critical importance (referred to as the \textit{ultrasound center} in the rest of this paper). The demand for its ultrasound center has been rapidly growing in the past few years since the strategic focus of the hospital on strengthening its obstetrics and gynecology (OB/GYN) specialty. Most of OB/GYN patients need an ultrasound exam after the initial consult with the primary doctor. The ultrasound exam can typically be scheduled and finished within the same day. After getting the exam results, patients return to the primary doctor for the second consult and get a final diagnosis, concluding their itinerary (a small proportion of them may need a second set of exams before the final diagnosis can be achieved). 
Preliminary analysis shows that the time spent to finish the ultrasound exam is the most time-consuming part for a patient's itinerary. 
Thus, it is important to understand the operations of the ultrasound center and develop accurate performance prediction on key metrics including the waiting time. Such performance prediction would then provide the basis to identify operational strategies that could reduce the total amount of time these patients need to spend in the hospital for their outpatient itinerary, improving the patient experience.

In this paper, we develop a high-fidelity simulation model that integrates predictive tools and queueing models to capture the operations in the ultrasound center. The model is calibrated with detailed patient-level data and can provide accurate prediction on the key performance metrics, such as distributions of \textit{time-dependent} queue length, waiting time, and sojourn time. The purpose of this research is multiple-fold: 
\begin{itemize}
    \item To empirically analyze the operations and key performance metrics for the ultrasound center in Luohu Hospital;
    \item To develop a high-fidelity simulation model to accurately capture key performance metrics;
    \item To provide a platform for hospital managers to understand the current performance and identify operational strategies that can improve the system performance.
\end{itemize}
In the following two subsections, we first describe the challenges we met in building the simulation model. Then, we highlight the contribution we made in incorporating a few salient features in the simulation model that are crucial for accurate prediction. 



\subsection{Challenges}
\label{subsec: challenge}

We model the ultrasound center as a multi-class, multi-pool parallel-server queue. Each class corresponds to one type of patients, and each pool of servers corresponds to one or a few examination rooms. However, different from the standard multi-class, multi-pool queue, operations in the ultrasound center have several unique features, including \textit{patient and server heterogeneity, complicated patient routing}, and \textit{time-dependency} on the server side. We specify each as follows.

First, there is great heterogeneity in the patients and exam rooms (servers). The center serves patients from more than 20 medical specialties, who come with very distinctive test items, often more than one item. This results in a vast combination of service requests and high variances in the service times. Meanwhile, the servers are highly different. Most rooms are capable of performing common tests, but some tests have to be done in rooms with certain equipment. The majority of technicians are known as ``general practitioners,'' who are qualified to perform most of the common tests. Several technicians, known as the ``specialists,'' are specialized to handle certain specific test items. Furthermore, different technicians vary in their skill levels and effectiveness, presenting a large difference in their service speed.

\begin{figure}[htp]
    \centering
    \begin{subfigure}[b]{0.33\textwidth}
        \includegraphics[width=\linewidth]{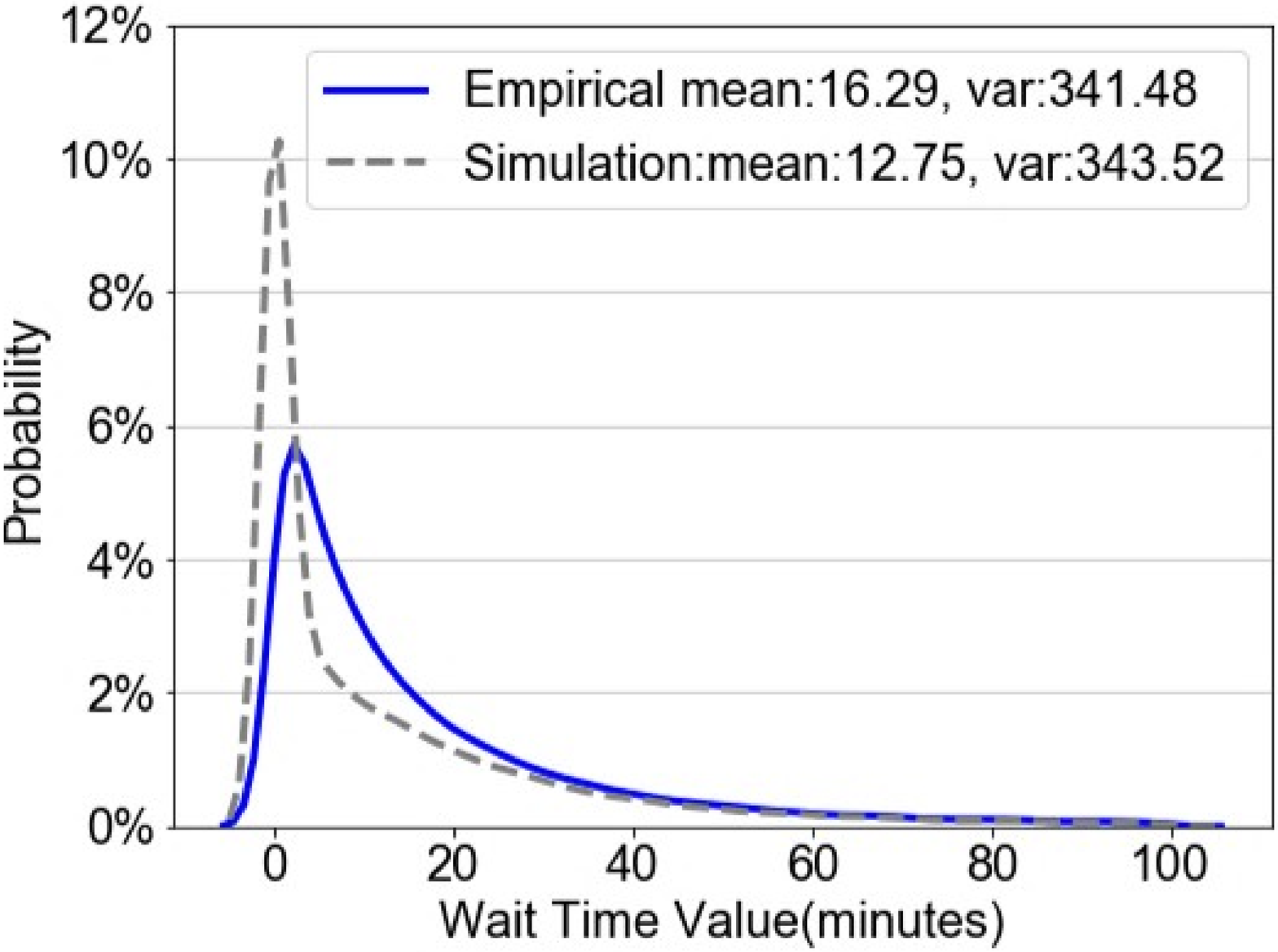}
        \caption{Wait Time Distribution}
        \label{fig: wait_time_model0}
    \end{subfigure}%
    \begin{subfigure}[b]{0.33\textwidth}
        \includegraphics[width=\linewidth]{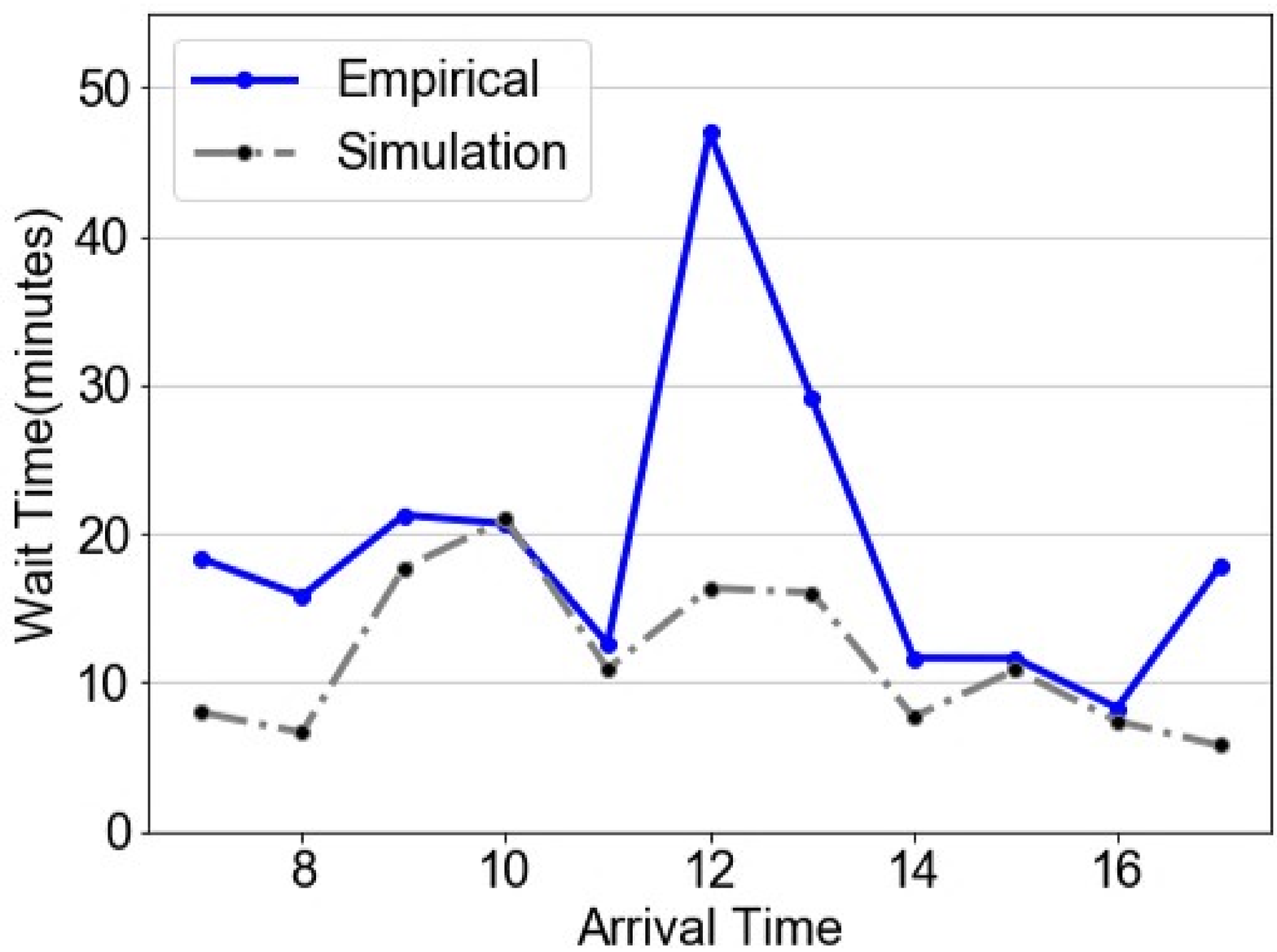}
        \caption{Average Wait Time }
        \label{fig: mean_wait_model0}
    \end{subfigure}
    \begin{subfigure}[b]{0.33\textwidth}
        \includegraphics[width=\linewidth]{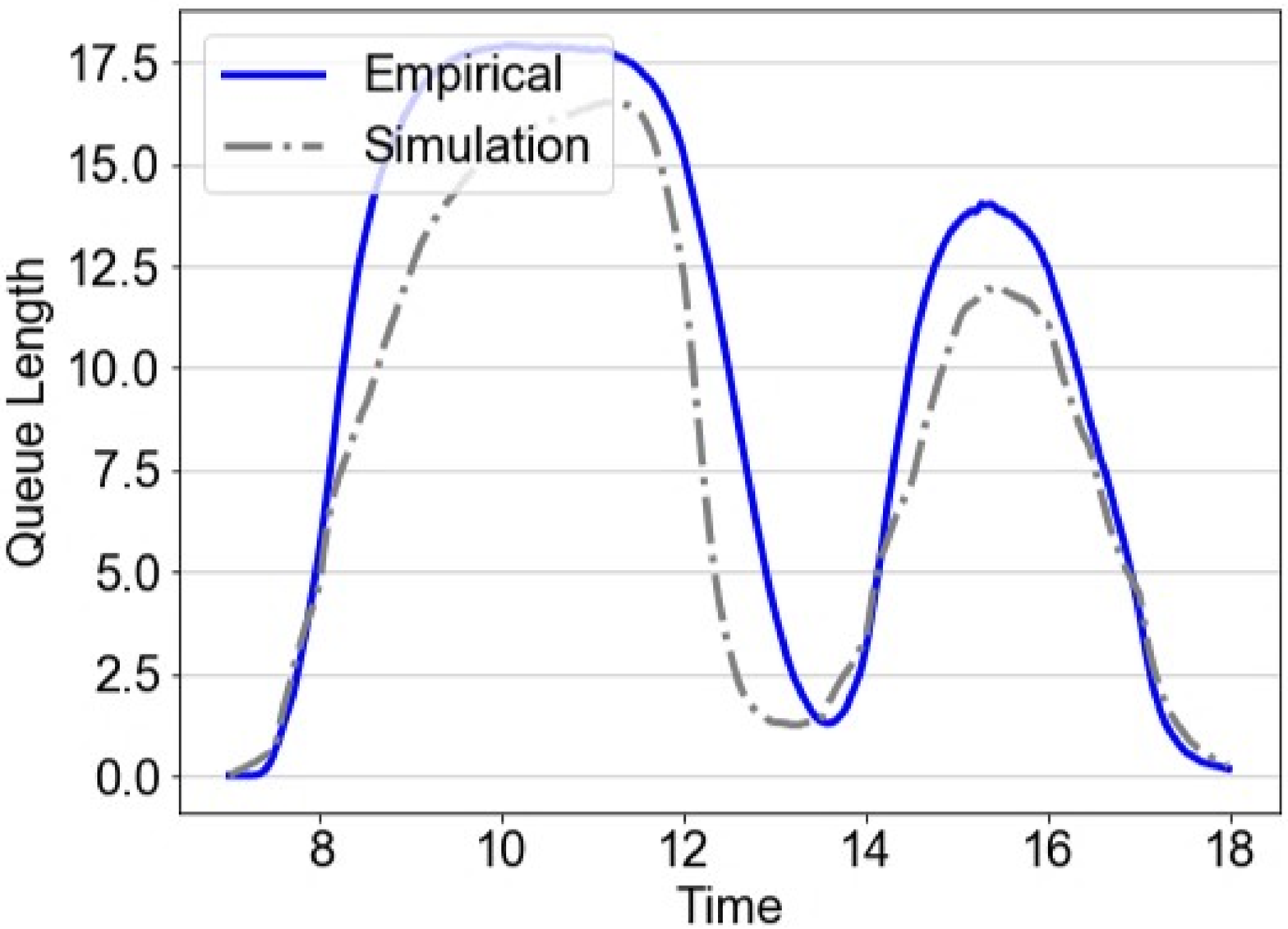}
        \caption{Average Queue Length }
        \label{fig: ql_model0}
    \end{subfigure}
    \caption{The performance of conventional multi-class model with join-shortest-queue (JSQ) routing.}
    \label{fig:JSQ_performance}
\end{figure}

Second, the routing from different patients to different rooms is highly complex. Modeling routing properly turns out to be one of the most important, yet challenging, steps for us to calibrate the simulation model with empirical performance. One source of the challenges comes from the great heterogeneity in patients and servers. Additionally, (i) there is no specific rule for the routing, and nurses who are in charge of the routing use their discretion to assign patients to different rooms; (ii) rooms are not necessarily open for the entire day due to different staffing level and technicians/doctors taking breaks occasionally (resulting in room being unavailable). Indeed, existing stylized routing policies such as first-come-first-serve (FCFS) or join-shortest-queue (JSQ) all fail to produce accurate prediction. Figure~\ref{fig:JSQ_performance} compares the empirical performance and the simulation output from a conventional multi-class queue and JSQ routing. The simulation output fails to capture the empirical performance, in particular, the waiting time distribution is far off from the empirical one though the average is reasonably close. 

Lastly, as in many healthcare systems, there is a strong time-dependency in the arrival rate. Our descriptive analysis also reveals a strong time-dependency in the service speed and server unavailability. We find that the service rate tends to be faster in the late morning than early morning, presenting additional heterogeneity in service speed even within the same server. There are two sources for server unavailability. First, not all rooms are open during the entire day. More rooms are open during the peak time (mid-morning) than in the afternoon. Second, some technicians may take a random ``break'' after serving a patient, resulting in random server unavailability that further complicates the simulation.

In summary, the unique features in the ultrasound center operations 
present the following challenges to construct a high-fidelity simulation model that can accurately capture empirical performance:
\begin{itemize}

\item The main challenge is to ``learn'' a proper routing model from the actual practice for the simulation model. 

\item To build a proper routing model, we further need:
(a) patient and server classifications, and (b) estimating server unavailability.
\end{itemize}

For (a), we need to strike a balance between having sufficient classes to capture the heterogeneity (in patients and servers) and avoiding too many classes that can lead to estimation inaccuracy. For (b), if we do not properly account for server unavailability, direct estimation of the routing policy from data may end up with counter-intuitive results. For example, without modeling room unavailability, direct estimation shows that more patients are assigned to rooms with a long queue. However, this result is biased because a room that is unavailable for the next half-hour will have a much longer queue.

\subsection{Previous Works}


There is a rich literature in adopting discrete event simulation to investigate the effectiveness of service policies in hospitals \cite{martin2020integrated,feng2020simulation,swan2019evaluating,zhang2018application}.  In these studies, researchers usually build simplified models to extract and analyze the key factors causing congestion. 
Many works in this area were based on flexible simulation models \cite{jin2013simulation,lin2013decision,rohleder2011using,guo2004outpatient,harper2003reduced,liang2015improvement,pan2015patient}. For example, \cite{guo2004outpatient} found that large variations in patients' service type and service duration could greatly affect the simulation performance. \cite{harper2003reduced} modeled the detailed dependency between oncologists and chemotherapy appointments to optimize the efficiency of patient flow. \cite{lin2013decision} presented an analysis on an eye clinic located in Singapore, carefully considering patient behavior in the appointment system (e.g., unpunctuality). These works dealt with a certain class of patients and an appointment mode, which are not in line with our circumstances, where there are high variations in both patient (customer) and doctor (server) sides.

From an operations perspective, the ultrasound center is a queueing network of multi-class customers and servers. Among the numerous papers on simulation models in healthcare settings, our work is most relevant to those studying emergency department (ED) and inpatient department for hospitals in North America or others \cite{wang2011improving,medeiros2008improving,swan2019evaluating,chavis2016simulation}. \cite{wang2011improving} developed an open queueing network with blocking to capture the situation when a hospital is full and patients have to be sent to another hospital. \cite{medeiros2008improving} considered an ED with a short-stay unit that provides an observation place to determine whether a patient will be sent to an inpatient unit or not. 
Comparing to these models, our ultrasound center presents several unique features as discussed above, which present us from off-the-shelf simulation models. 

Directly predicting waiting times with machine learning tools is an alternative to our model-based approach. For example, \cite{hermanto2018waiting,limlawan2020development} apply artificial neural networks to predict waiting time using patient-level features. Our approach is different from this line of research, as it is a ``white box" model and captures the underlying queueing dynamics. Capturing the queueing dynamics is important for counterfactual study, where one needs to evaluate the performance changes under different operational strategies (e.g., capacity changes). Black-box machine learning models often lack the ability to capture the complex dependence between performance metrics and system parameters.

\subsection{Main Contribution}

\begin{figure}[htp]
    \centering
    \begin{subfigure}[b]{0.33\textwidth}
        \centering
        \includegraphics[width=\linewidth]{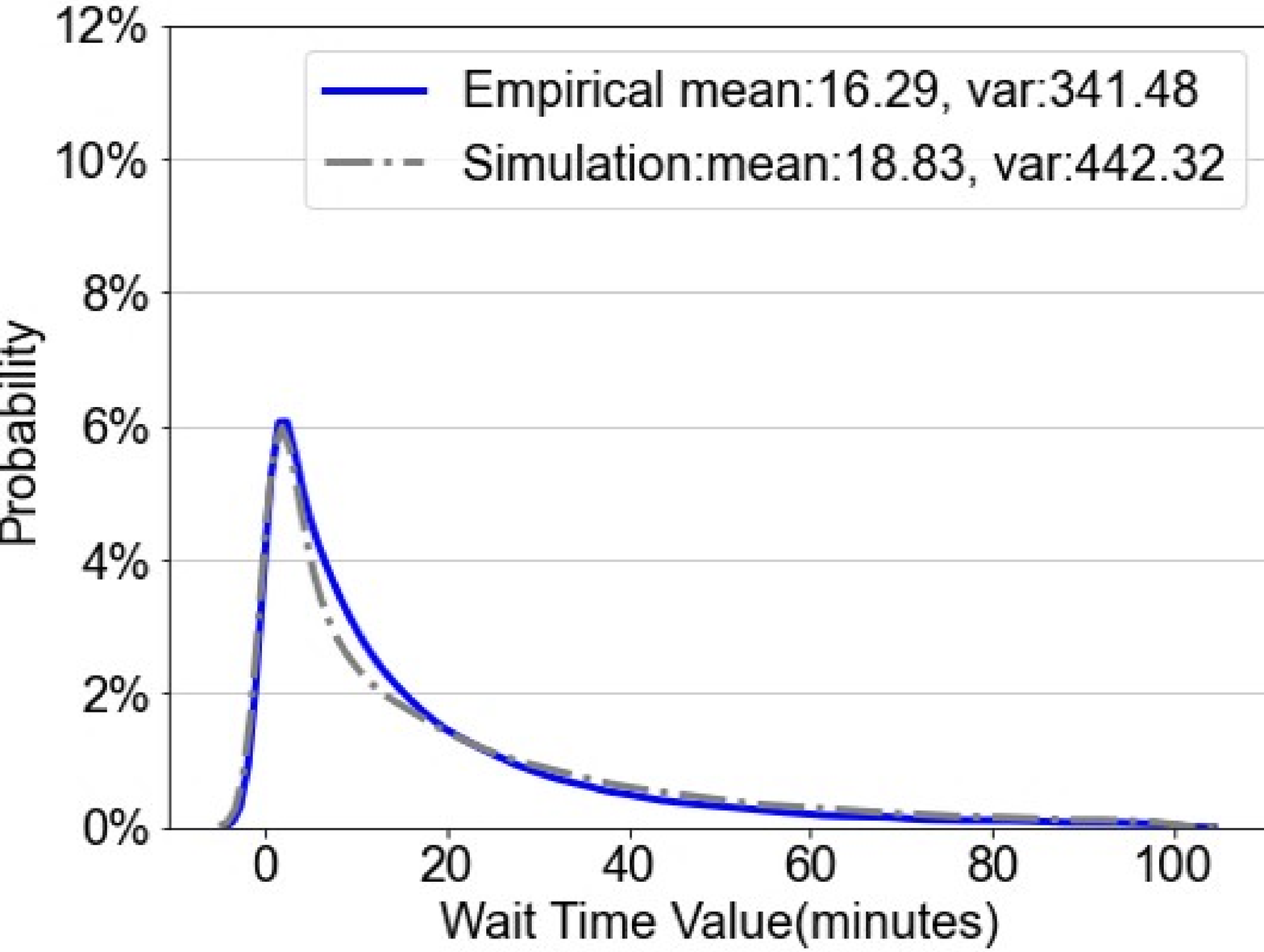}
        \caption{Wait Time Distribution}
        \label{fig: wait_time_model1}
    \end{subfigure}%
    \begin{subfigure}[b]{0.33\textwidth}
        \centering
        \includegraphics[width=\linewidth]{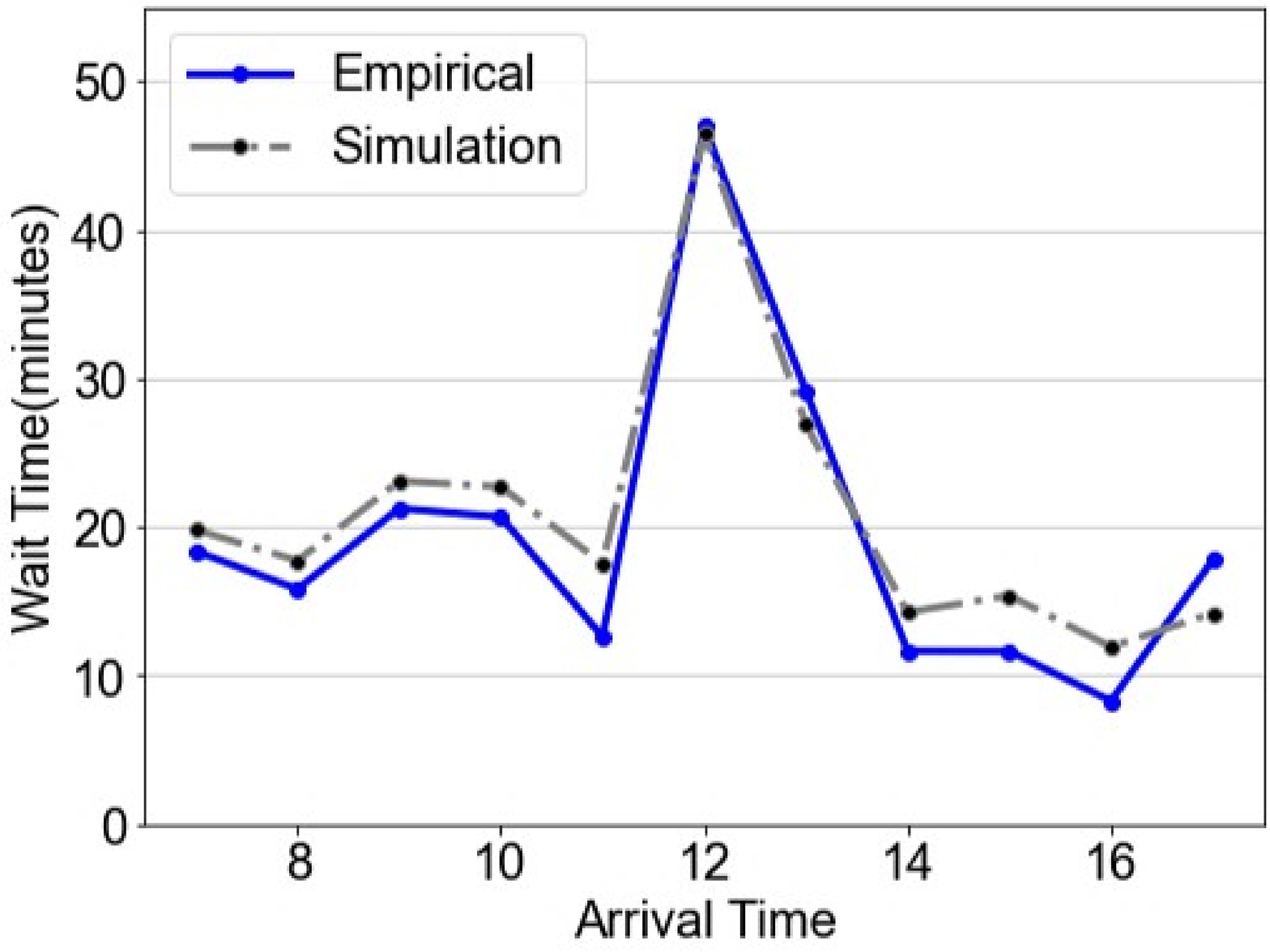}
        \caption{Average Wait Time}
        \label{fig: mean_wait_model1}
    \end{subfigure}
    \begin{subfigure}[b]{0.33\textwidth}
        \centering
        \includegraphics[width=\linewidth]{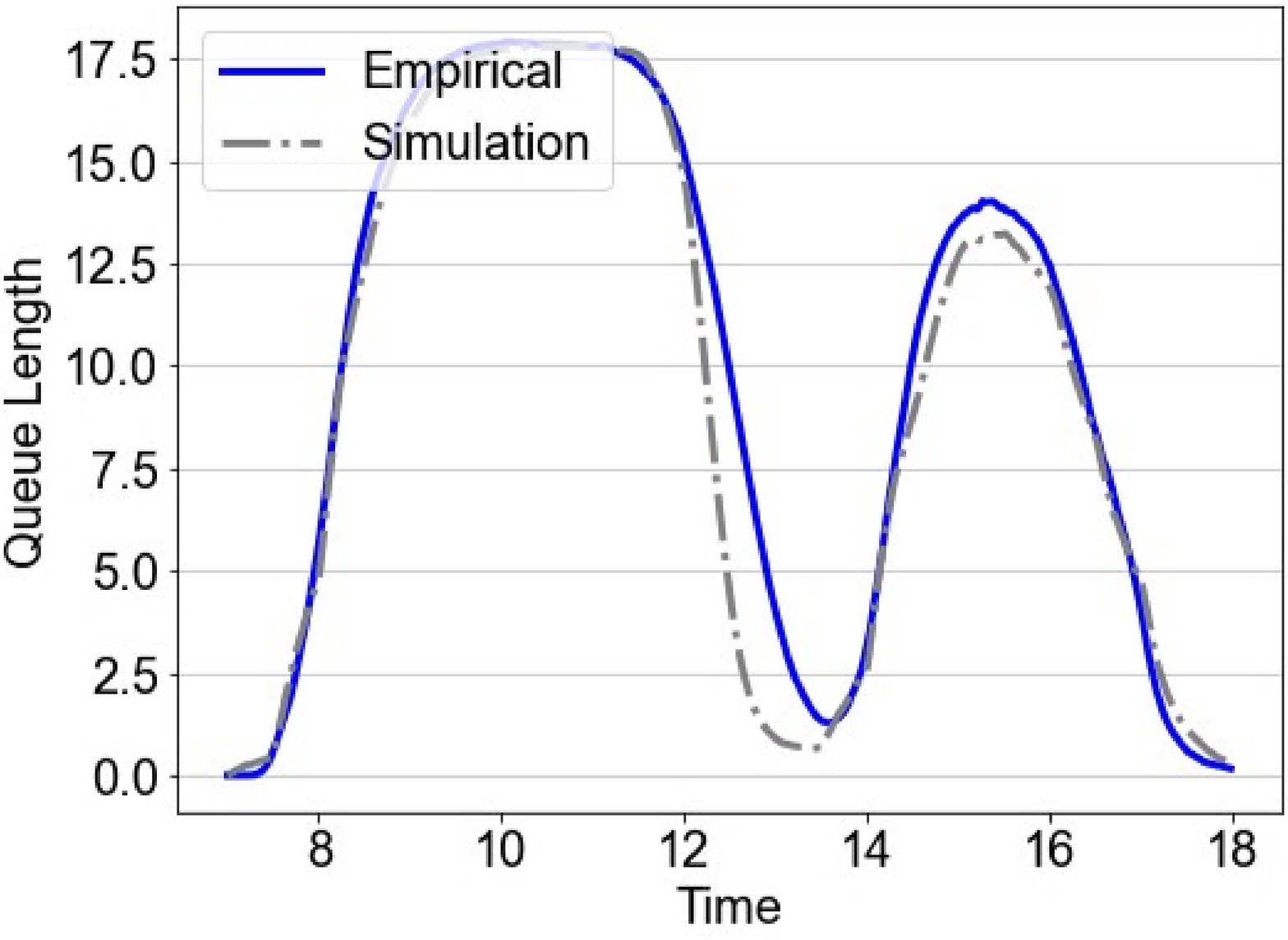}
        \caption{Average Queue Length}
        \label{fig: ql_model1}
    \end{subfigure}
    \caption{The performance of our high-fidelity simulation model. The horizontal axis of (b) (c) ranges from 7 to 17, which points to the hourly index in one day. For the other figures in our paper, the time axis has the same meaning as we show here.}
    \label{fig:model_good_waiting}
\end{figure}

We develop a high-fidelity simulation model that integrates machine learning tools and queueing models in a novel manner, instead of using either alone. We leverage machine learning as predictive tools to calibrate key modeling components in our stochastic queueing network. This integrated model produces accurate performance prediction enhanced by machine learning, while the queueing-based model allows us to retain desirable properties such as of being structural and interpretable. Figure~\ref{fig:model_good_waiting} demonstrates the output from our high-fidelity simulation model, which can accurately capture various key performance metrics, including the waiting time distribution that was far off from the conventional model as shown in Figure~\ref{fig:JSQ_performance}. 
More validation results are in Section~\ref{sec:model-valid}. 

We highlight our contribution to the literature as follows:
\begin{itemize} 
    \item \textbf{Two-level patient routing component. } To address the main challenge of building a proper routing component, we combine queueing structures with data to analyze routing behaviors in practice, rather than imposing stylized routing policies. Specifically, we develop a two-level hierarchical routing policy. In the first level, we decided which group of exam rooms to route a patient. Then, in the second level, we decide which room, within the selected group from the first level, to route the patient. At each level, we apply machine learning methods to learn the routing rule from data. This two-level structure allows us to separate the key determinants in the routing decisions: patient's examination items (medical factor) in the first level, and the queue length and the room availability (operational factor) in the second level. 

    \item \textbf{Clustering-based patient and room classification. } We leverage clustering algorithms to classify patients and servers. For patients, the clustering is based on similarity of the service time distributions. For servers, the clustering is based on the empirical routing probability and service time distributions. In this way, we group rooms that serve similar types and proportions of patients from different specialties. These clustering algorithms provide us the flexibility to choose a proper number of classes and the desired proximity on key features for calibrating the model (e.g., service time).  

    \item \textbf{Estimating server availability. }
    As discussed, the server availability is affected by (i) the staffing (room open/close) policy and (ii) technician taking a random break after serving the last patient. For (i), we calibrate a time-varying room open policy from data. For (ii), we analyze the sequence of patients, who received exams in the same room, on their arrival, service start, and service end timestamps. We identify two types of gap times. The first type is the break time, defined as the duration between the service-end time of the previous patient and the service-start time of the next patient in a busy period. The second type is on the patient side -- walking time, defined as the duration between the arrival time and service-start time of the next patient after the previous patient has finished service. It turns out incorporating these two gap times in the simulation can further improve the prediction accuracy. See more details in Section~\ref{subsec: room open}.

\end{itemize}

\textbf{Relevance of our simulation model in decision making.} 
The simulation model we build in this research provides an accurate prediction on key performance metrics, 
which can be used to provide patients with an estimated waiting time on finishing the ultrasound exam. 
Such ``delay announcement'' is often used in call center operations and has been shown to greatly improve customer experience. Moreover, this simulation model provides a high-fidelity platform to evaluate the impact of different operational changes. 
For example, in \cite{chen2021saa}, the authors develop a patient scheduling policy that explicitly accounts for patient revisits in a day. 
The solution algorithm relies on the estimation of the sojourn time distribution at the examination center (ultrasound for OB/GYN patients) as a key input to search for the optimal schedule. The empirical estimates of the sojourn time can be possibly used as an input for the initial iteration of solution algorithm. However, in later iterations, the sojourn time distribution needs to be re-evaluated when all primary doctors' offices use the new schedule. This will cause a major shift to the arrival pattern to the ultrasound center and hence, change the sojourn time. In this case, one cannot use empirical estimates anymore. Having a high-fidelity simulation model that allows evaluation of the sojourn time distribution under the new arrival pattern is critical. Compared with pure black-box machine learning tools, our queueing-based model can better capture the complicated dependence between arrival pattern and system congestion.

\textbf{Paper outline.} In the rest of the paper, we first give an overview of the ultrasound center and present descriptive statistics in Section \ref{sec:descrip}. We also describe the classification of patients and rooms in Section \ref{sec:descrip}. Then we specify the details of our simulation model in Section \ref{sec:sim-model}. We present the simulation results that validate our model in Section \ref{sec:model-valid}. We conclude the paper in Section \ref{sec:conclud}.

\section{DATA AND DESCRIPTIVE ANALYSIS}
\label{sec:descrip}

Patients who request tests from the ultrasound center come from various departments (medical specialties) and their request examination items vary greatly. There are 189 different ultrasound examination items out of the 175702 total records in our 1-year data in 2018. The top 10 examination items account for over 75\% of the tests with more than 2000 occurrences in a year. Table~\ref{tab: top 10 exam item} lists the top 10 examination items. Appendix~\ref{app: data merging} details our process of merging different data sources and the key attributes/timestamps in the data. 

\begin{table}[htp]
	    \centering
\scalebox{0.8}{
	    \begin{tabular}{|c|c|c|}
	        \hline
	        Index & Inspection Item & Propotion \\
	        \hline
	        A & Transabdominal and transvaginal color Doppler ultrasound & 32.3\% \\
	        \hline
	        B & Color Doppler ultrasound examination of uterine attachment & 8.79\% \\
	        \hline
	        C & Fetal Level 2 Examination & 7.07\% \\
	        \hline
	        D & Superficial color Doppler ultrasound: breast and axillary lymph nodes & 6.32\% \\
	        \hline
	        E & Color Doppler ultrasound: liver, gallbladder, spleen, pancreas and portal vein system & 5.26\% \\
	        \hline
	        F & Intracavitary three-dimensional color ultrasound imaging & 3.96\% \\
            \hline
            G & Four-dimensional Level 2 fetal examination & 3.39\% \\
            \hline
            H & NT & 3.25\% \\
            \hline
            I & Superficial color Doppler ultrasound: thyroid and lymph nodes & 3.20\% \\
            \hline
            J & Cardiac Color Doppler Examination Package 2 & 3.02\% \\
            \hline
	    \end{tabular}
}	    
\caption{Top 10 Ultrasound Examination Items.}
\label{tab: top 10 exam item}
\end{table}

\begin{figure}[htp]
    \centering
    \begin{subfigure}[b]{0.45\textwidth}
    \centering
        \includegraphics[width=5cm, height=3.5cm]{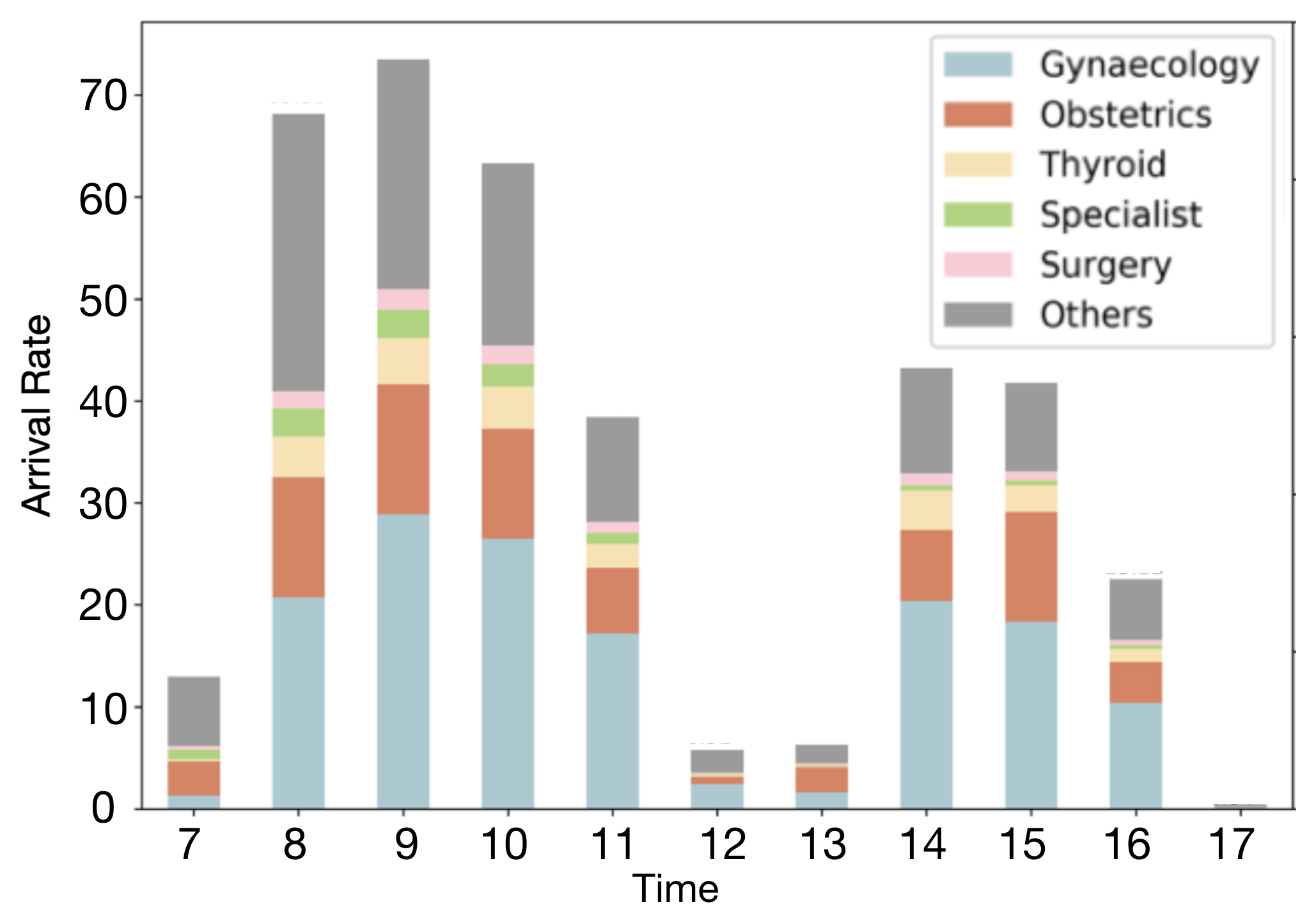}
        \caption{Arrival Rate by Department}
        \label{fig: arr department}
    \end{subfigure}%
    \begin{subfigure}[b]{0.45\textwidth}
    \centering
        \includegraphics[width=5cm, height=3.5cm]{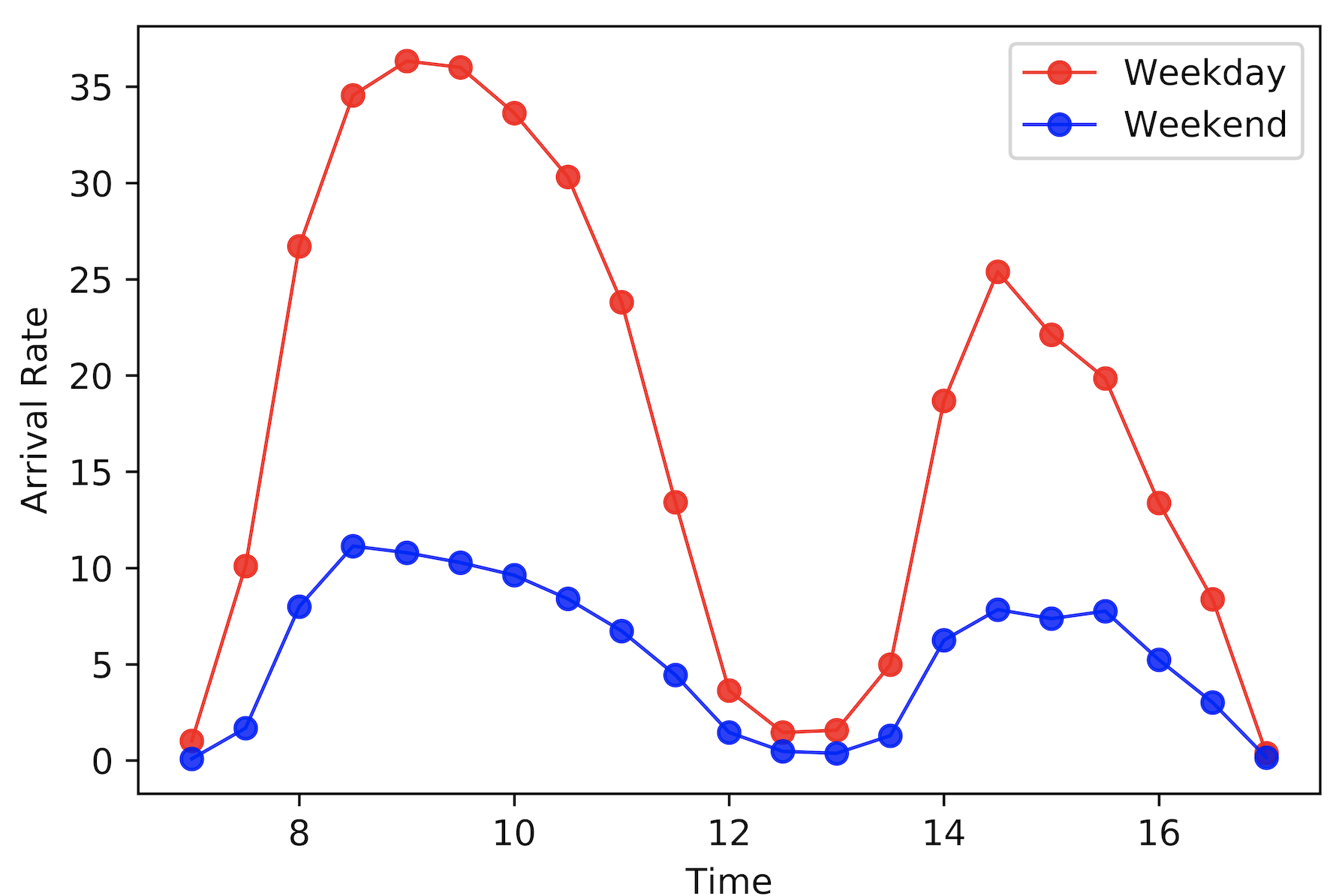}
        \caption{Arrival Rate on Half Hour by Weekday/Weekend}
        \label{fig: arr weekend}
    \end{subfigure}
    \caption{Arrival Rate for Ultrasound Center}
    \label{fig: arrival-rate}
\end{figure}

Figure~\ref{fig: arrival-rate} plots the hourly arrival rates to the ultrasound center, which shows a time-dependent pattern. Figure~\ref{fig: arr department} decomposes the hourly arrival rates into the corresponding rate from each department such as OB/GYN and Surgery. The peak arrival occurs in the morning (8-11 AM) each day. In addition to the time-of-day pattern, the arrivals also demonstrate a day-of-week pattern. The two curves in Figure~\ref{fig: arr weekend} show the hourly arrival rates averaged over all weekdays and weekends, respectively. Next, we discuss how to classify the patients and the servers (examination rooms). 


\subsection{Classification of Examination Item and Room Types}
\label{subsec: classify item types}

\begin{figure}[htbp]
    \centering
        \begin{subfigure}[b]{0.3\textwidth}
    \centering
        \includegraphics[width=4.5cm, height=3.5cm]{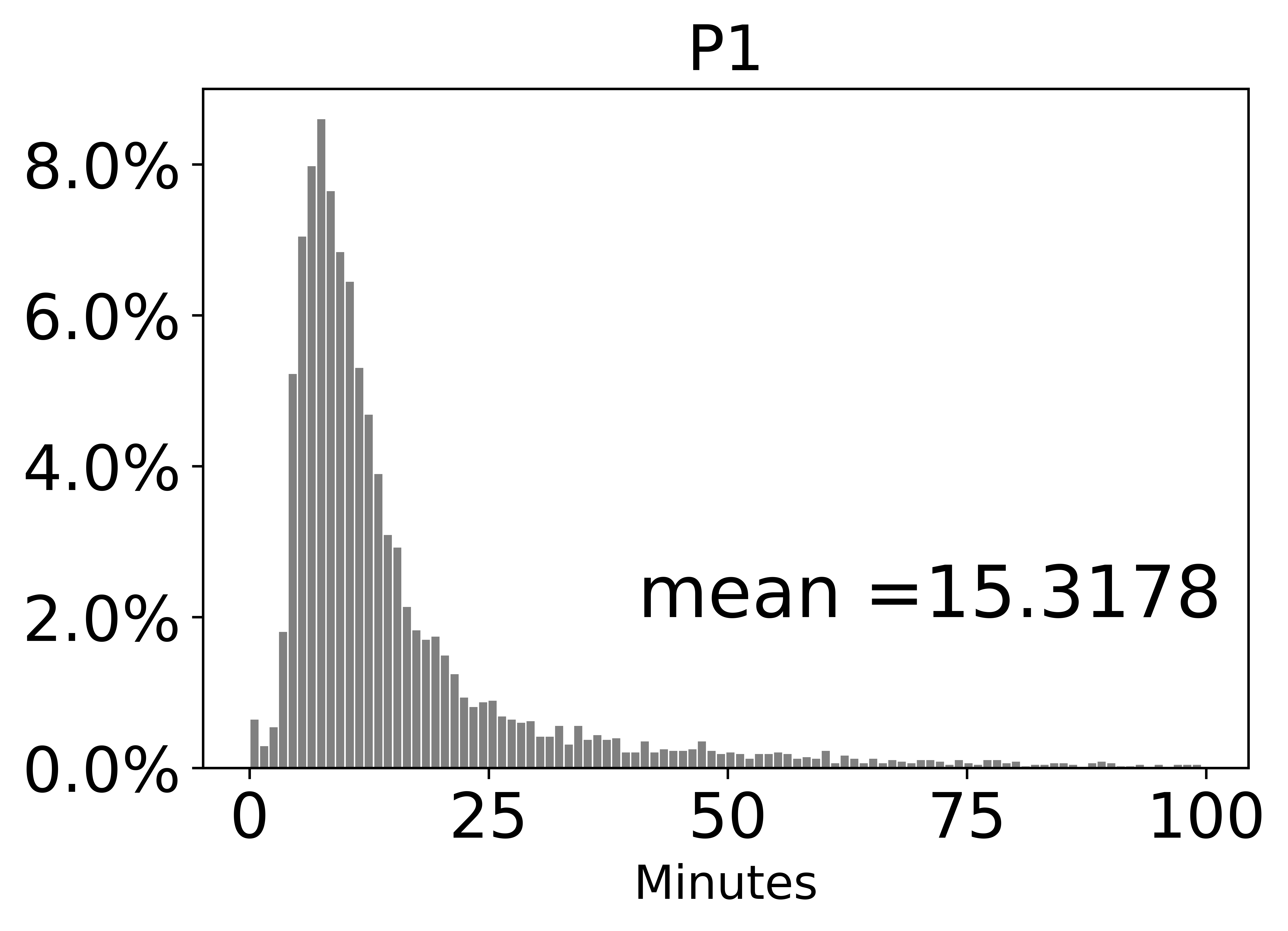}
        \caption{Service time distribution for group 1 (P1) exam type}
        \label{fig: ser-time-P1}
    \end{subfigure}%
    ~~
    \begin{subfigure}[b]{0.3\textwidth}
    \centering
        \includegraphics[width=4.5cm, height=3.5cm]{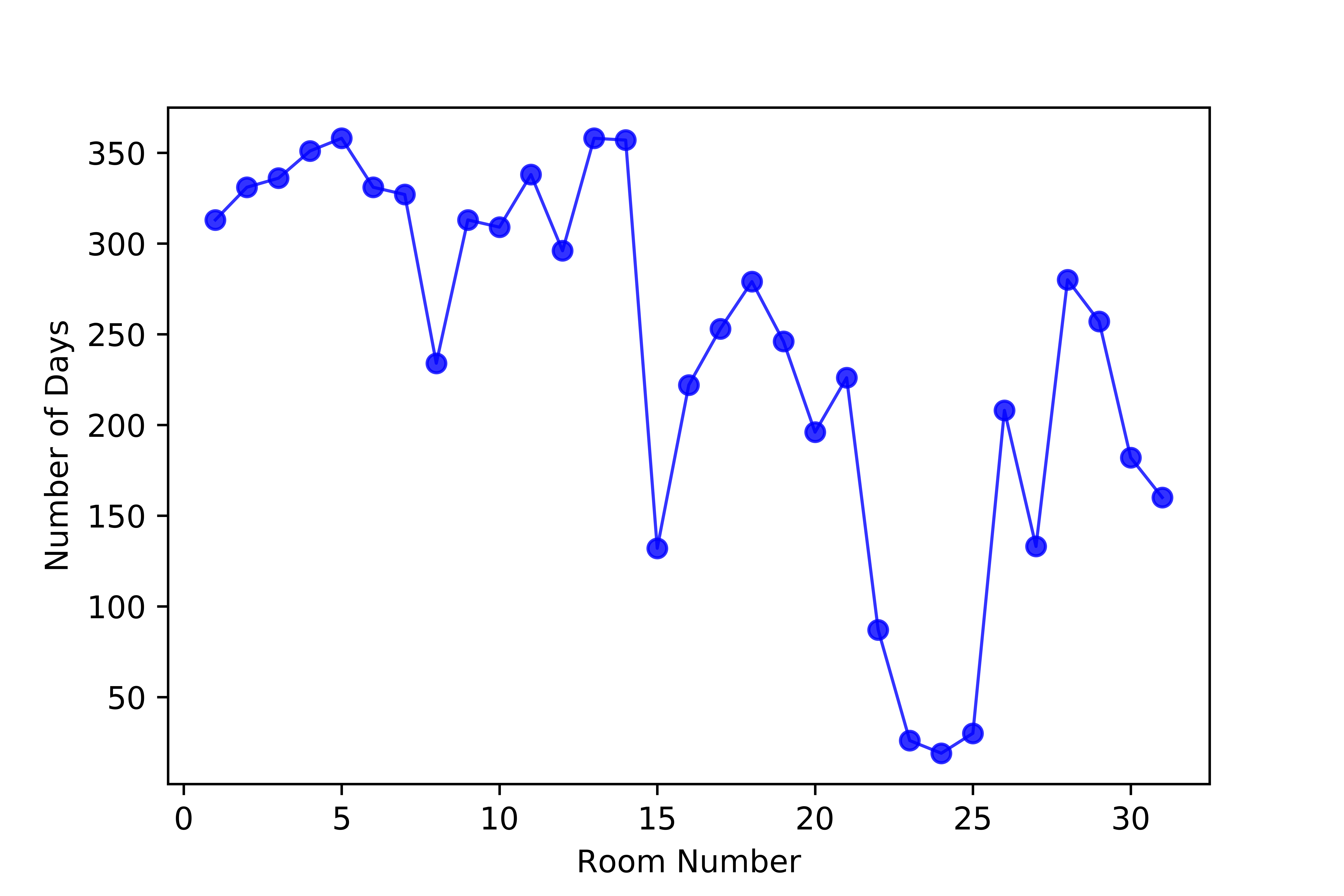}
        \caption{Number of opening days in a year}
        \label{fig: room open}
    \end{subfigure}%
    ~~ 
    \begin{subfigure}[b]{0.3\textwidth}
    \centering
        \includegraphics[width=4.5cm, height=3.5cm]{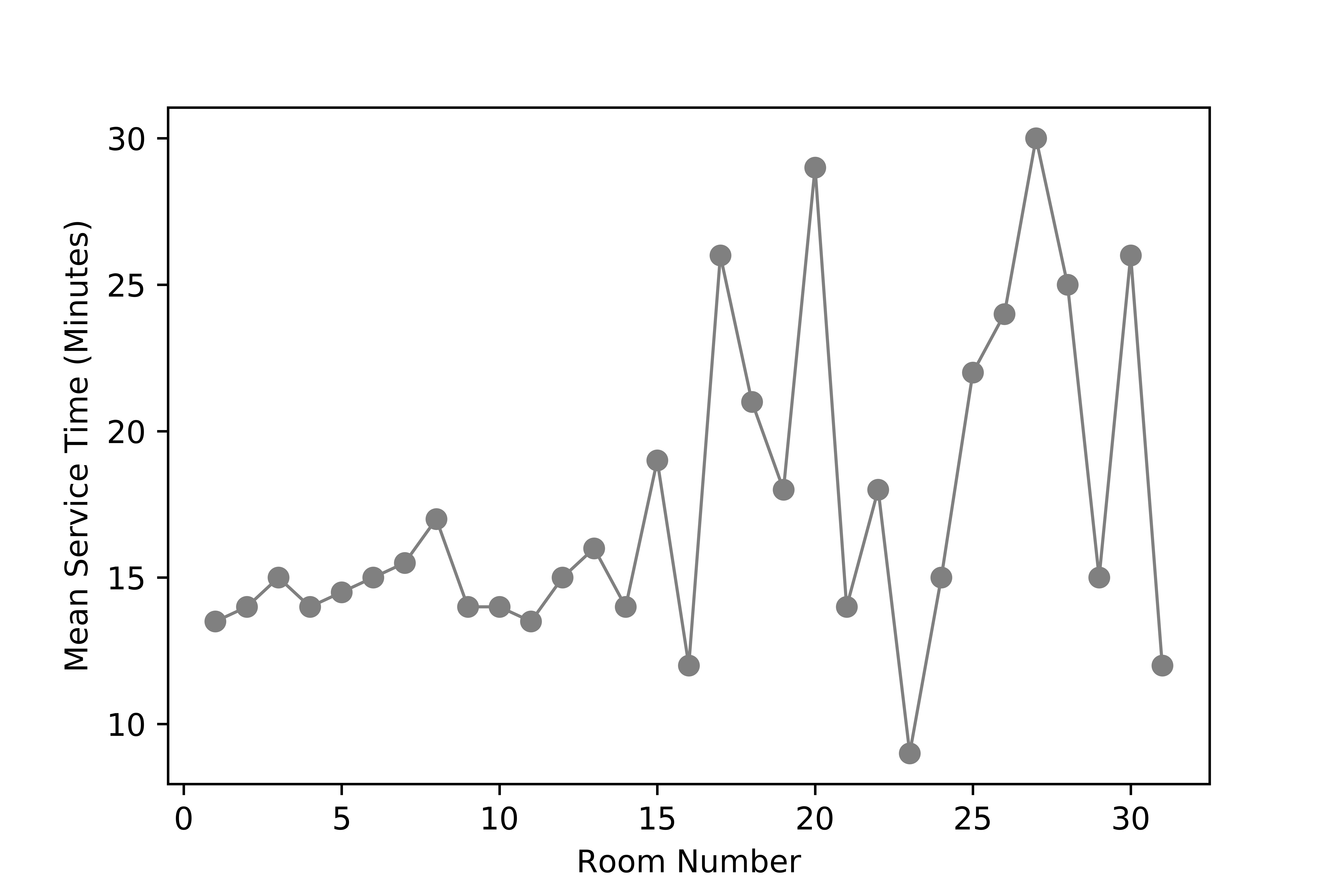}
        \caption{Mean service time for each room}
        \label{fig: room service}
    \end{subfigure}
    \caption{Exam Item and Room Features for Ultrasound Center}
    \label{fig: room feature}
\end{figure}

To classify patients, we note that, in addition to the large number of examination items, a significant number of patients come to the ultrasound center with more than two examination items. Thus, if we were to classify patients according to their required examination items, it would lead to too many classes and cause estimation errors due to the small sample size in each combination. To avoid this issue, we leverage clustering algorithms to classify patients, based on the mean and standard deviation of their service time and the frequency of going to different rooms. We further separate the 17.48\% of patients who need multiple examination items into a different category.  
Appendix~\ref{app:class-exam-type} details the total six groups of examination items classified from the clustering algorithm (labeled as P1 through P6 in the rest of the paper).  Figure~\ref{fig: ser-time-P1} plots the empirical service time distribution for the first group (P1). The empirical service time distributions for other groups are provided in Appendix~\ref{app:class-exam-type}.  
 

For the server side, the ultrasound center at Luohu Hospital has 32 rooms that are equipped with different kinds of exam machines. According to our healthcare partner, most examination items can be performed in the majority of rooms, but a few examination items could only be done in certain specific rooms. Additionally, not every room is open during the entire working hour in a day. Figure~\ref{fig: room open} shows the number of open days, out of 365 days in a year, for each of the 32 rooms. Figure~\ref{fig: room service} shows the mean service time for each of the 32 rooms. We can see that both the number of open days and service times vary greatly among different rooms. 

\begin{figure}[htp]
	\centering
	\includegraphics[width=0.45\linewidth]{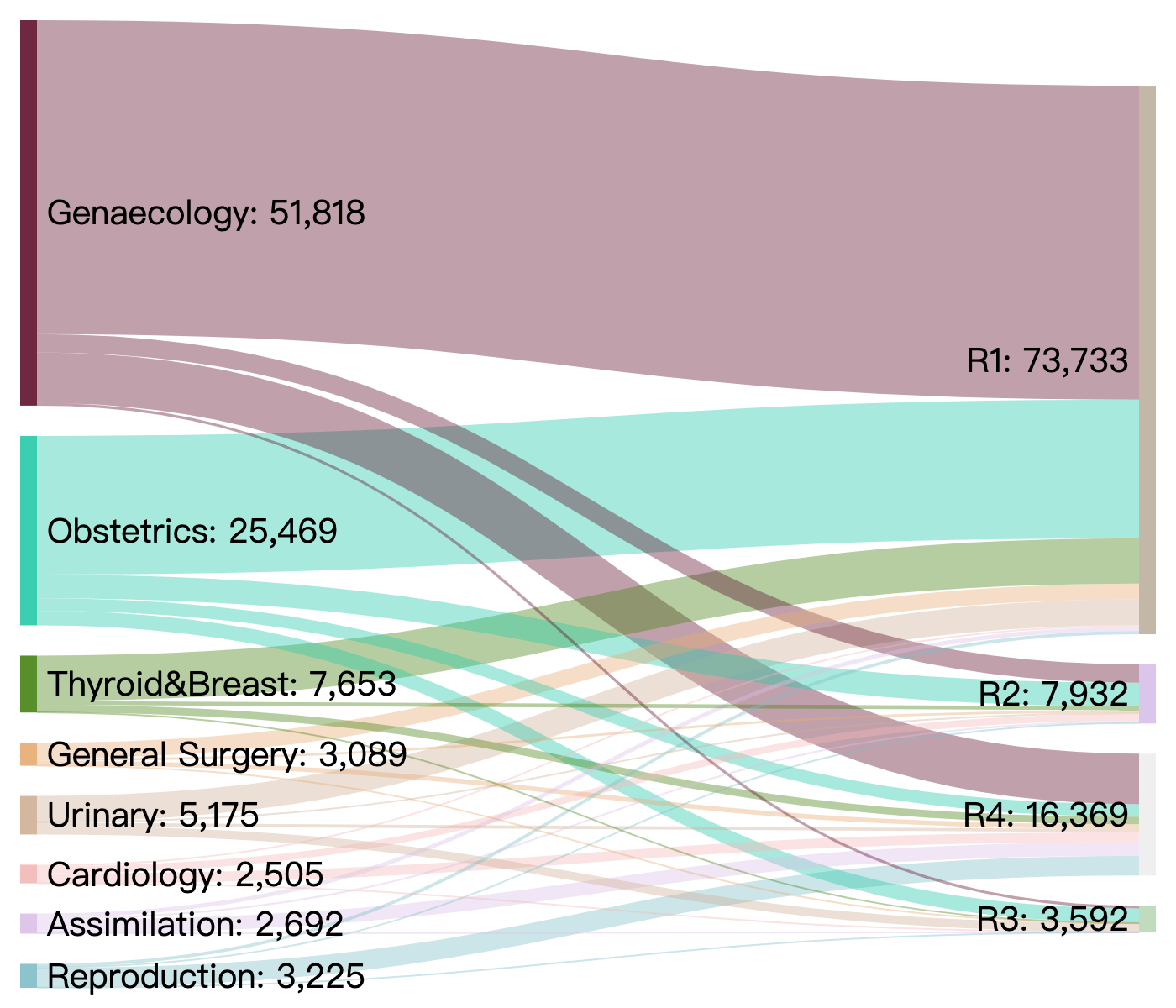}
	\caption{Patient Routing Process}
	\label{fig: ultra room routing}
\end{figure}

Similarly, we leverage clustering algorithms to classify the rooms into a handful of types. For this part, we use the Hierarchical Clustering Algorithm base on the following room features: (i) number of open days in one year, (ii) mean service time, which is related to service capacity, and (iii) proportion of examination items performed, which is related to routing. Table~\ref{tab: ultra room types} shows the four room types classified from the clustering algorithm, along with the room ID each type contains and the major types of examination items performed. Figure~\ref{fig: ultra room routing} plots the number of patient records from different departments in our 1-year data that are routed to each of the four room types, with the width of the line scaled proportionally to reflected the different volumes. 


        \begin{table}[htp]
	    \centering
	 \scalebox{0.85}{
	    \begin{tabular}{|c|c|c|}
	        \hline
	        Room Type & Corresponding Rooms Number & Patient Type \\
	        \hline
	        R1 & 1,2,3,4,5,6,7,8,10,12 & P2,P3,P5,P6 \\
	        \hline
	        R2 & 9,11,13,14,29 & P1,P2,P3,P5,P6\\
	        \hline
	        R3 & 17,18,20,26,27,28,30 & P2,P3,P4,P5,P6 \\
	        \hline
	        R4 & 15,16,19,21,22,23,24,25,31 & P2,P3,P5,P6 \\
	        \hline 
	    \end{tabular}
	    }
	    \caption{Classification of Patients by Examination Items.}
	    \label{tab: ultra room types}
	\end{table}

\section{Simulation Model}
\label{sec:sim-model}

\begin{figure}[htp]
	\centering
	\includegraphics[width=0.8\linewidth]{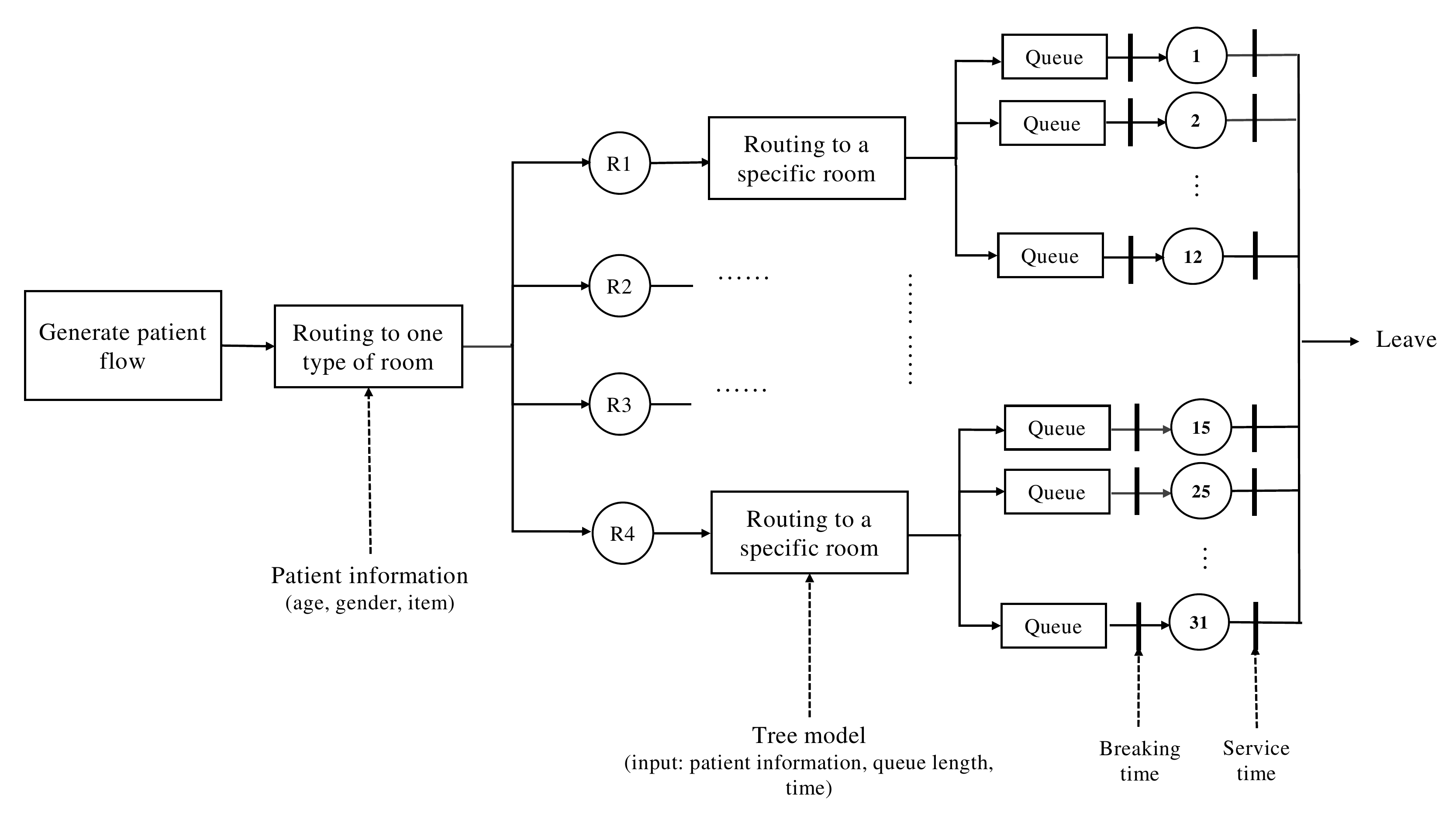}
	\caption{Flowchart of queueing simulation at ultrasound station.}
	\label{fig: patient routing}
\end{figure}

In this section, we introduce our high-fidelity simulation model to capture the patient flow and operations of the ultrasound center. This model is a multi-class, multi-pool parallel-server queueing system. The model contains four major components: (1) patients arriving, (2) checking room availability, (3) routing patients to the specific room, and (4) patients receiving service and leaving the system after service completion. Figure~\ref{fig: patient routing} shows the flowchart of the simulation process. The main novelties of our simulation model include developing a two-level, machine-learning-based routing policy in component (3), and adding random break time to the patient service process in (4). Thus, we will briefly describe components (1) and (2) first, and then specify components (3) and (4) in the following subsections. 

\subsection{Arrivals}
\label{subsec: ultra-queueing generation}

Based on the examination item types introduced in Section~\ref{subsec: classify item types}, we classify patients according to a combination of three attributes: item type, age group, and gender. The classification gives us in total 75 distinct patient classes (see details in Appendix~\ref{app: patient types}). We model the arrival process for each patient class as an independent time-nonhomogeneous Poisson process, with the hourly arrival rates for each estimated empirically from data. 

\subsection{Room Availability}
\label{subsec: room open}

Our initial model that uses the average proportion of opening hours/days to sample the room availability status performs poorly. Further analysis shows that the room opening pattern strongly correlates with individual days. For our final model, we estimate the patterns of room opening hours for each day. In the simulation, we first sample a day from our dataset at random at the beginning of each simulated day. We then use the estimated opening pattern of that day to determine the room availability in the corresponding simulated day.

\subsection{ROUTING}
\label{subsec: ultra-queueing guidance}

As discussed in Section~\ref{subsec: challenge}, models with stylized routing policies perform poorly in capturing the empirical performance, e.g., see Figure~\ref{fig: wait_time_model0}.
We develop a novel two-level hierarchical routing policy, with each level being estimated from data directly. In the first level, we decide on which one of four types of rooms to route the patient to. Then in the second level, we decide on which room within that type to route the patient. Correspondingly, we constructed two prediction models to learn the routing decisions correspond to the two levels. Our analysis shows that adopting the two-level structure is important in calibrating the simulation model, specifically, separating the second level from the first level. This is because assigning patients to different types of rooms (captured by the first level) is mainly from medical consideration. Within a given type of room, the specific room to assign the patients will then largely depend on operational features such as queue length and room availability. With the two-level structure, we are able to differentiate the two types of considerations and obtain an accurate prediction on the routing decisions.

\subsubsection{First Level: Routing to One Type of Rooms}

The first level in our routing component is to assign an incoming patient to one of four room types (R1, R2, R3, R4). In this part, we formulate it as a classification problem, with the true class label as the actual room type from data. We then fit different machine learning models to predict this label, including multinomial logit model, random forest, gradient boosting tree, and neural network. We compare their prediction accuracy performance and pick the one with the best out-of-sample prediction power. The random forest model is found to be the best among all models we test. 
Features used in the random forest model are summarized in Table~\ref{tab: routing features}. 


	\begin{table}[htp]
	    \centering
	    \scalebox{0.8}{
	    \begin{tabular}{|c|c|}
	        \hline
	        feature name & explanation \\
	        \hline
	        age & age of patient\\
	        \hline
	        item type & classified examination item types, encoded as P1 to P6 in Sec \ref{subsec: classify item types}\\
	        \hline
	        arrival hour & hour of the day when patient arrives\\
	        \hline
	        weekday & day of the week when patient arrives\\
	        \hline
            queue length & the number of patient in queues when patient arrives\\
            \hline
            numServer & number of open test rooms when patient arrives\\
            \hline
	    \end{tabular}
	    }
	    \caption{Table of features for routing model: first level in the routing component.}
	    \label{tab: routing features}
	\end{table}

Notice that in this first level, the queue length is generated by summing up all queues in each type of room. Similarly, ``numServer'' -- the number of open rooms, is also computed among all rooms contained in each type. The prediction accuracy of this random forest model is provided in Appendix \ref{app: other routing}. 


\subsubsection{Second Level: Routing to a Specific Room}

After selecting a room type for an arriving patient, we decide on which room within the type to route the patient to. Similarly, we consider it as a supervised learning problem, but now the classification labels are changed to individual rooms. We fit one routing model for each room type, which leads to a total of four models.
Similar to the first level, we test different classification methods for the second level. Again, random forest is shown to be the best. The input features to this second-level random forest model are mostly the same as Table \ref{tab: routing features}, except following two differences: 
\begin{itemize}
    \item \textit{queue length}: queue length refers to the number of waiting patients at an individual test room, instead of the entire room type.
    \item \textit{numServer}: in the first level we adopt numServers to represent room open status within one type. Here we use a binary indicator that shows whether an individual room is open or not in the current hour. 
\end{itemize}
Table~\ref{'tab:RF_par'} summarizes the hyperparameters chosen for random forest models in the two levels. 

\begin{table}[htbp] 
\centering
\scalebox{0.8}{
    \begin{tabular}{|c|c|c|c|c|c|}
    \hline
    Parameter         & First level & \makecell[c]{Second level \\ (room type 1} & \makecell[c]{Second level\\  (room type 2)}& \makecell[c]{Second level\\  (room type 3)}& \makecell[c]{Second level\\  (room type 4)} \\ \hline
    number of estimator       & 100           & 250                         & 150&250&300                          \\ \hline
    bootstrap         & False           & False                       & False  &False&False                     \\ \hline
    criterion         & Gini Index           &Entropy & Gini Index                     & Gini Index 
    &Gini Index\\ \hline
    max features      & $\sqrt{p}$          &$\log_2{p}$ & $\sqrt{p}$                      & $\sqrt{p}$ & $\sqrt{p}$                       \\ \hline
    min samples leaf  & 20           & 10 &1 &1                         & 1                           \\ \hline
    min samples split & 20           & 5 &2 &15 &11                                                  \\ \hline
    max depth         & 9           & 9                           & 9 &10 &10                           \\ \hline
    \end{tabular}
}
\caption{Hyperparameter for random forest model. $p$ is the input dimension of Random Forest model.}
\label{'tab:RF_par'}
\end{table}

\subsection{Service Process}

The service time of each patient is sampled empirically from historical patient records with the same room, hour-of-day, and patient type combination. A patient leaves the system after getting service in the system. We illustrate two additional features in generating service times in our simulation model: random \textit{server break time} and \textit{patient walking time}. 

From both the data and our field observation in the hospital, we find that patients are not always called into the room as soon as the last patient in the queue leaves. It is possible that technicians need to take time to organize materials, adjust the machines, take a short break, or on some occasions, leave the room to deal with urgent situations. To model this server break time, we analyze a sequence of patients served in the busy period of each room, and take the gap between two consecutive patients as the break time. That is, the duration between the service-end time of the previous patient and the service-start time of the next patient. We then fit distributions of the break time for each of the four room types and for each hour of the day. Details of the break time distributions are in Appendix~\ref{app: ultra stat}. Similarly, patients need to spend some time walking to the exam room after being routed there. To estimate this walking time, we use the duration between a new patient's arrival time and his/her service-start time in an idle period. We also fit this distribution for each of the four room types and each hour of the day; see Appendix~\ref{app: ultra stat}.

	


\section{MODEL VALIDATION}
\label{sec:model-valid}

\subsection{Validation on Routing}

\begin{figure}[htp]
	\centering
	\includegraphics[width=0.8\linewidth]{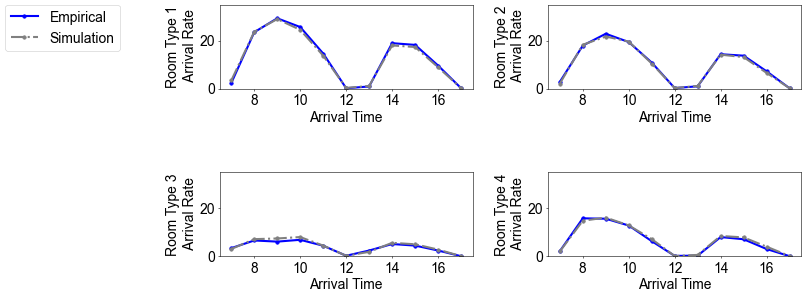}
	\caption{First level: number of routed patients from four types.}
	\label{fig:routingarri1}
\end{figure}

We first discuss validate the two-level routing component. We compare the average number of patients routed to each room type (first level) or each room (second level) with the empirical counterparts. Figure~\ref{fig:routingarri1} compares for the first level. We can see that the predicted number of patients routed to each room type matches closely with the empirical value in each hour. The absolute difference in the hourly arrival rate is between 0.04 and 2.08 across different hours and room types. The relative difference is between 0.60\% and 22.55\%, with the relative difference defined as the absolute difference between simulation and empirical values divided by the empirical value. 


\begin{figure}[htp]
	\centering
    \includegraphics[width=0.7\linewidth]{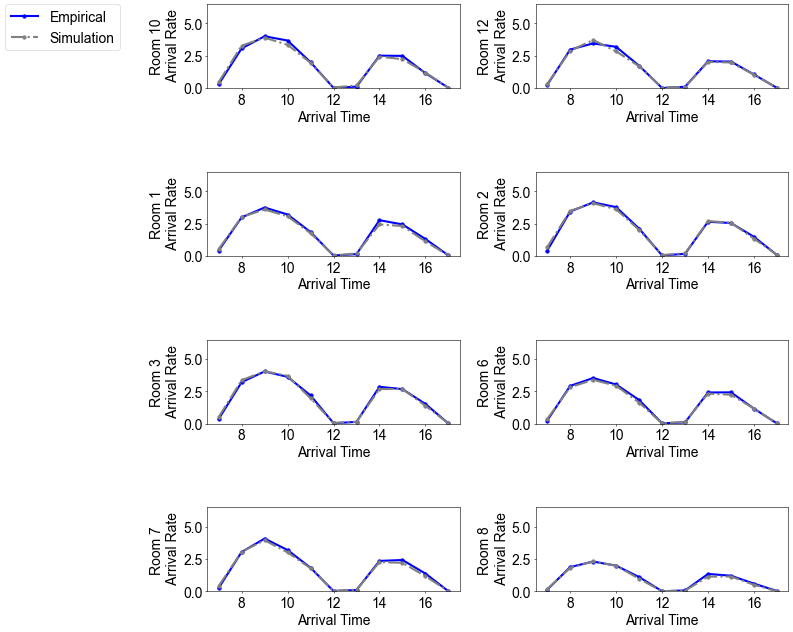} 
    \label{fig: arrRate1}
    \caption{second level: number of routed patients from each room of room types type 1.}
    \label{fig:routingarri2_1}
\end{figure}

For the second level, Figure~\ref{fig:routingarri2_1} compares the empirical and predicted hourly number of patients routed to each Type 1 room. The figures for the other three room types are available in Appendix~\ref{app: other routing}. Again, we can see the predicted values match well with the empirical values. For room type 1, the absolute difference in the hourly arrival rates is between 0 and 0.53 across different hours, with the relative difference between 0\% and 15.31\%. 

\subsection{Validation on Key Performance Metrics}

Figure~\ref{fig:JSQ_performance} in the introduction shows the performance comparison between the empirical performance and the simulation output from the conventional multi-class queue with JSQ routing. 
The middle column in Table~\ref{tab:compare} summarizes statistics on the differences in the key performance metrics. It is worth noting that while the differences (absolute and relative) in the mean waiting time are moderate, the differences in the mean queue length and waiting time distributions are significant, as observed from Figure~\ref{fig:JSQ_performance} and Table~\ref{tab:compare}. The largest relative difference in the hourly mean queue length can be as high as 32\%. 

For our high-fidelity simulation model, Figure~\ref{fig:model_good_waiting} shows a similar set of performance comparison between the simulation output and the empirical performance. The last column in Table~\ref{tab:compare} summarizes statistics on the differences in the key performance metrics in this high-fidelity model. The improvement on the waiting time distribution is the largest: the Kolmogorov-Smirnov difference (KS diff) is reduced from 0.371 to 0.058. The difference in the mean queue length also shows a significant reduction (from 2.23 to 0.82), with the largest relative difference in the hourly queue length reduced to 20\% (from 32\% previously). The mean waiting time also shows some improvement, with the difference reduced from 3.11 minutes to 2.40 minutes. Overall, our high-fidelity model produces much more accurate performance prediction compared to the conventional model. 


\begin{table}[htp]
    \centering
    \begin{tabular}{c|c|c}
                    & Conventional model & High-fidelity model \\
                    \hline 
Diff in mean queue length (patient)  &  2.23 (16.49\%) & 0.82 (6.73\%)  \\
Diff in mean waiting time (min)      &  3.11 (21.25\%) & 2.40 (19.34\%) \\
KS diff in waiting time distribution & 0.371 & 0.058
    \end{tabular} 
    \caption{Comparison on simulation output. The numbers in the parenthesis for the first two rows are the corresponding relative differences.}
    \label{tab:compare}
\end{table}
\vspace{-0.2in}

\section{CONCLUSION}
\label{sec:conclud}
For many health service system, conventional modeling-based analysis has its limitation as being inflexible in capturing complicated behavioral features, e.g., the routing decisions in our studied setting. This calls for the need of developing high-fidelity simulation models that can capture these complicated features and make more accurate predictions. Taking our healthcare partner as an example, the salient features in the operations of their ultrasound center would make a conventional modeling-based analysis infeasible. These features include patient and server heterogeneity, time-dependency, complicated patient routing, and server vacation. Thus, it is imperative for us to develop a high-fidelity model, instead of a stylized model, to predict the performance of the ultrasound service system.

It is important to note that our simulation model approximates the real system at the appropriate granular level so that the model is both predictive as well as tractable. To that end, we carefully examined the historical data and identify behavioral patterns which helps simplify the simulation model without compromising the prediction accuracy. For example, to capture the highly complex routing behavior, we directly used machine learning methods to learn the routing behavior from data. Then we choose a hierarchical routing policy that closely approximates the current policy and also can be efficiently simulated. In addition, we develop a clustering-based classification for both patients and servers (rooms) so that items in the same class share similar operational characteristics (e.g., service time distributions). This classification captures the patient and server heterogeneity without significantly increasing the problem size. Numerical results show that the output of our simulation model can accurately recover the real system states. This increases the credibility of the potential managerial implications. In particular, this simulation model can provide direct help for the hospital manager to forecast the intra-day demand and make better staffing and scheduling decisions. Moreover, many of the features that we identified and modeled in this paper, such as time-varying arrival and service rates, patient and server heterogeneity, server vacation, etc., are likely to exist in other hospitals. Thus, the methods we proposed in developing the simulation model have potentials to be generalized to study other ultrasound centers and service systems. 


One future research direction is to develop an analytically tractable model that still captures some of the critical features in our simulation model. This would allow us to further understand the underlying driving forces of the ultrasound system and develop generalizable managerial insights. Another potential research direction is to develop a decision support system, based on this simulation model, that could facilitate the hospital manager to make real-time scheduling decisions such as the one in \cite{chen2021saa}.

\bibliographystyle{plain}
\bibliography{references}

\newpage
\appendix

\begin{center}
    \Large\textbf{Appendix}
\end{center}

\section{Description of data source merging}\label{app: data merging}

There are two sets of data that we used in our simulation, which are known as \textit{dataset A} and \textit {dataset B}. Both of them record ultrasound test information from Jan 2018 to Dec 2018 but contain different features of each record. To integrate these two datasets properly, here we first specify the attributes from \textit{A} and \textit{B}:
\begin{itemize}
    \item Attributes that are shared by both datasets (the names outside parenthesis are from \textit{dataset A}, and names inside parenthesis come from \textit{dataset B}): department, checking items, patient ID, registration time (queueing starting time), service-calling time (queueing ending time)
    
    \item Attributes only appearing in \textit{dataset A}: report generation time, report verification time
    
    \item Attributes only appearing in \textit{dataset B}: patient gender, patient age, technician, test rooms
\end{itemize}
To determine the waiting time and service time of a patient, there are three timestamps that are necessary to record: arrival time, service starting time, service ending time. By checking some samples from both datasets and consulting the hospital managers, we find:
\begin{itemize}
    \item Arrival time is represented in registration time (from \textit{dataset A}), which is the same as queueing starting time (from \textit{dataset B}).
    
    \item Service starting time is written in service-calling time (from \textit{dataset A}), which is the same as queueing ending time (from \textit{dataset B}).
    
    \item Service ending time is contained in report generation time (from \textit{dataset A}).
\end{itemize}
To maintain the consistency of our modeling, we adopt arrival time, service starting time and service ending time as the unique names of queueing timestamps. Furthermore, \textit{dataset A} and \textit{dataset B} are merged by (1) aligning registration time with queueing starting time and (2) aligning service-calling time with queueing ending time. Finally, attributes from the integrated data can be concluded in the following three types:
\begin{itemize}
	\item \textbf{Timestamps}: arrival time, service starting time and service ending time.
	\item \textbf{Patient information}: patient ID, patient gender, patient age, department.
	\item \textbf{Test information}: checking items,  test rooms, technician.
\end{itemize}

\section{Prediction Performance of Routing Models}\label{app: other routing}

\subsection{Prediction performance of random forest models}

To train and validate the two-level routing model, we randomly sample 4/5 of our data as a training dataset, and the rest 1/5  serves as a test dataset. There are two perspectives of performance evaluation. 
We assess the model by some traditional measurements, for example, multi-class AUC (area under the ROC curve, where ``ROC'' refers to receiver operating characteristic curve).

\noindent\textbf{First level. }
 The ovr (one verses rest) AUC and accuracy are shown in the following table
	\begin{table}[htp]
	    \centering
	    \begin{tabular}{c|c}
	    \hline
	         &  on four types of rooms\\
	         \hline
	       ovr AUC (train)  & 0.701\\
	       ovr AUC (test)  & 0.37\\
	       accuracy (train) & 0.953 \\
	       accuracy (test)  & 0.36\\
	       \hline
	    \end{tabular}
	    \caption{AUC and accuracy from first level routing model}
	    \label{tab:my_label}
	\end{table}
	
	Furthermore, we adopt the permutation test to evaluate the importance of different features. The results are shown as:
	\begin{figure}[htp]
		\centering
		\includegraphics[width=0.5\linewidth]{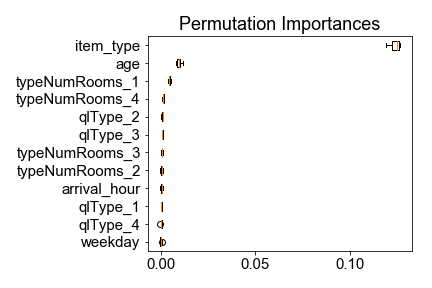}
		\caption{Permutation importance for each feature in the model.}
		\label{fig:routingimportance}
	\end{figure}

\noindent\textbf{Second level. }
	The AUC (`ovr') and accuracy are shown in Table~\ref{tab:layer2auc}, and the hourly arrival rate comparison from each room of room type 2,3,4 are showed in Figure~\ref{fig:arrRate234} 
	\begin{table}[htp]
		\centering
		\begin{tabular}{c|cccc}
		\hline
		&Type 1&Type 2&Type 3&Type 4\\
		\hline
		AUC (test set)	&0.713& 0.705& 0.881& 0.830 \\
		AUC (train set) &	0.774& 0.744& 0.929& 0.868 \\
		accuracy (test set)&	0.17& 0.27& 0.38& 0.39 \\
		accuracy (train set)&	0.20& 0.28& 0.43& 0.42 \\
		\hline
		\end{tabular}

	\caption{AUC and accuracy from second level routing model.}
	\label{tab:layer2auc}
	\end{table}
	
\subsection{Other routing models}
We have tried several other machine learning methods for the routing component: multinomial logit model (MNL), extreme gradient boosting tree (xgb), and Neural Network (NN). The following table gives the performance at the second-level routing.

\begin{table}[htp]
		\centering
		\begin{tabular}{c|cccc}
		\hline
		&Type 1&Type 2&Type 3&Type 4\\
		\hline
		AUC (test set) of MNL	&0.721& 0.711& 0.852& 0.818 \\
		AUC (test set) of xgb&	0.719& 0.702& 0.872& 0.821 \\
		AUC (test set) of NN&	0.718& 0.708& 0.854& 0.822 \\
		\hline
		\end{tabular}
	\caption{AUC from second level routing model with other models.}
	\end{table}
	

\subsection{Validation on routing}

\begin{figure}[htp]
    \centering
    \begin{subfigure}[b]{\textwidth}
        \centering
        \includegraphics[width=0.65\linewidth]{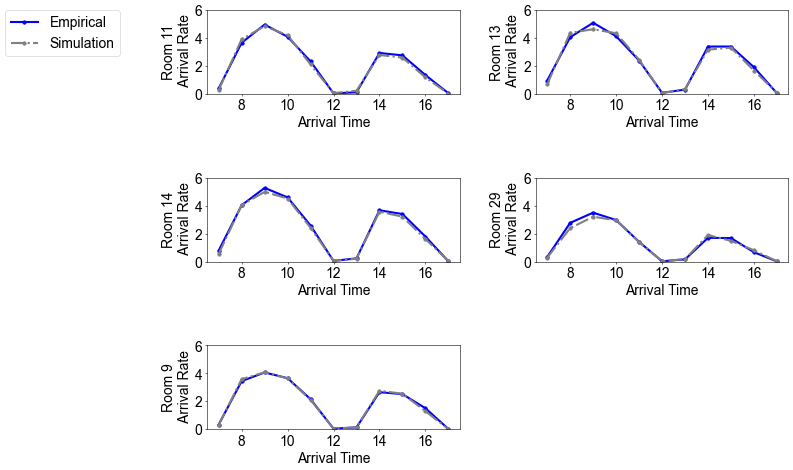}
        \caption{Room Type 2}
        \label{fig: arrRate2}
    \end{subfigure}%
    
    \begin{subfigure}[b]{\textwidth}
        \centering
        \includegraphics[width=0.65\linewidth]{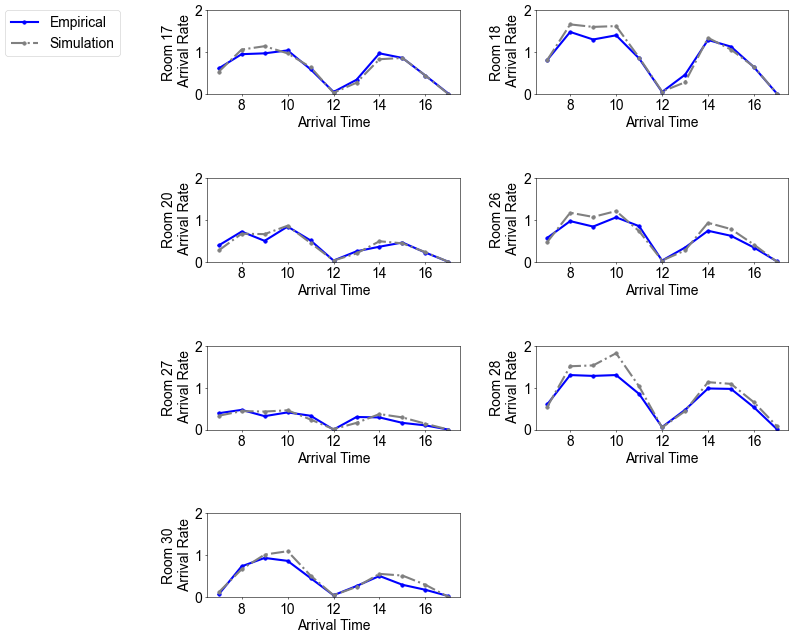}
        \caption{Room Type 3}
        \label{fig: arrRate3}
    \end{subfigure}%
    
    \begin{subfigure}[b]{\textwidth}
        \centering
        \includegraphics[width=0.65\linewidth]{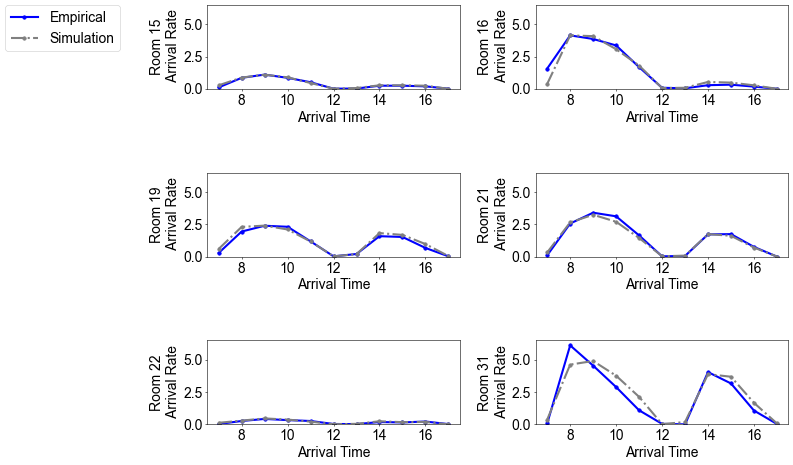}
        \caption{Room Type 4}
        \label{fig: arrRate4}
    \end{subfigure}%
    \caption{Hourly Arrival Rate from Each Room of Room Type 2,3,4.}
    \label{fig:arrRate234}
\end{figure}

Figures \ref{fig:arrRate234} compares the average number of patients routed to each room from the predicted routing component, for room type 2, 3, and 4, respectively, versus the empirical values. The average relative difference of arrival rate is 6.50\% for room type 2, 20.37\% for room type 3(since the arrival rate values of these rooms are quite small), and 17.77\% for room type 4. 


\section{Functions of Clinics}
\label{app: functions of clinics}
In the ultrasound station, because there are different inspection equipment in different clinics, different clinics are mainly responsible for various sources of patients. Table~\ref{tab:function-clinic} summarizes the main specialties of patients that each examination room serves.

\begin{table}[htp]
	\centering
	\begin{tabular}{c|c}
	\hline
	Room Number& Patient Specialty\\
	\hline
	1-2	& Salpingogram, Pelvic Ultrasound, Gynecology\\
	3-7 & General Department, Superficial Ultrasound, Obstetrics , Gynecology\\ 
	8-14 & Obstetrics , Gynecology, Pelvic Ultrasound\\
	15 & General Department\\
	16 & Expert Clinic\\
	17-20 & Cardiovascular \\
	21 & General Department\\
	22 & Intervention Clinic\\
	27-32 & Fetal Screening\\ 
	\hline
	\end{tabular}
\caption{Main functions of examination rooms.}
\label{tab:function-clinic}
\end{table}

\section{Grouping and clustering patient types}
\label{app: patient types}

Among patients, there are 6 exam item types; see Appendix~\ref{app:class-exam-type}. 
The age is divided into 10 types:
0-0.5;  0.5-5.5;  5.5-10.5;  10.5-20.5;  20.5-30.5;  30.5-40.5;  40.5-50.5;  50.5-60.5; 60.5-70.5;  70.5 and larger. There are 2 genders: male and female. Depending on these three attributes, we have $6\times10\times2$ combinations. However, some combinations have no patients in empirical data. For example, for item types P1,P3,P4, we only have female patients. Therefore, we end up having a total of 75 patient classes.

\subsection{Classification of Exam Items}
\label{app:class-exam-type}

\begin{figure}[htp]
		\centering
\includegraphics[width=0.3\linewidth]{P1.png}
\includegraphics[width=0.3\linewidth]{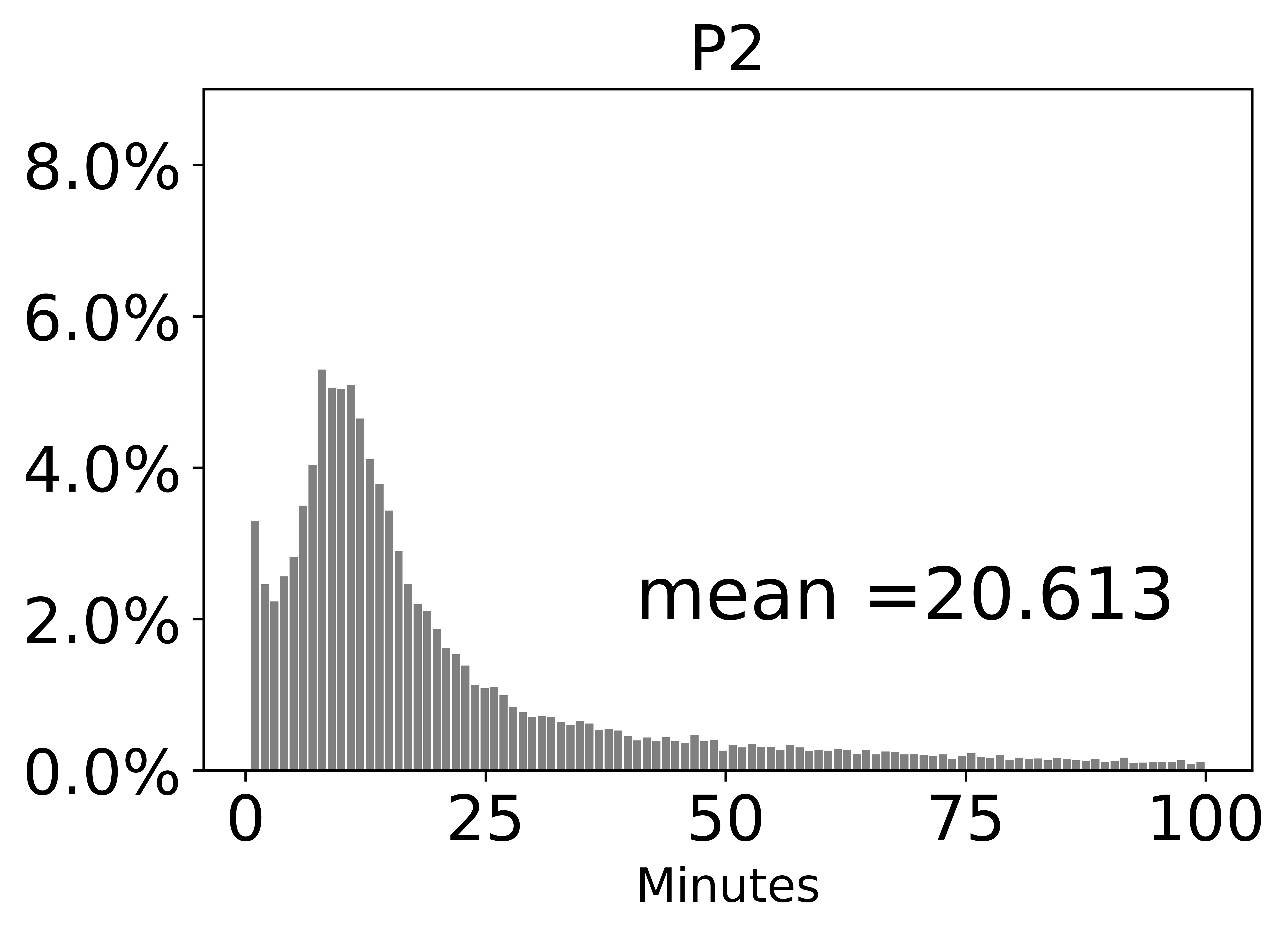}
\includegraphics[width=0.3\linewidth]{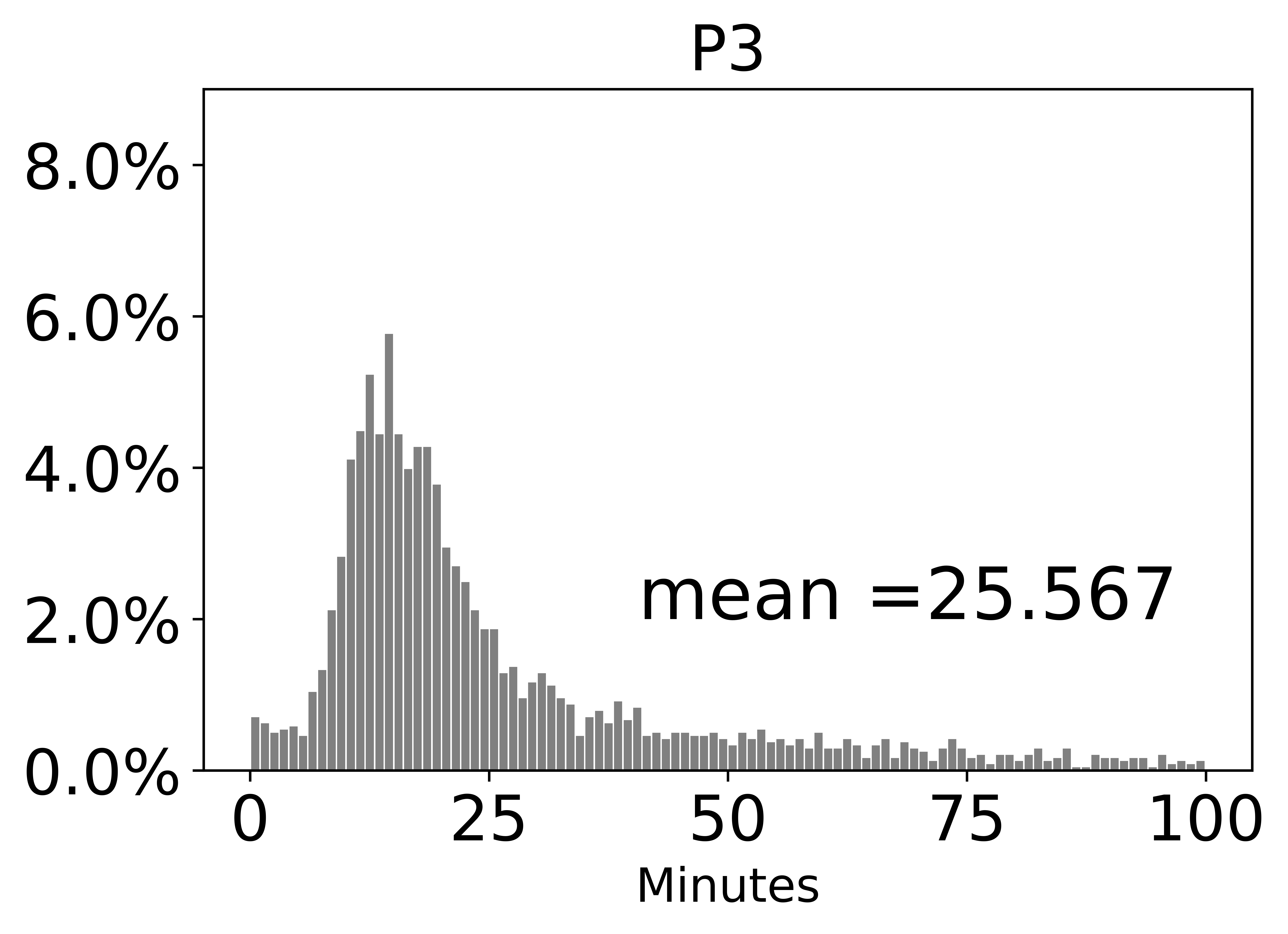}\\
\includegraphics[width=0.3\linewidth]{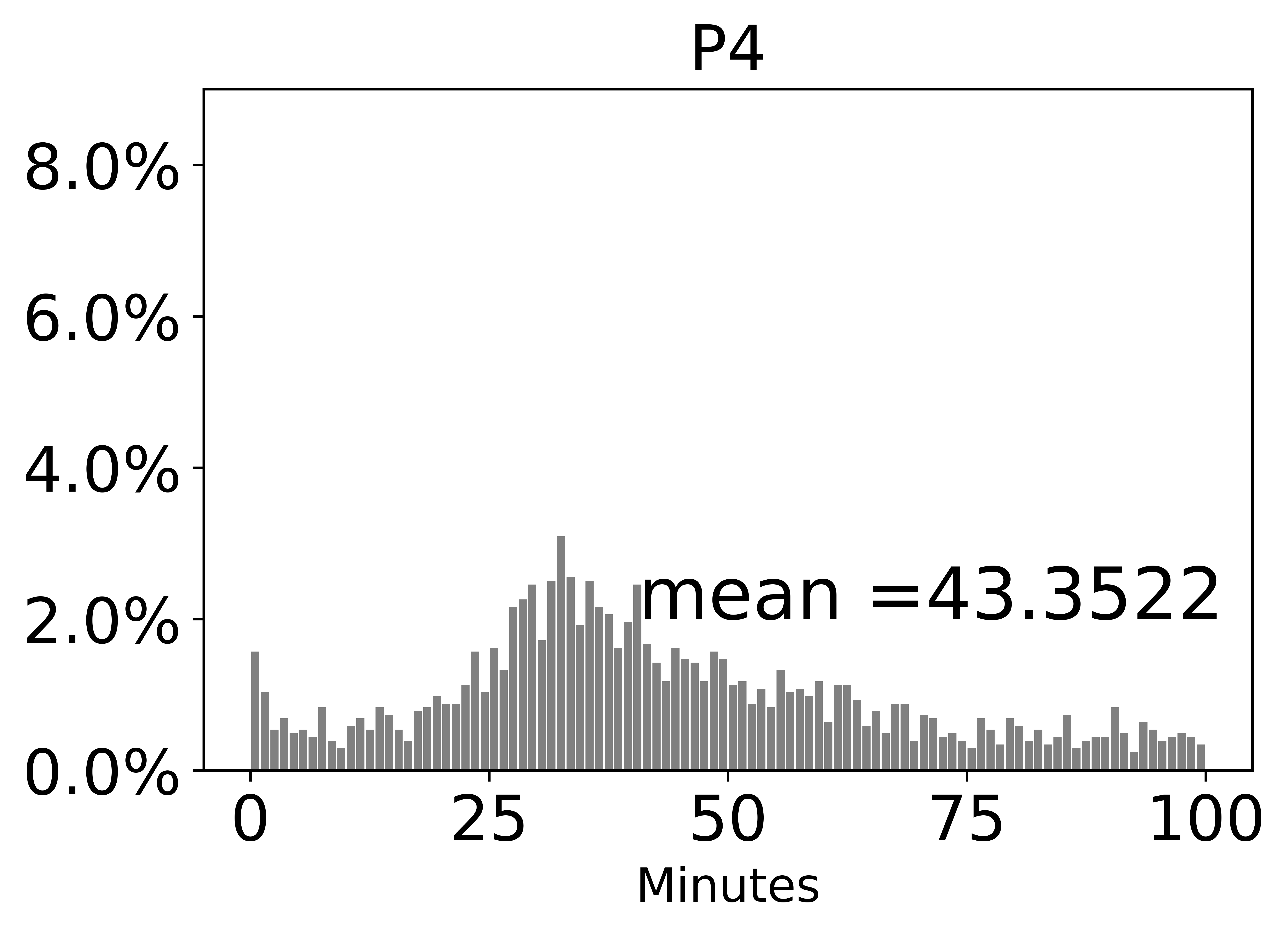}
\includegraphics[width=0.3\linewidth]{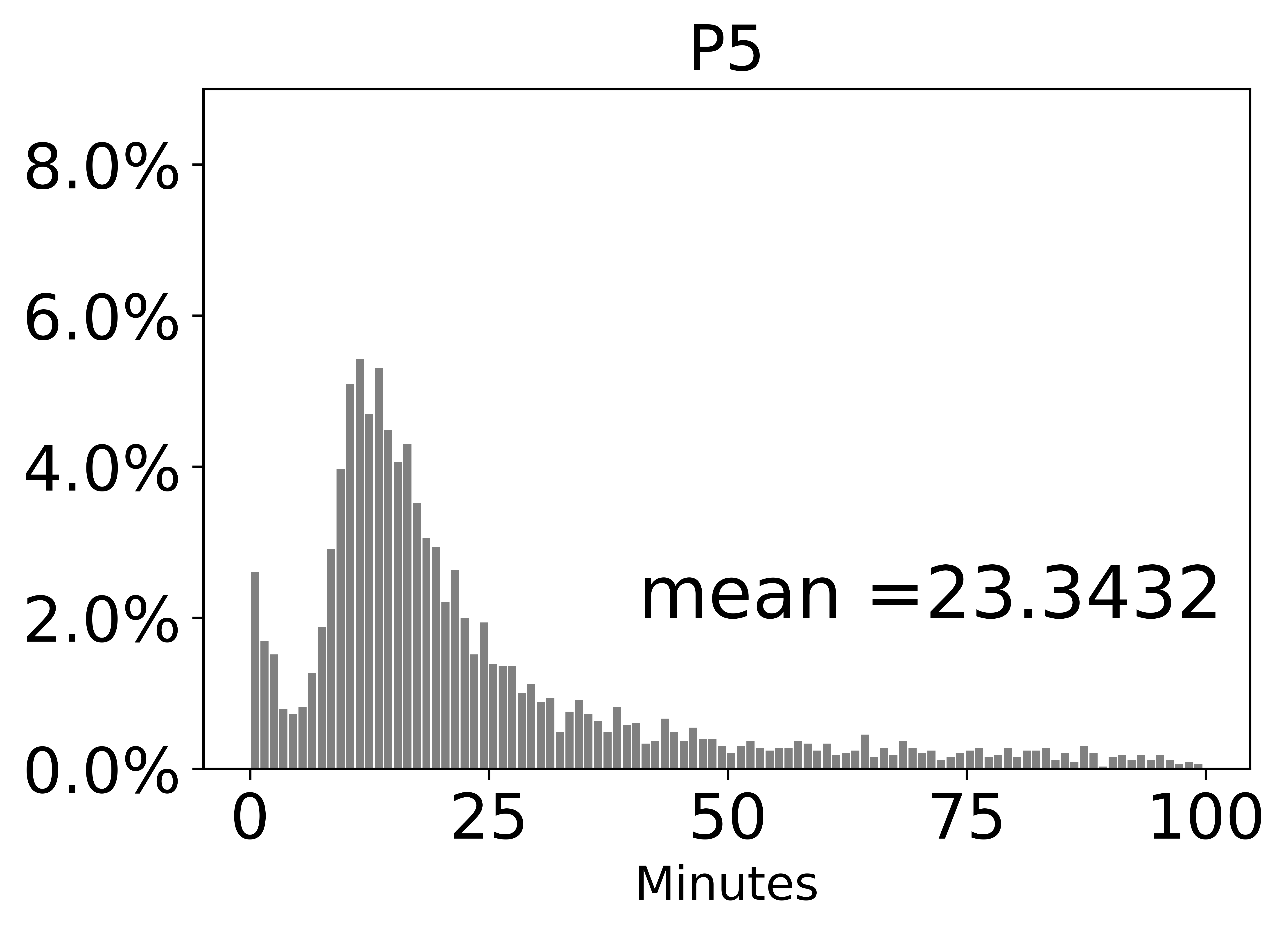}\includegraphics[width=0.3\linewidth]{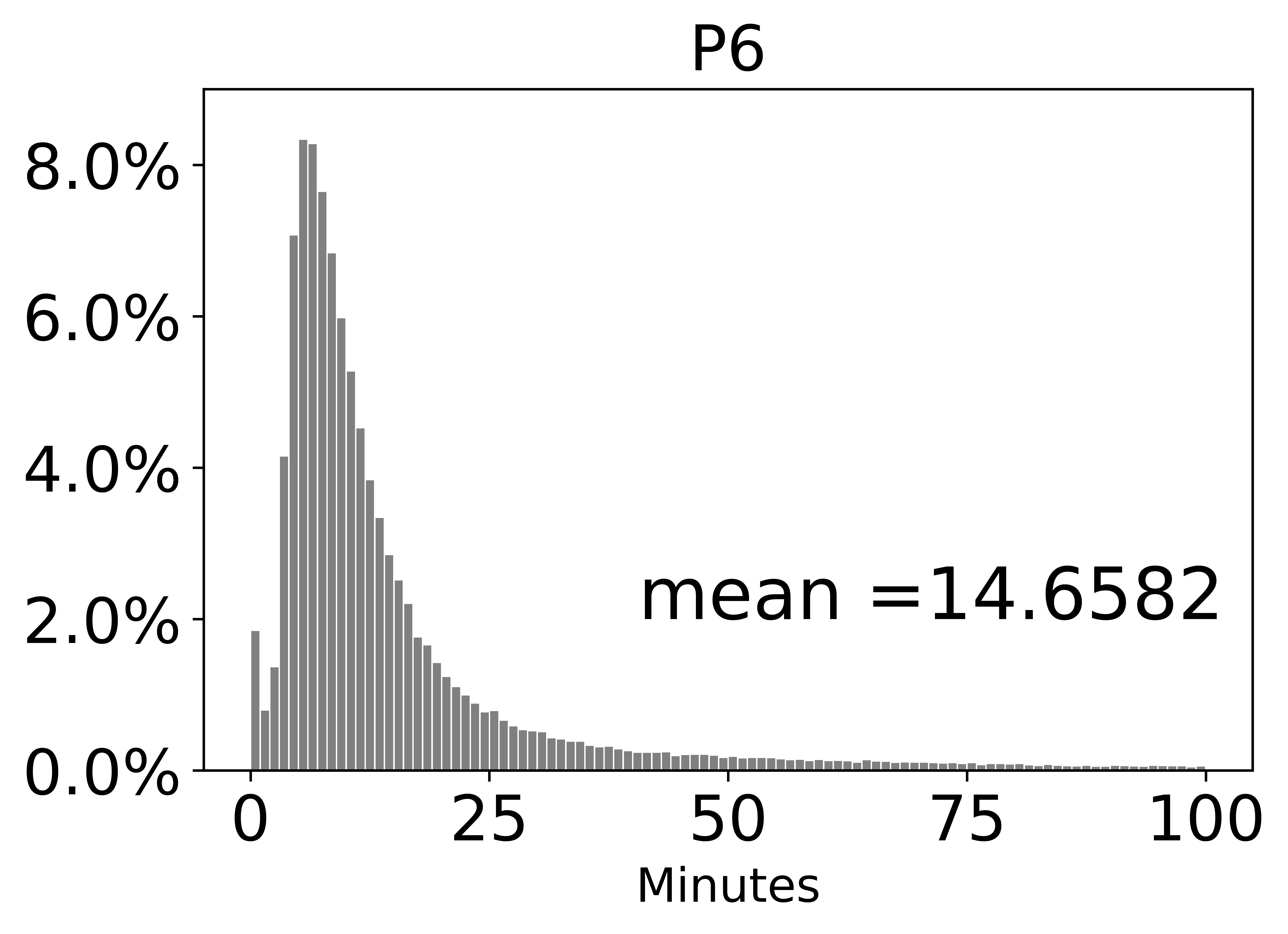}
\caption{Service time distribution for the six groups of exam item groups as given in Table~\ref{tab: classify patients}.} 
\label{fig: Service time distribution}
\end{figure}

We performed Expectation-Maximization Clustering Algorithms using Gaussian Mixture Models for patients who only come with one examination item by their service time (mean and std) and empirical routing probability. In addition, we separate the 17.48\% of patients who need multiple examination items into another category. Table~\ref{tab: classify patients} shows the total six groups of examination items classified from the clustering algorithm. Figure~\ref{fig: Service time distribution} plots the service time distributions for the six groups of examination items. In the simulation model, we will use the combination of age group, gender, and these six exam groups to classify patients. 


        \begin{table}[htp]
	    \centering
	    \begin{tabular}{|c|c|c|}
	        \hline
	        Examination Type & Corresponding Inspection Items & Proportion \\
	        \hline
	        P1 & Intracavitary three-dimensional ultrasound imaging & 3.16\% \\
	        \hline
	        P2 & Patients who need to take more than one item & 17.48\% \\
	        \hline
	        P3 & NT & 2.24\% \\
	        \hline
	        P4 & Four-dimensional fetal secondary ultrasound & 1.5\% \\
	        \hline
	        P5 & Heart color Doppler combination & 2.14\% \\
	        \hline
	        P6 & Other patients who only have one inspection item & 73.48\% \\
            \hline
	    \end{tabular}
	    \caption{Classification of Patients by Examination Items.}
	    \label{tab: classify patients}
	\end{table}

\section{Additional statistics for ultrasound simulation}\label{app: ultra stat}
\begin{itemize}

\item Break Time Analysis: The break time at different types of rooms is shown as:

\begin{figure}[htp]
	\centering
	\includegraphics[width=0.4\textwidth]{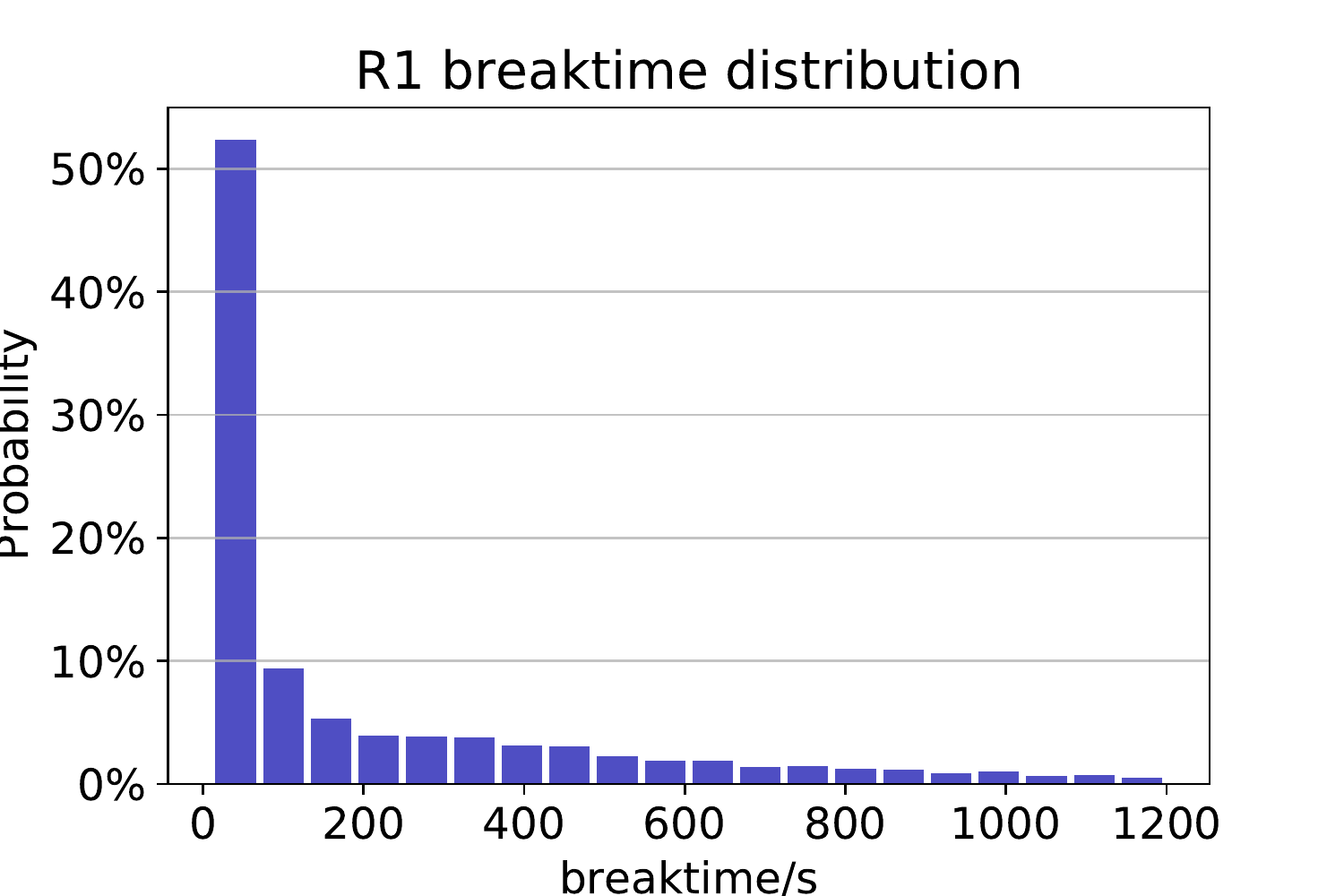}
	\includegraphics[width=0.4\textwidth]{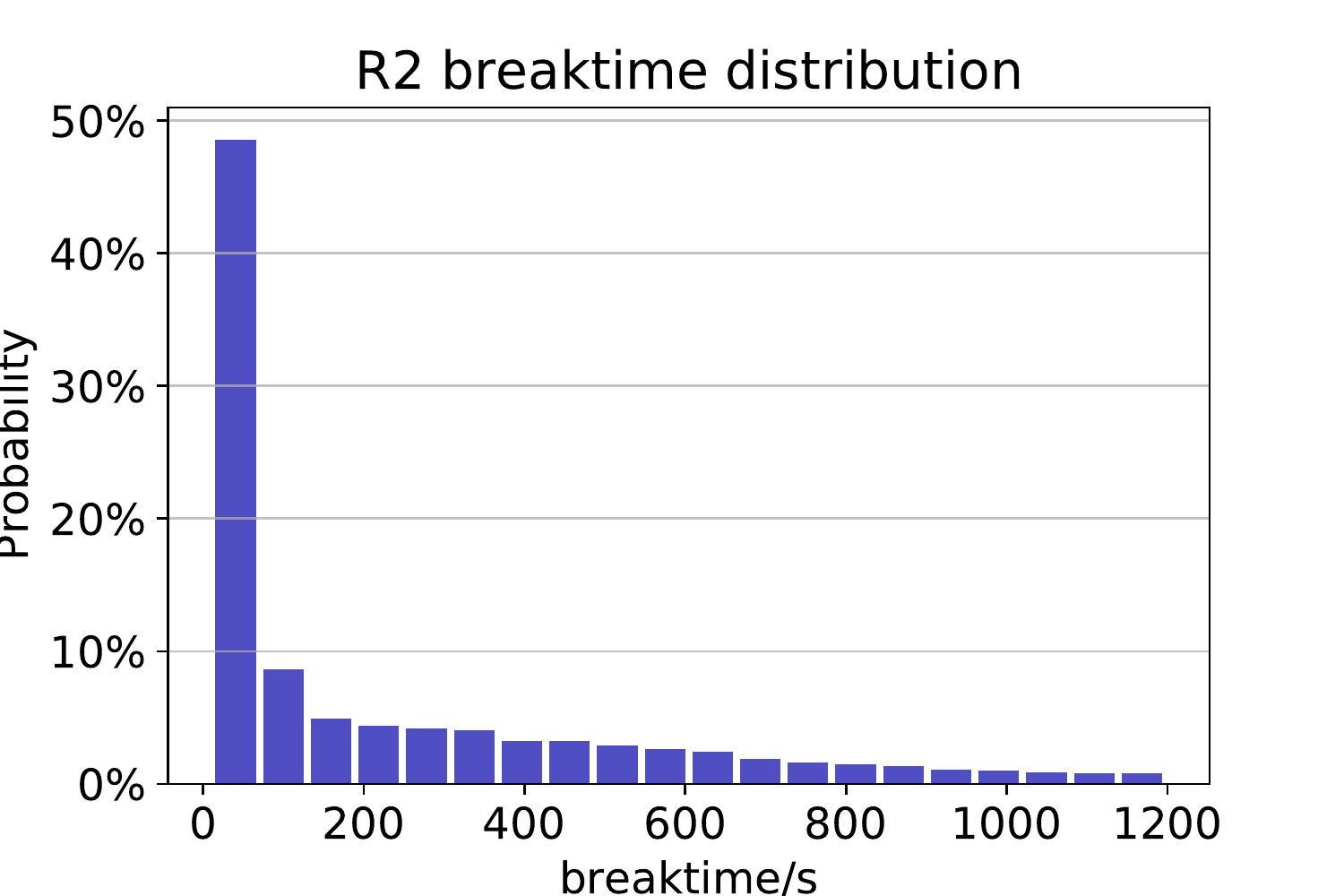}
	\includegraphics[width=0.4\textwidth]{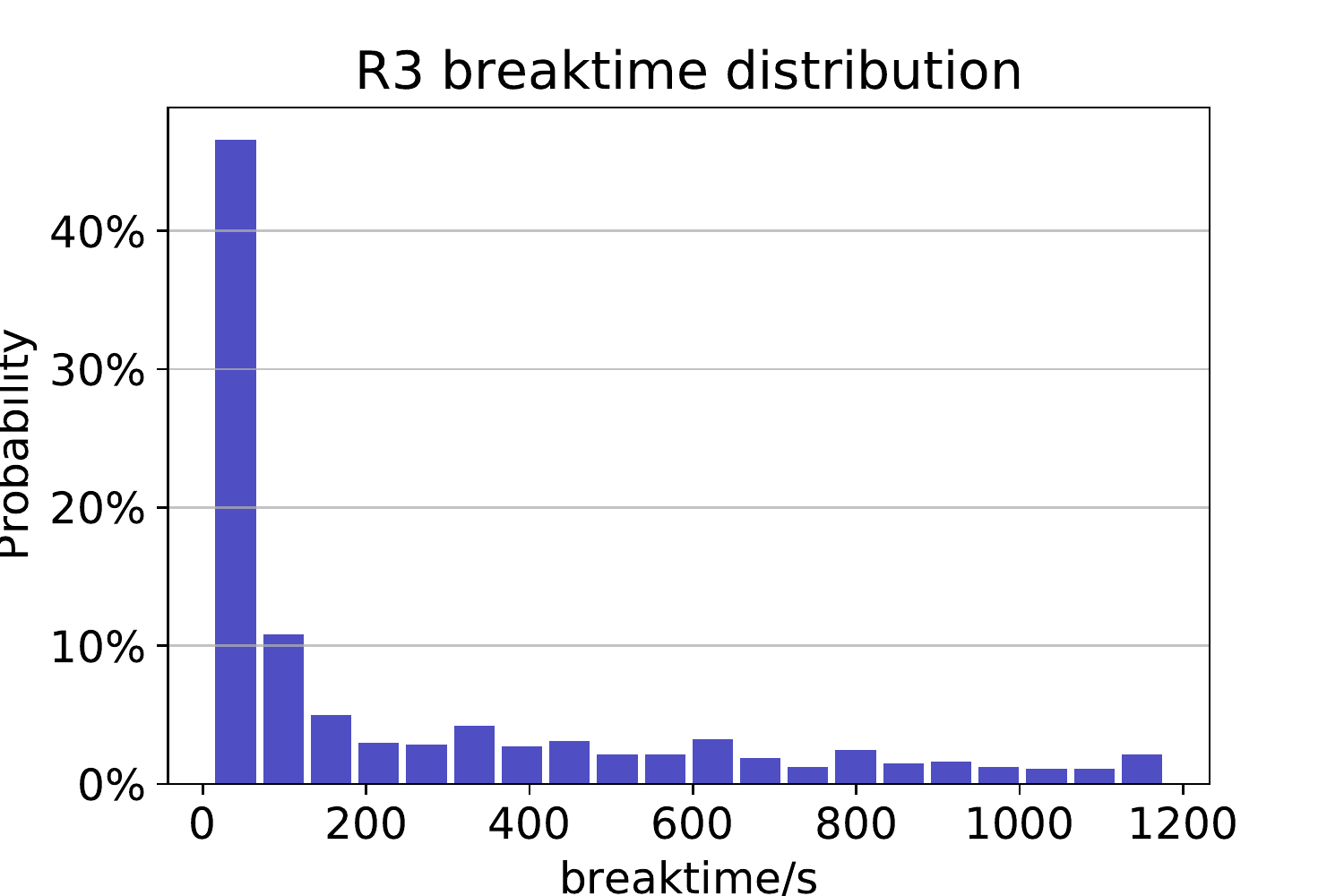}
	\includegraphics[width=0.4\textwidth]{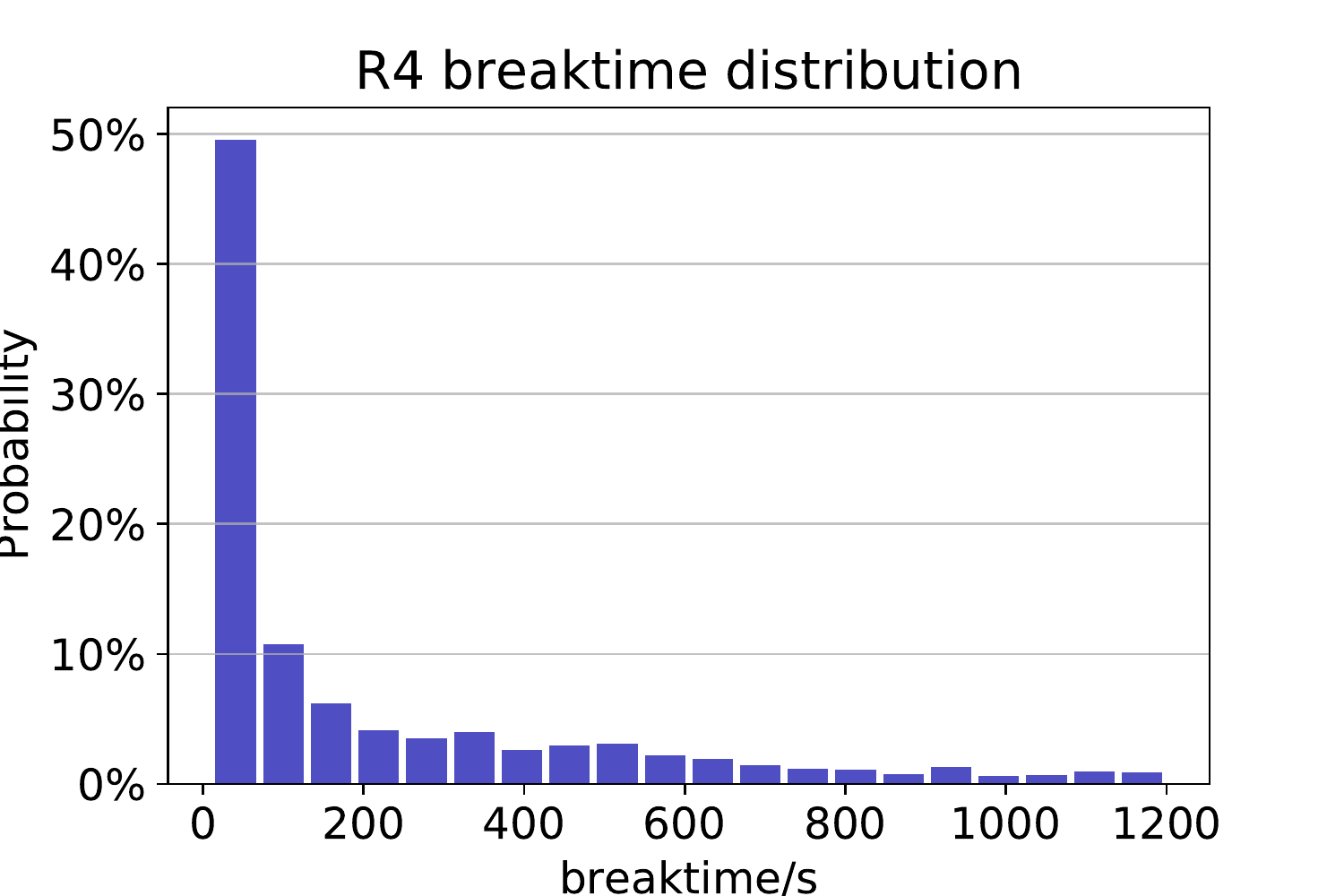}
	\caption{The break time distribution by each type of room. If the arrival time of $n+1_{th}$ patient advanced of the end time of $n_{th}$ patient and the start time of $n+1_{th}$ patient minus the end time of $n_{th}$ patient larger than $10s$, then we deem a break happens and the break time equals to the start time of $n+1_{th}$ patient minus the end time of $n_{th}$ patient.}
\end{figure}
\newpage
\item Break Time Analysis: The break time at different hours is shown as:

\begin{figure}[htp]
	\centering
	\includegraphics[width=0.3\textwidth]{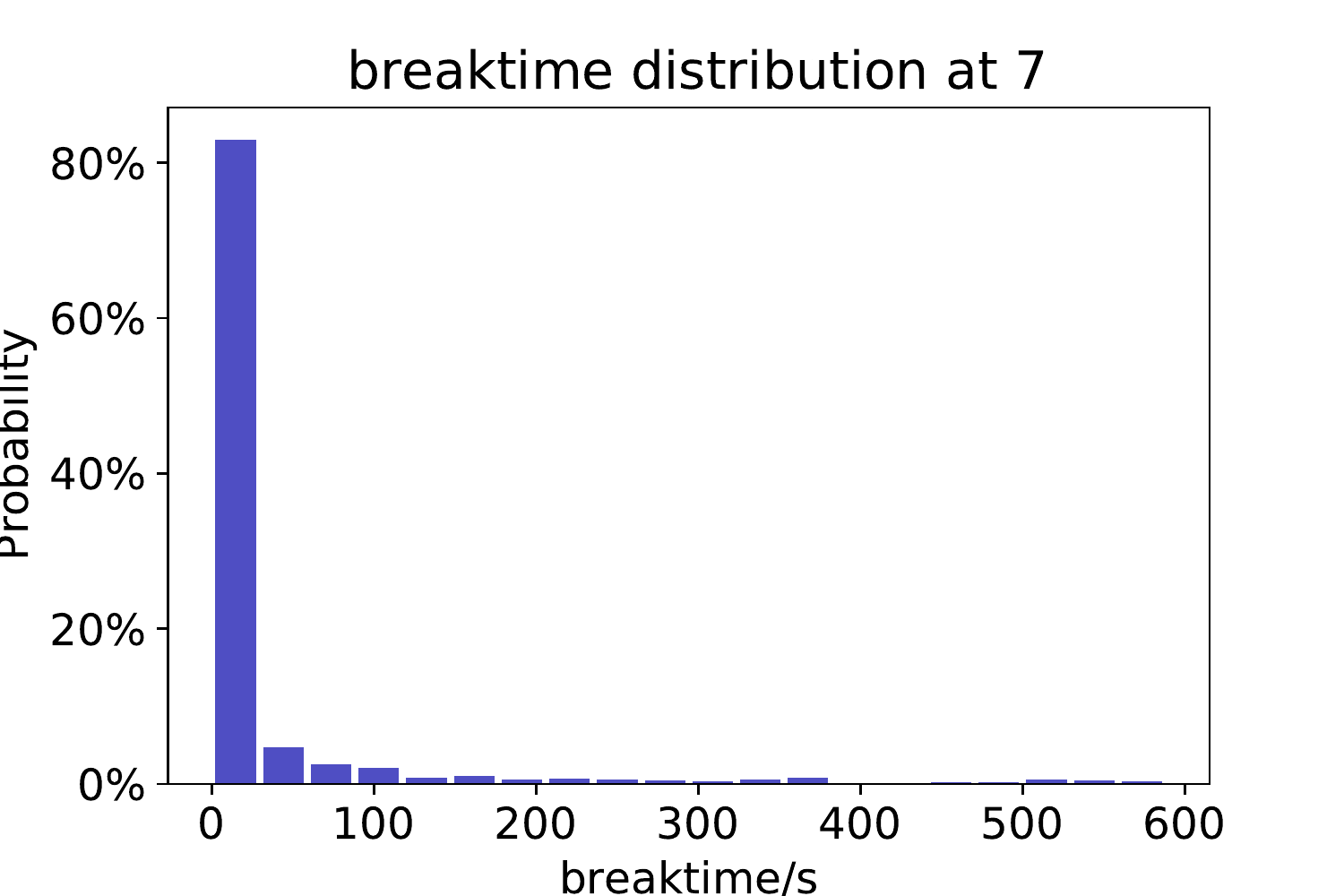}
	\includegraphics[width=0.3\textwidth]{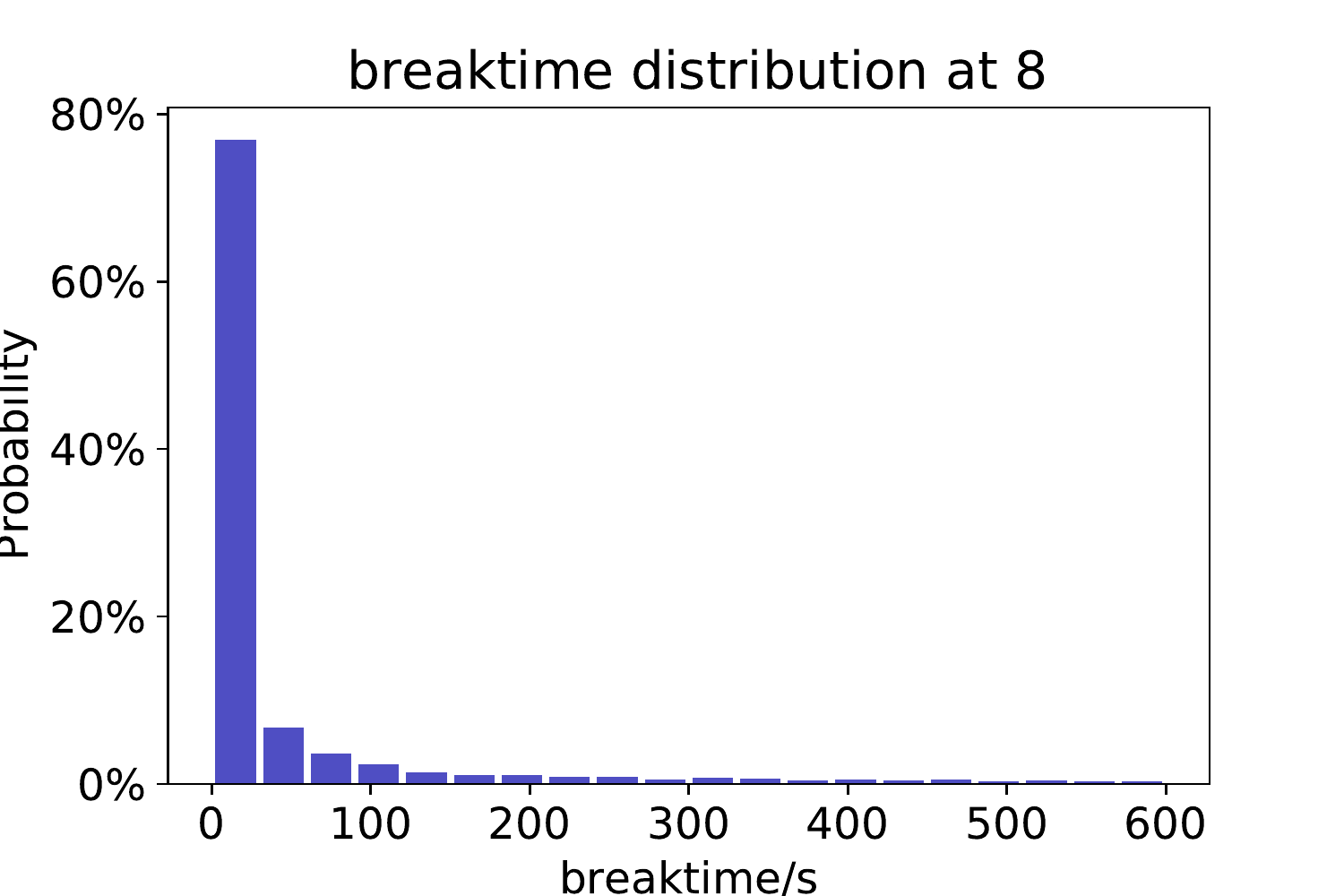}
	\includegraphics[width=0.3\textwidth]{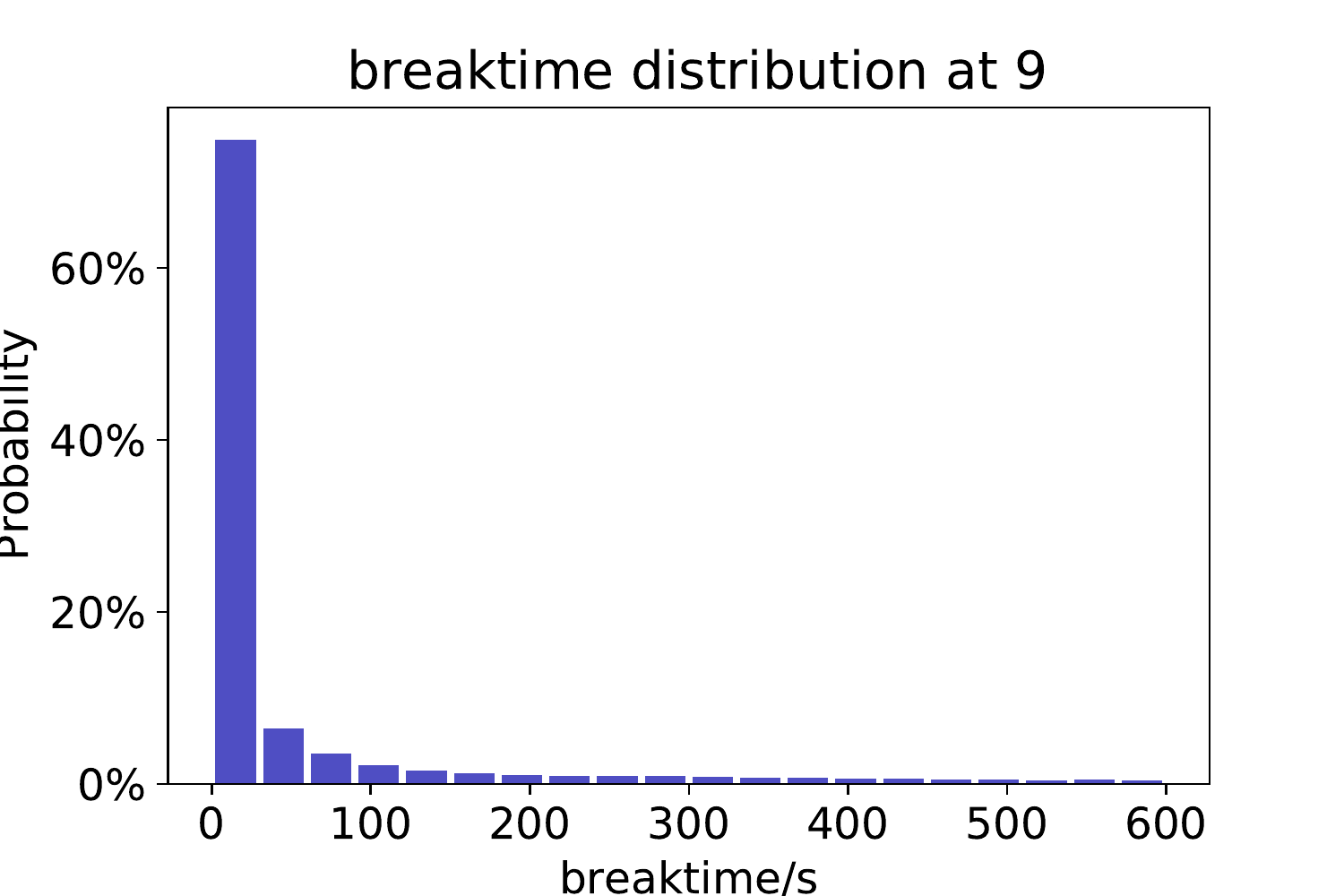}
	\includegraphics[width=0.3\textwidth]{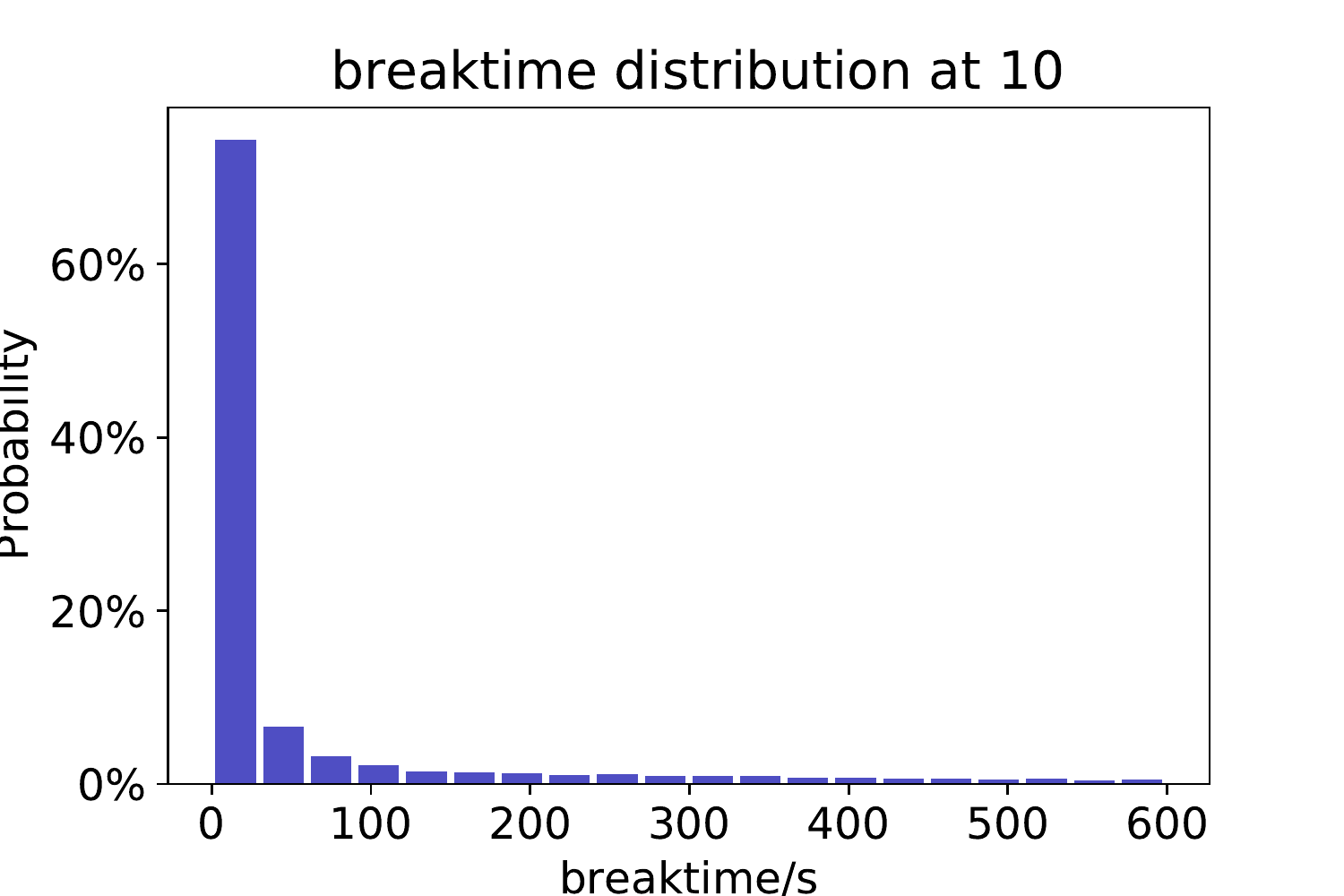}
	\includegraphics[width=0.3\textwidth]{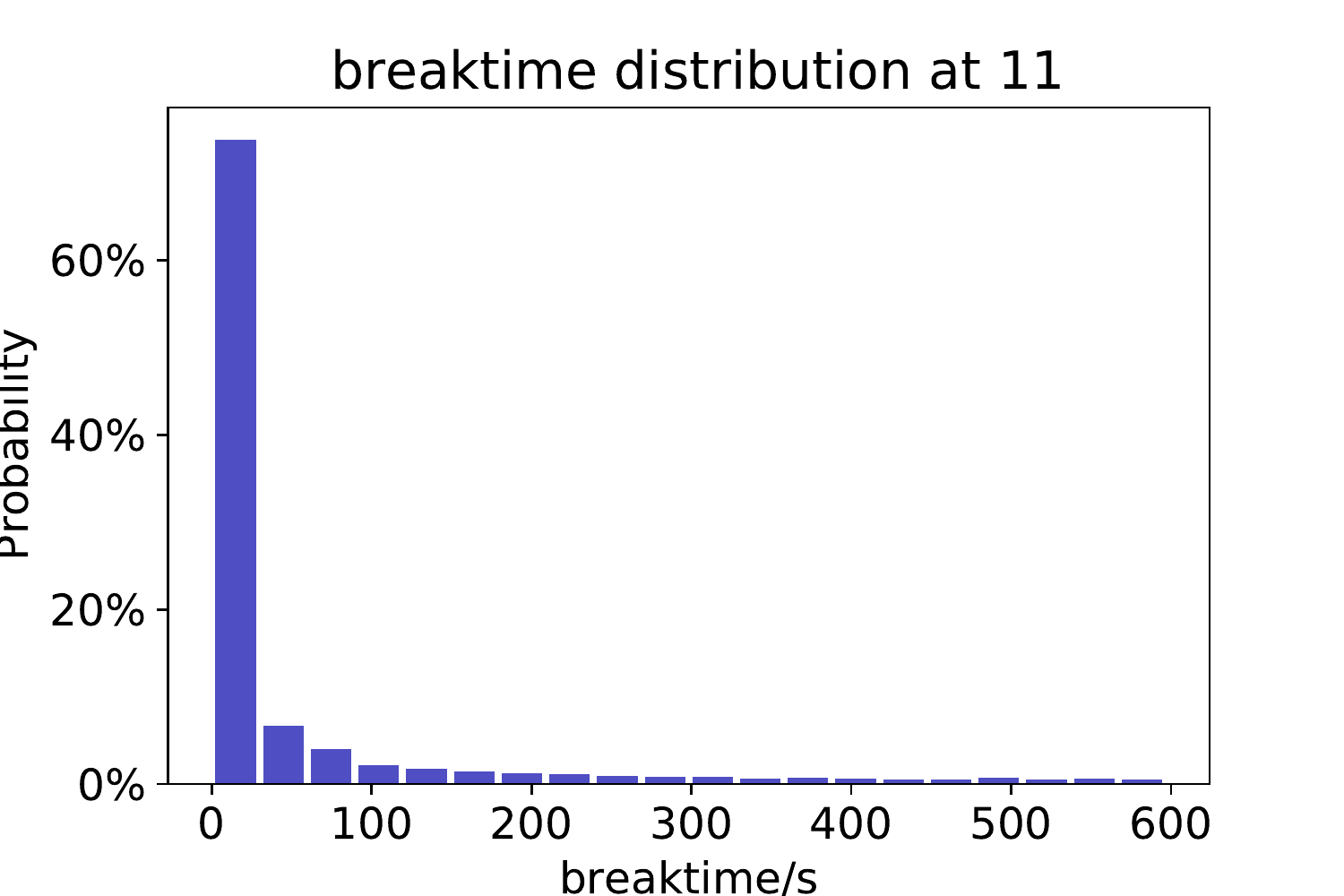}
	\includegraphics[width=0.3\textwidth]{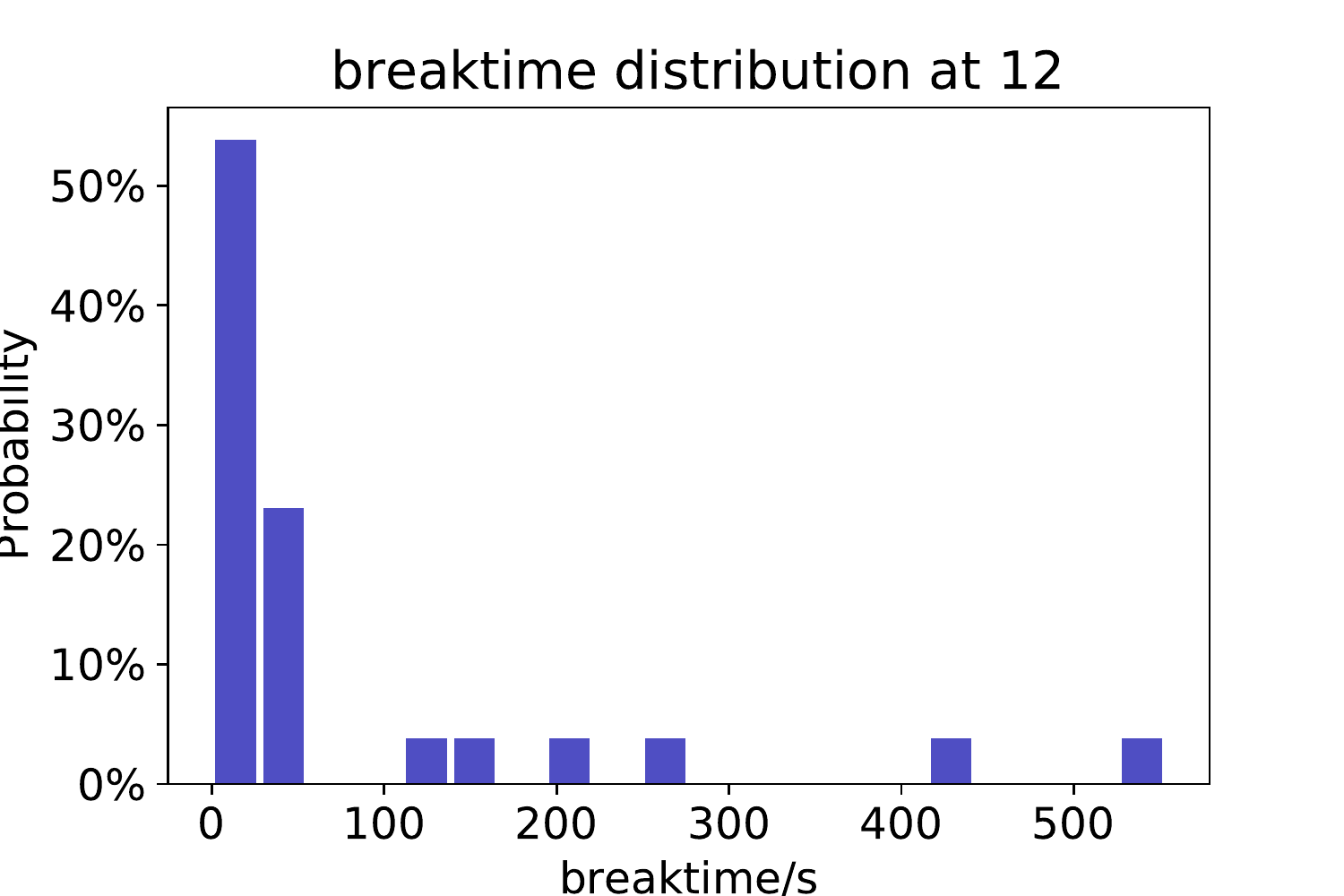}
	\includegraphics[width=0.3\textwidth]{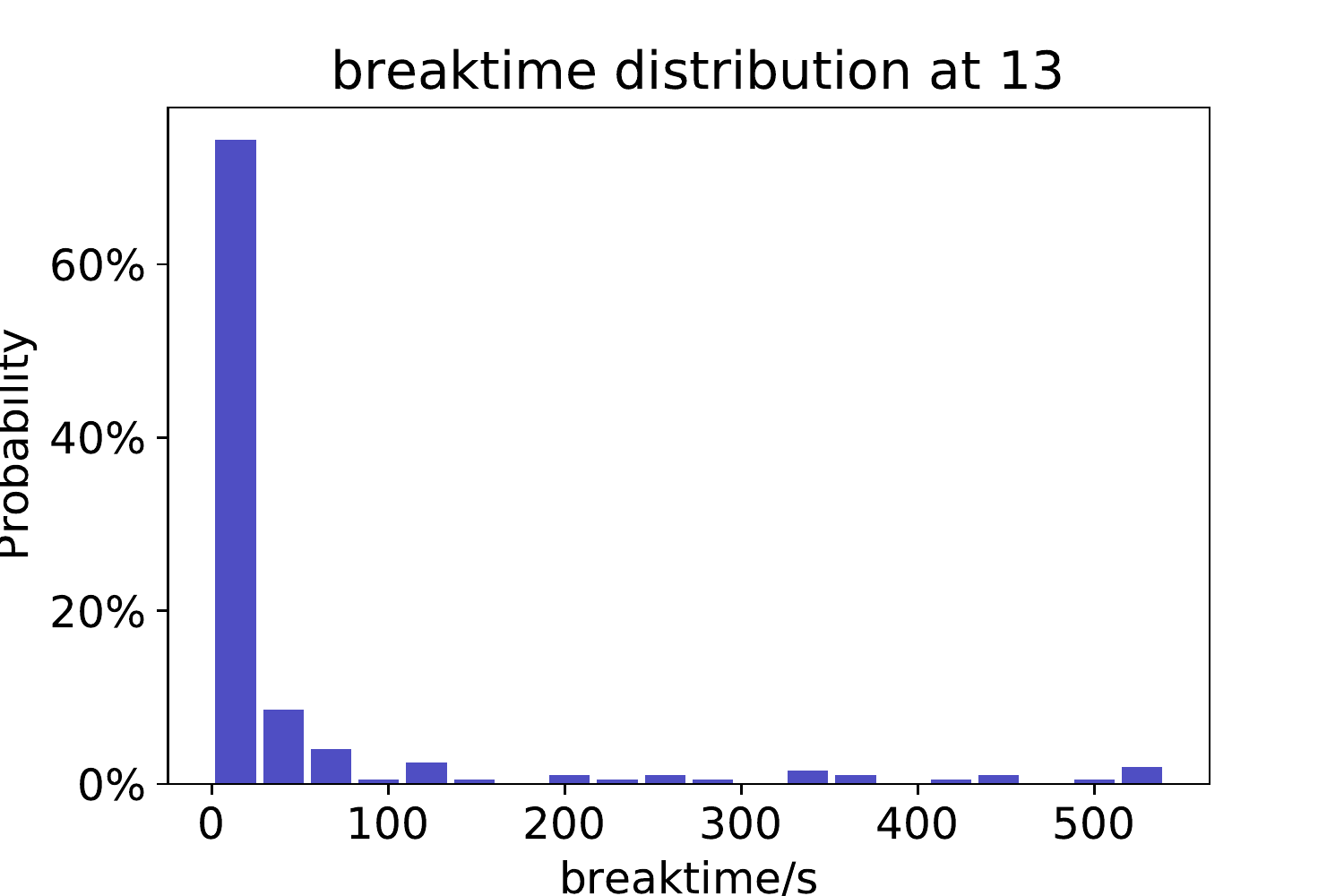}
	\includegraphics[width=0.3\textwidth]{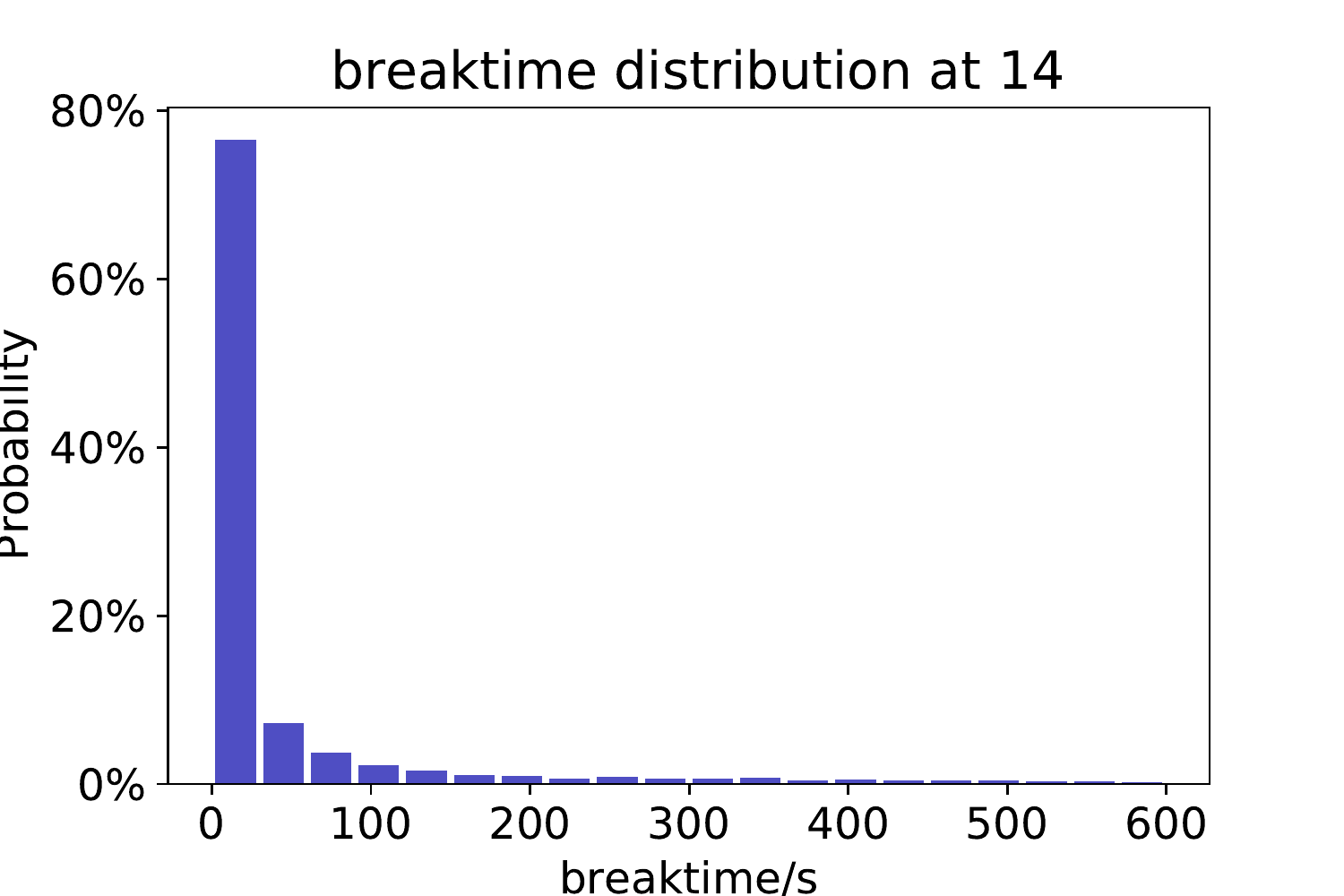}
	\includegraphics[width=0.3\textwidth]{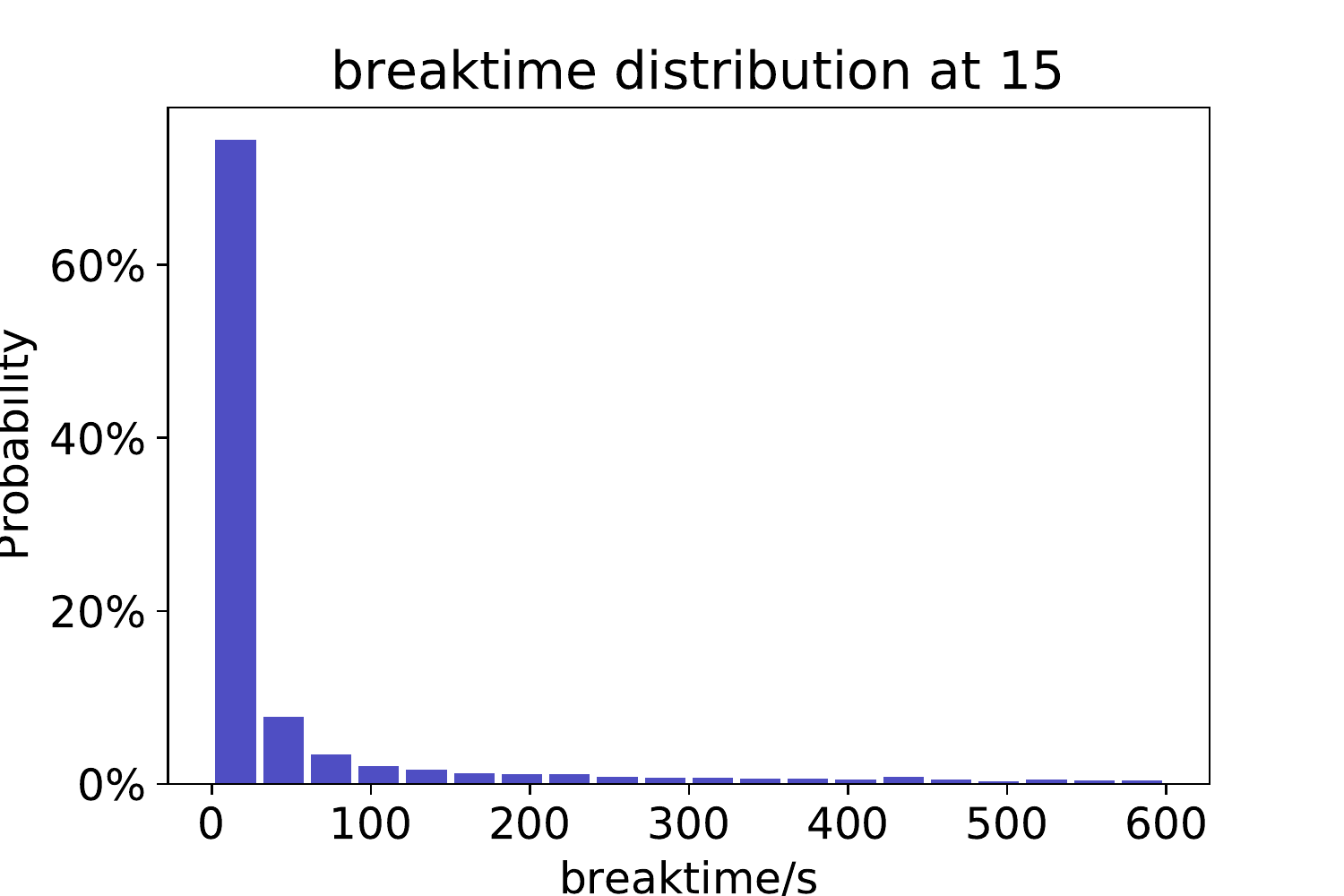}
	\includegraphics[width=0.3\textwidth]{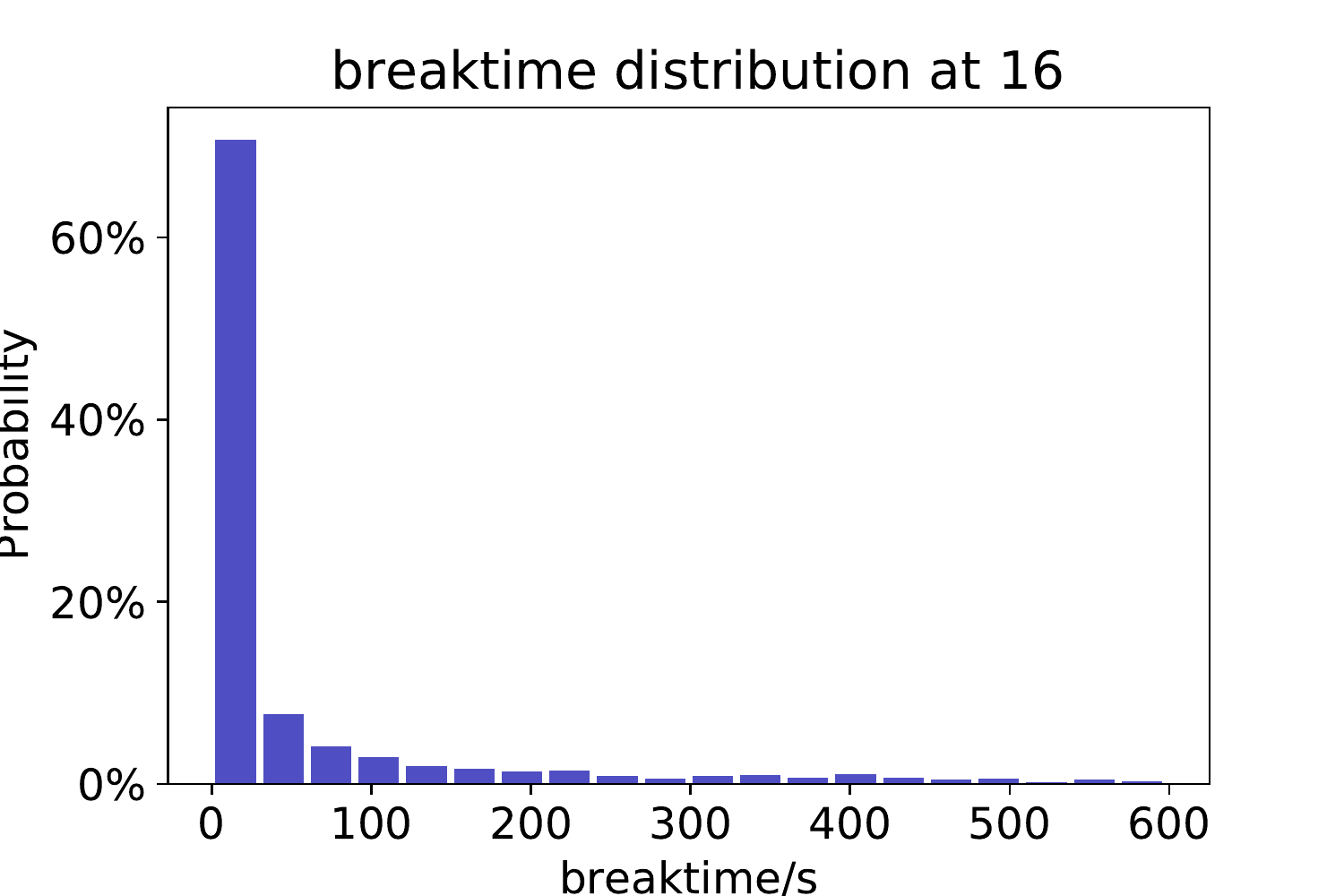}
	\includegraphics[width=0.3\textwidth]{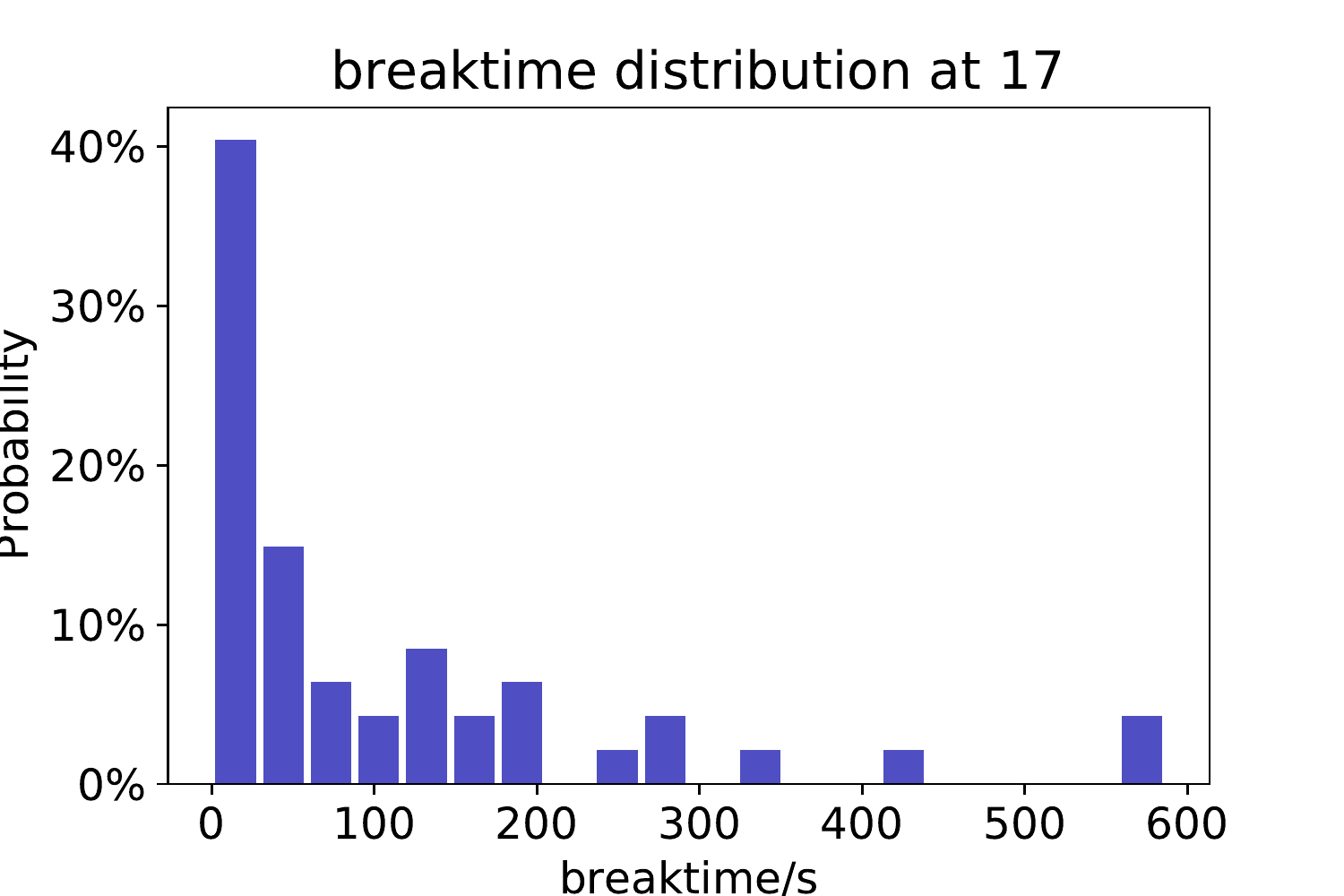}
	\caption{The break time distribution by each hour.}
\end{figure}
\newpage
\item Walking Time Analysis: The walking time at different types of rooms is shown as:

\begin{figure}[htp]
	\centering
	\includegraphics[width=0.4\textwidth]{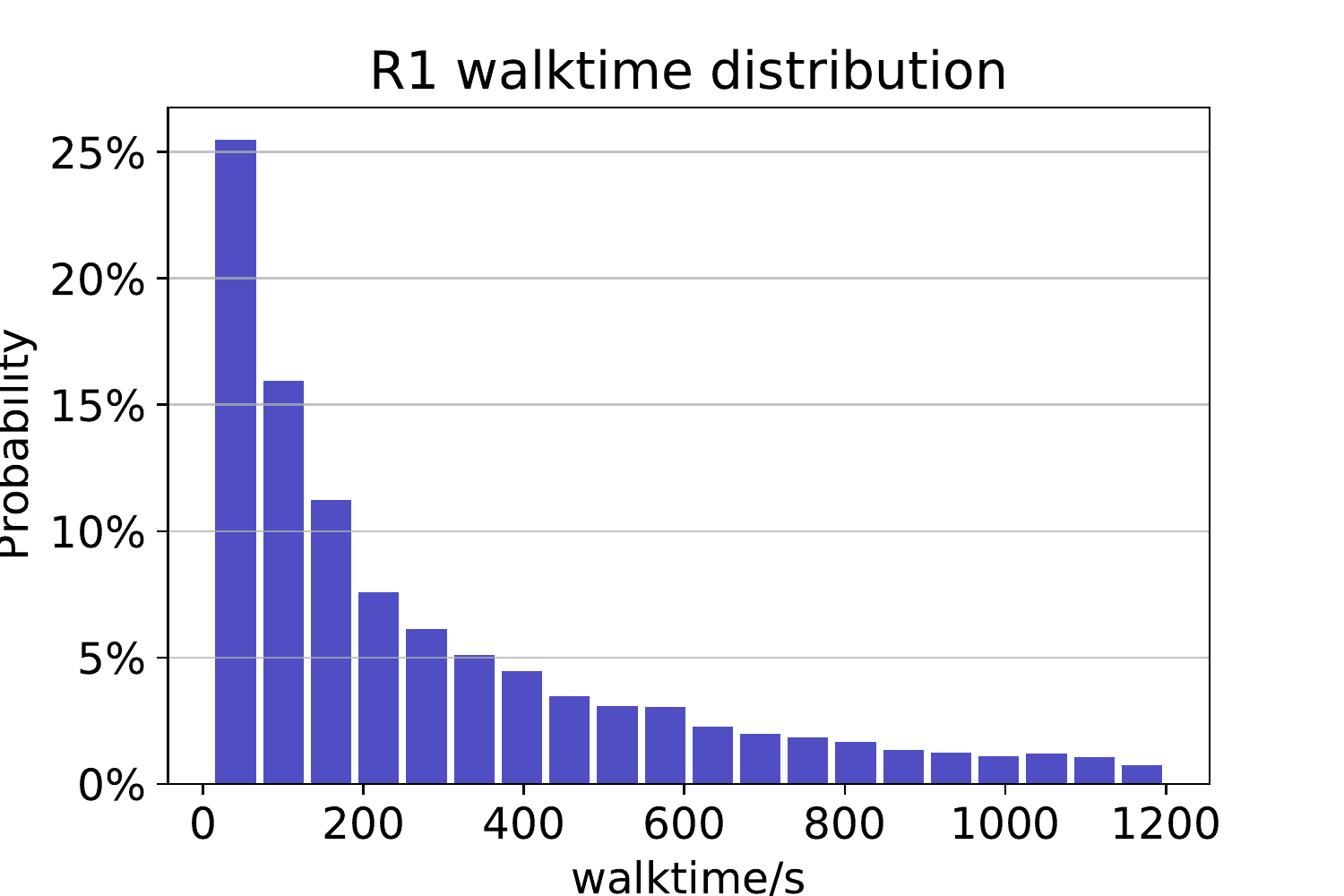}
	\includegraphics[width=0.4\textwidth]{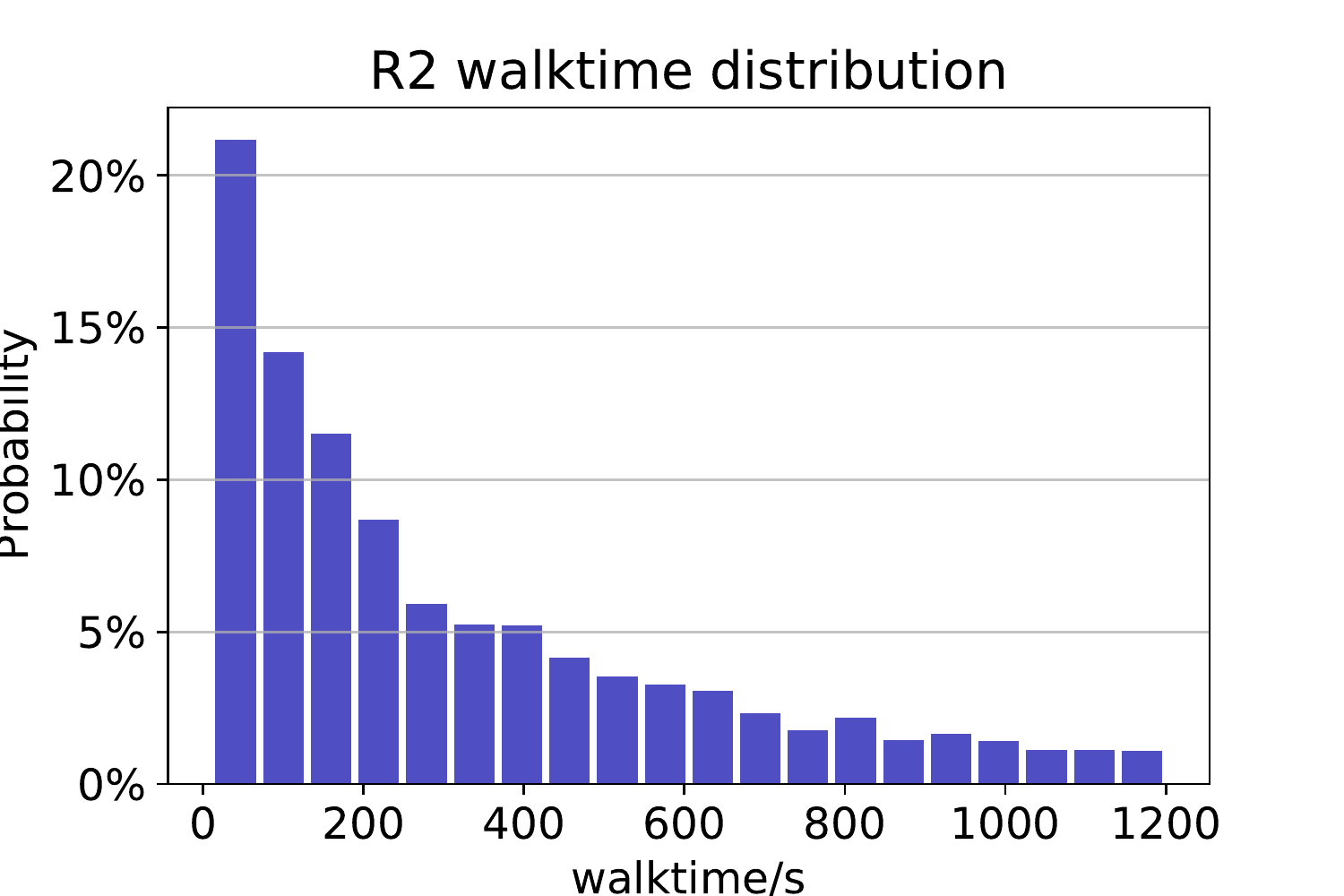}
	\includegraphics[width=0.4\textwidth]{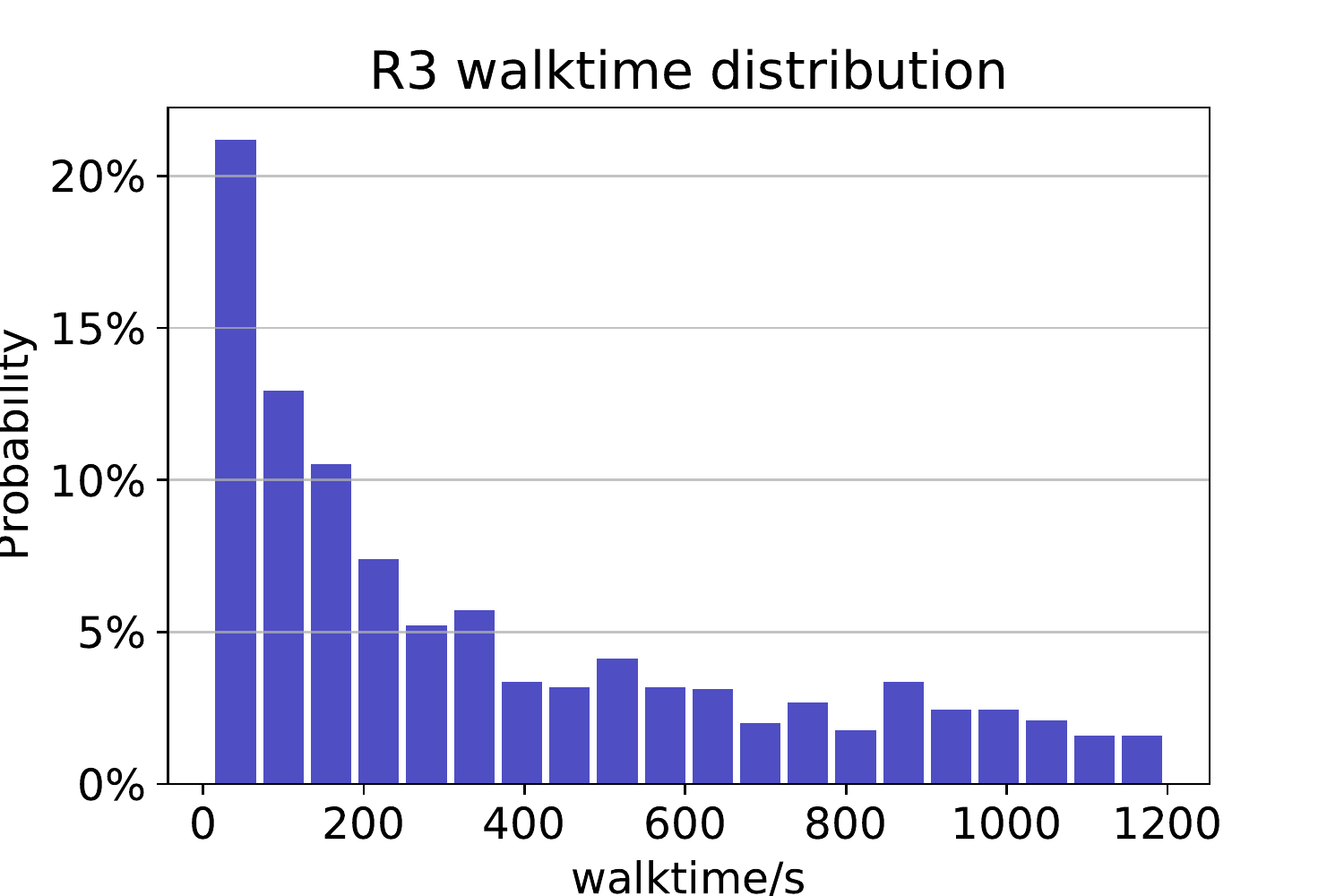}
	\includegraphics[width=0.4\textwidth]{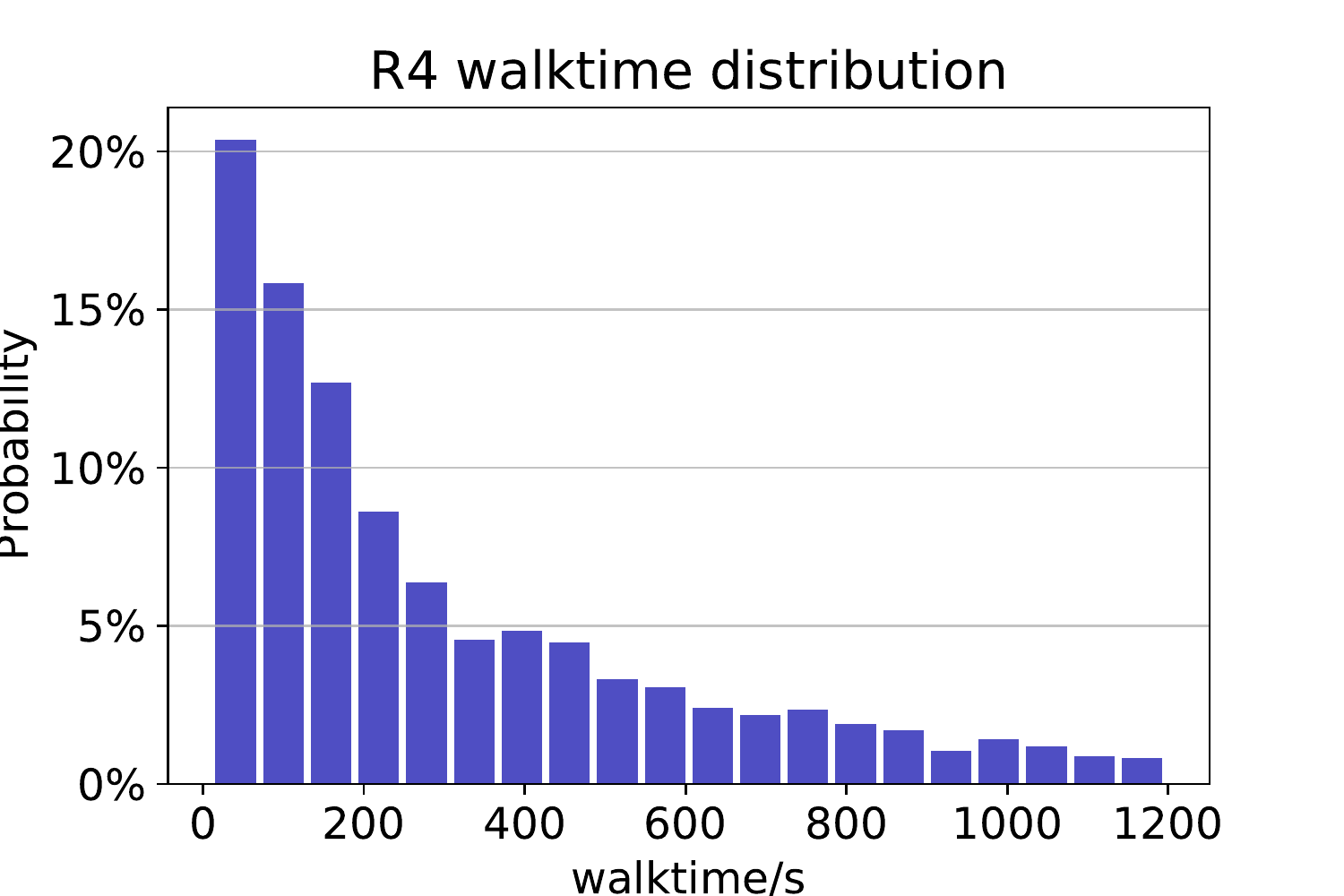}
	\caption{The walk time distribution by each type of room. If the arrival time of $n+1_{th}$ patient later than the end time of $n_{th}$ patient and the start time of $n+1_{th}$ patient minus the arrival time of $n+1_{th}$ patient larger than $10s$, then we deem a walk happens and the walk time equals to the start time of $n+1_{th}$ patient minus the arrival time of $n+1_{th}$ patient.}
\end{figure}

\newpage
\item Walk Time Analysis: The walk time at different hours is shown as:

\begin{figure}[htp]
	\centering
	\includegraphics[width=0.3\textwidth]{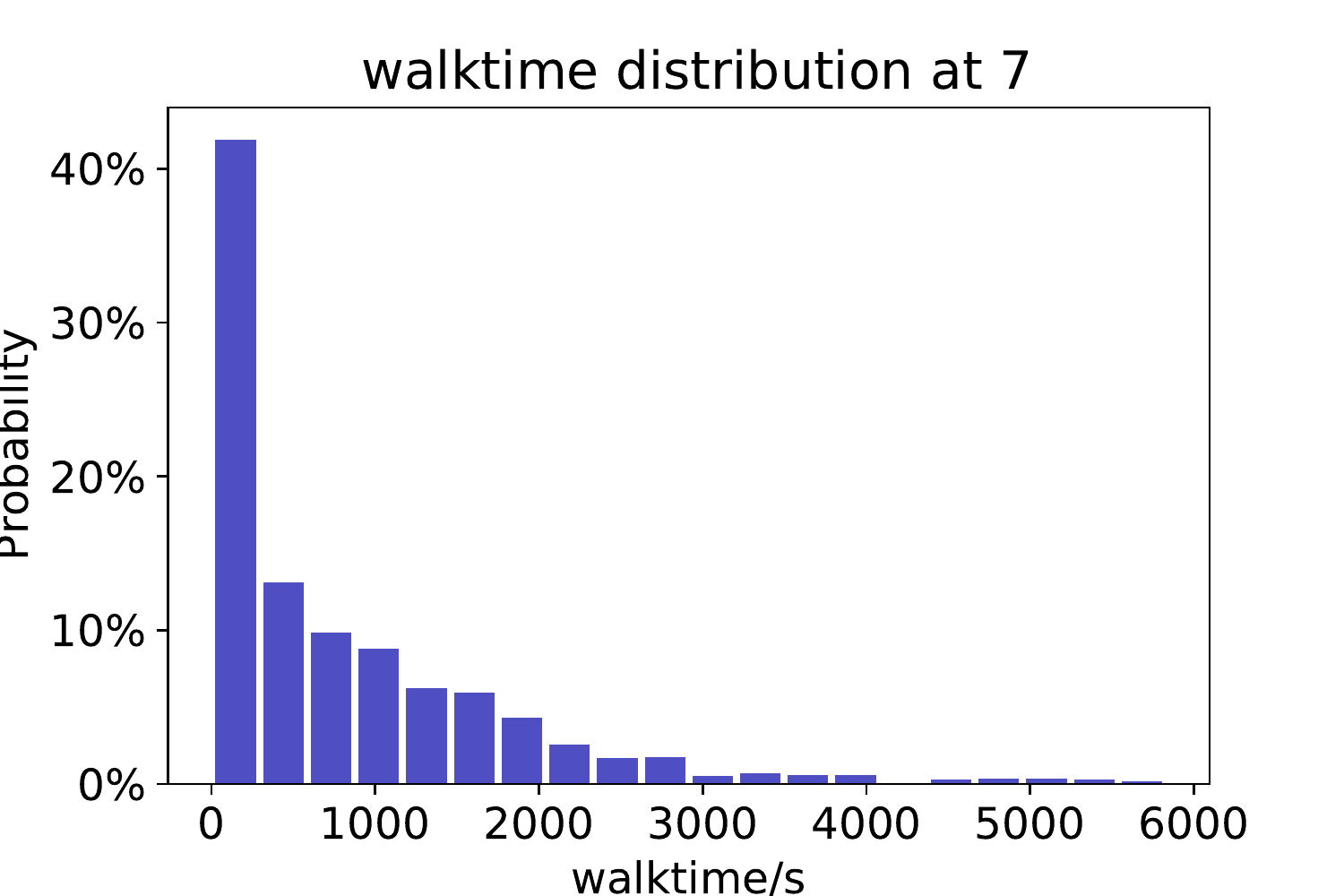}
	\includegraphics[width=0.3\textwidth]{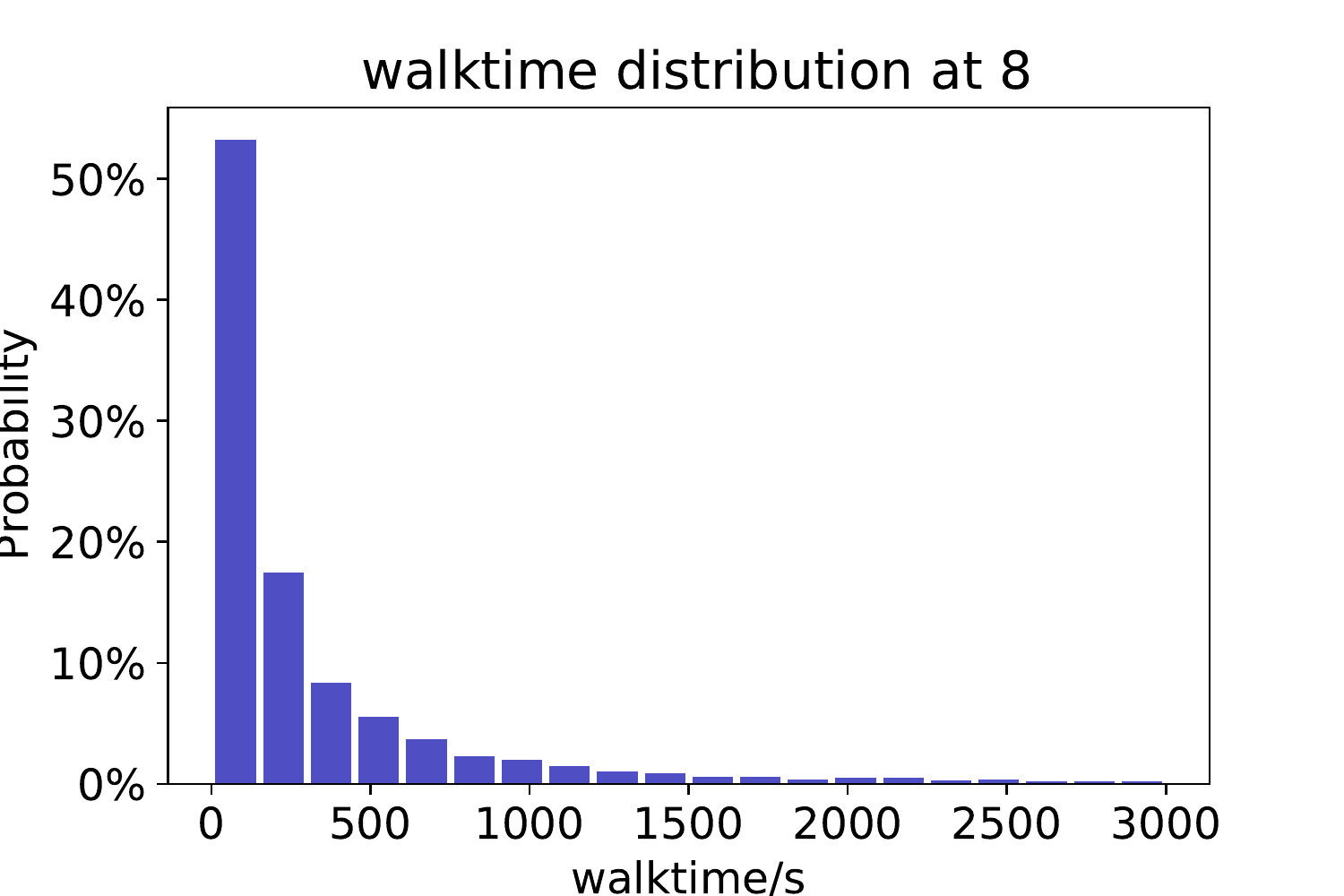}
	\includegraphics[width=0.3\textwidth]{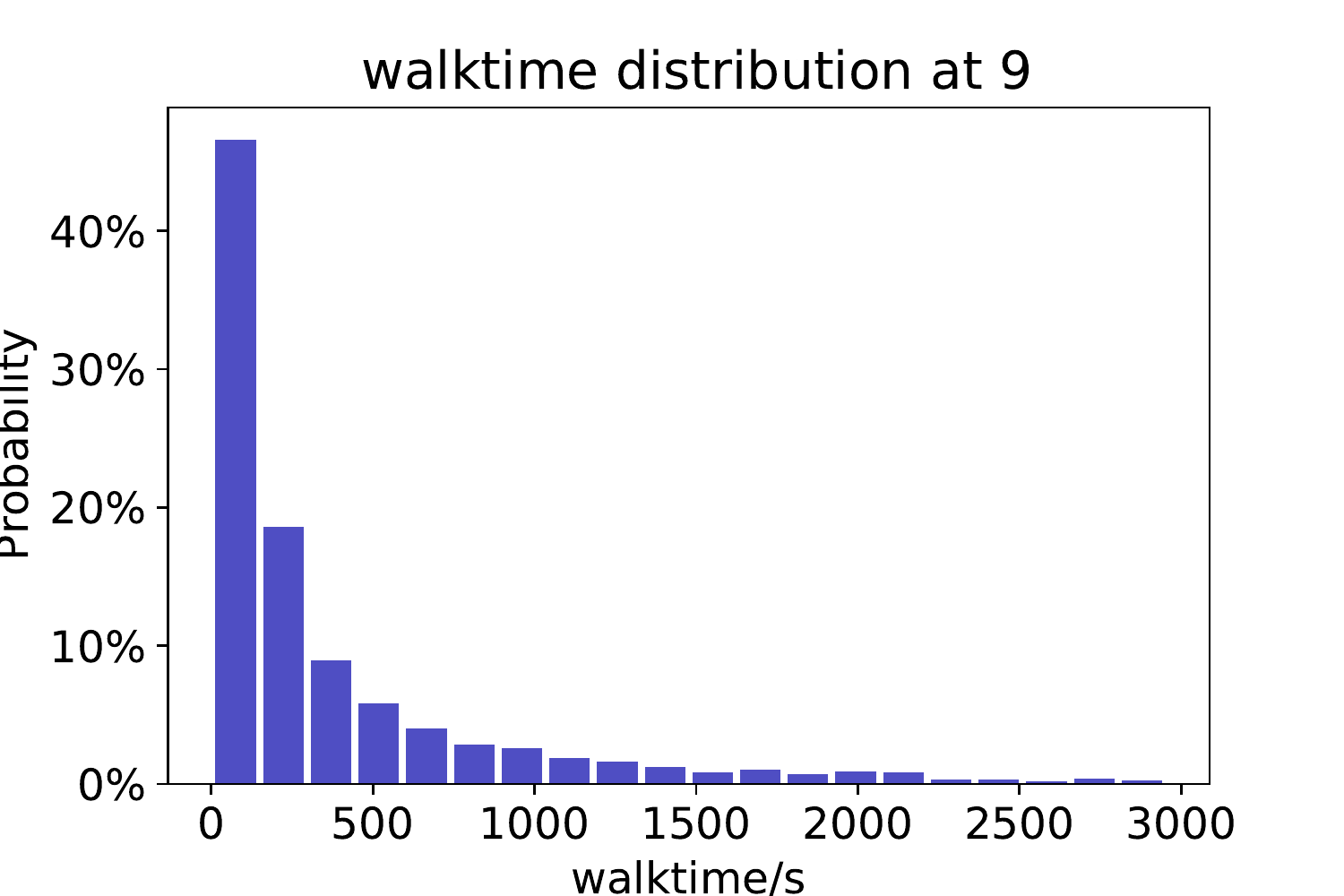}
	\includegraphics[width=0.3\textwidth]{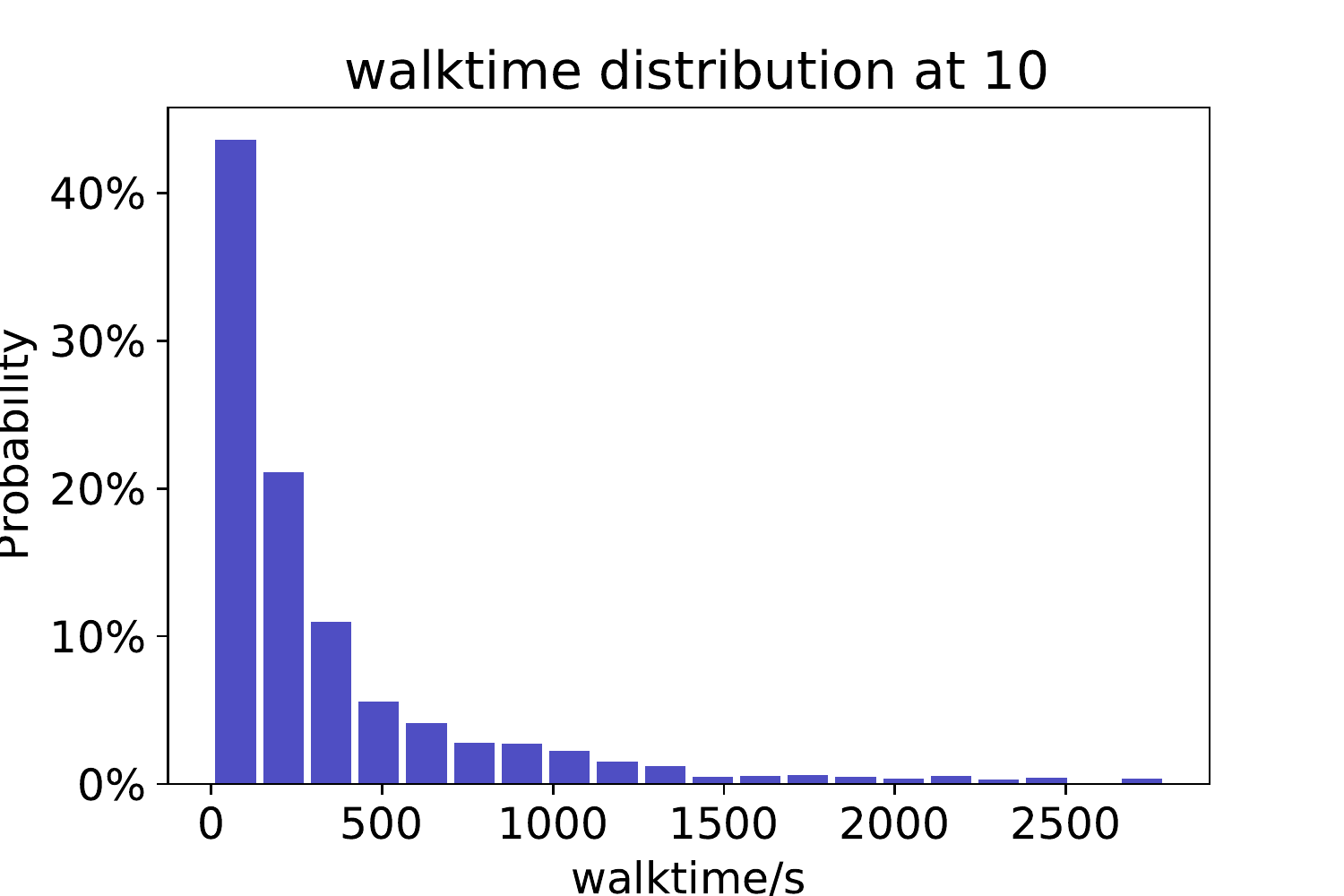}	\includegraphics[width=0.3\textwidth]{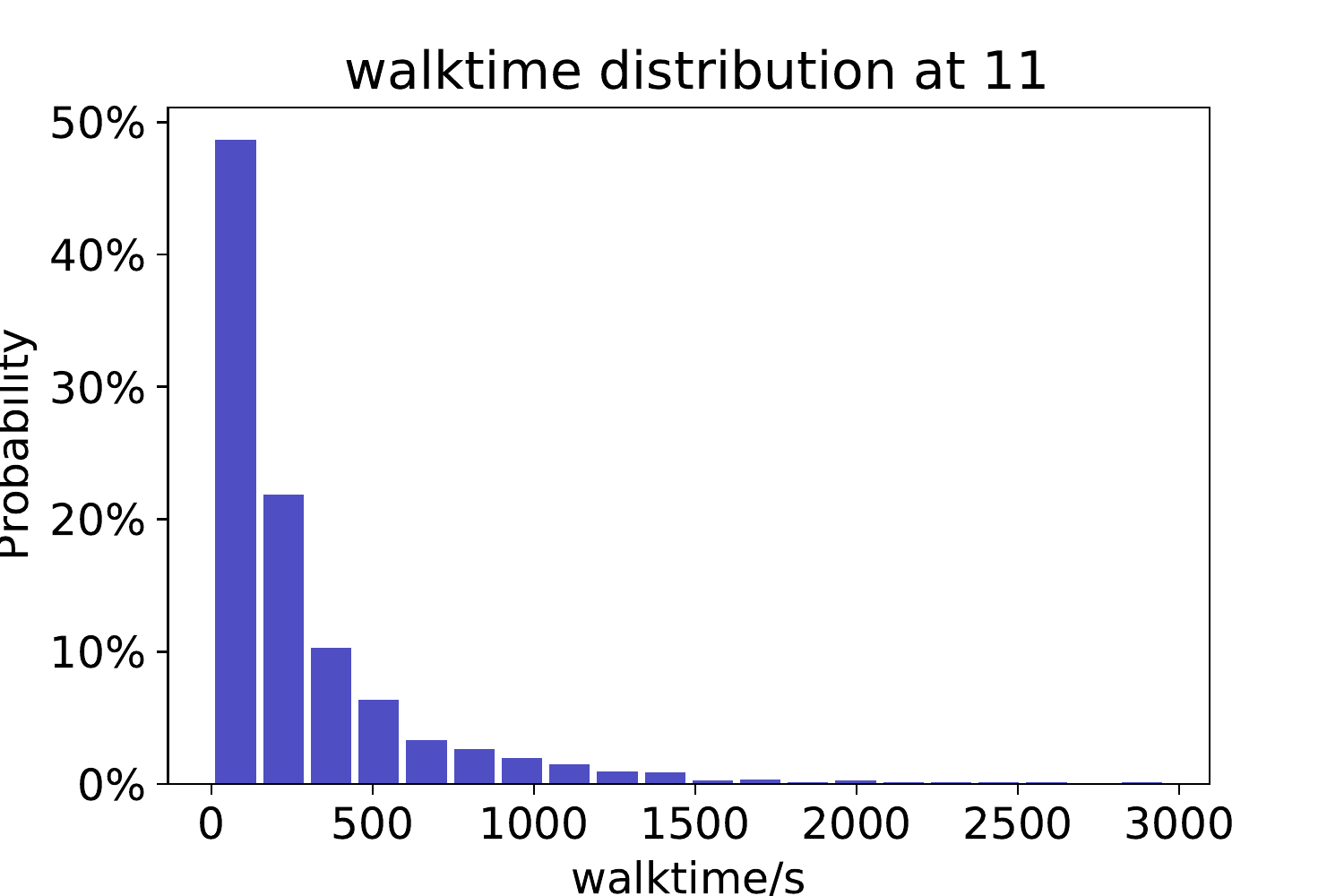}
	\includegraphics[width=0.3\textwidth]{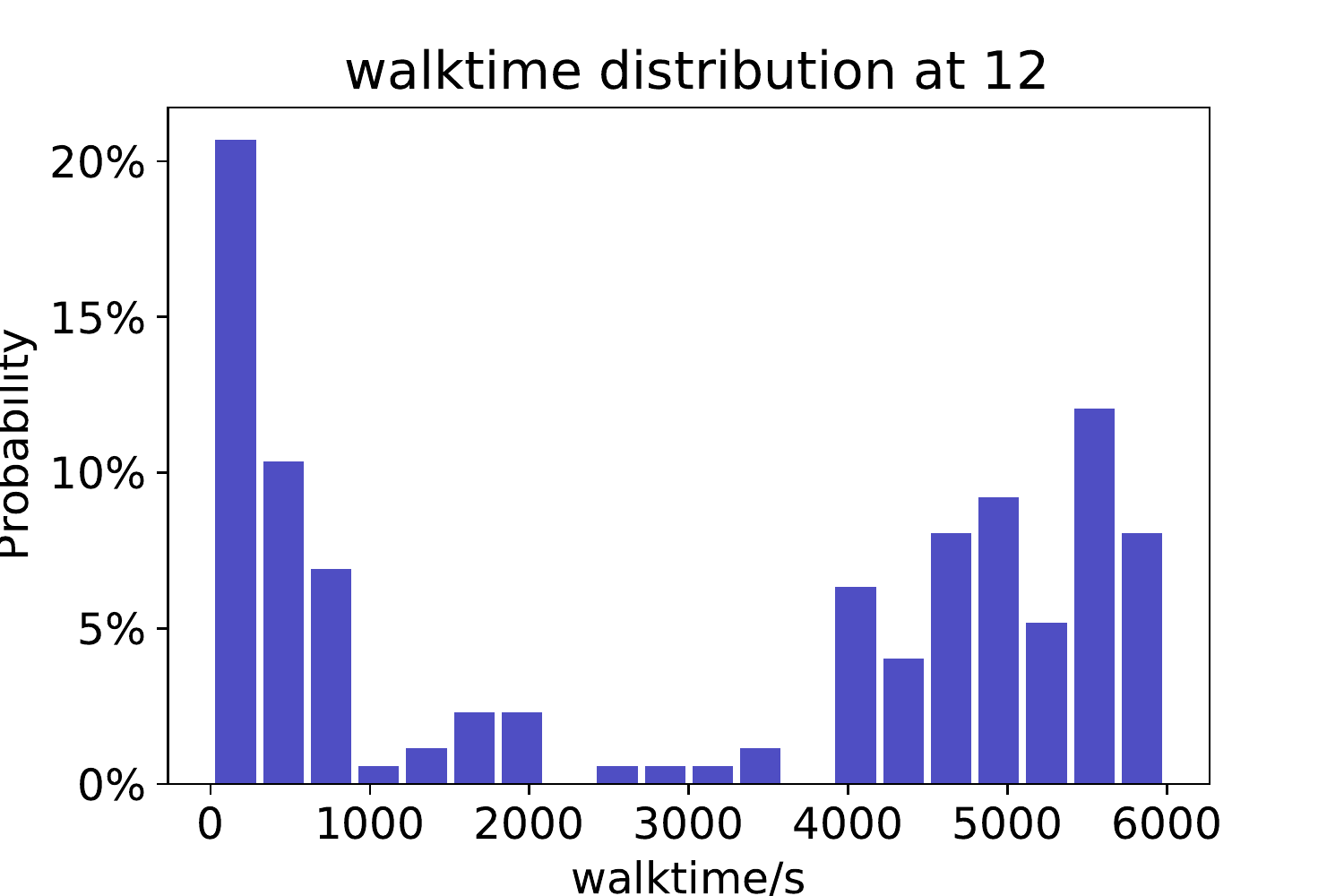}
	\includegraphics[width=0.3\textwidth]{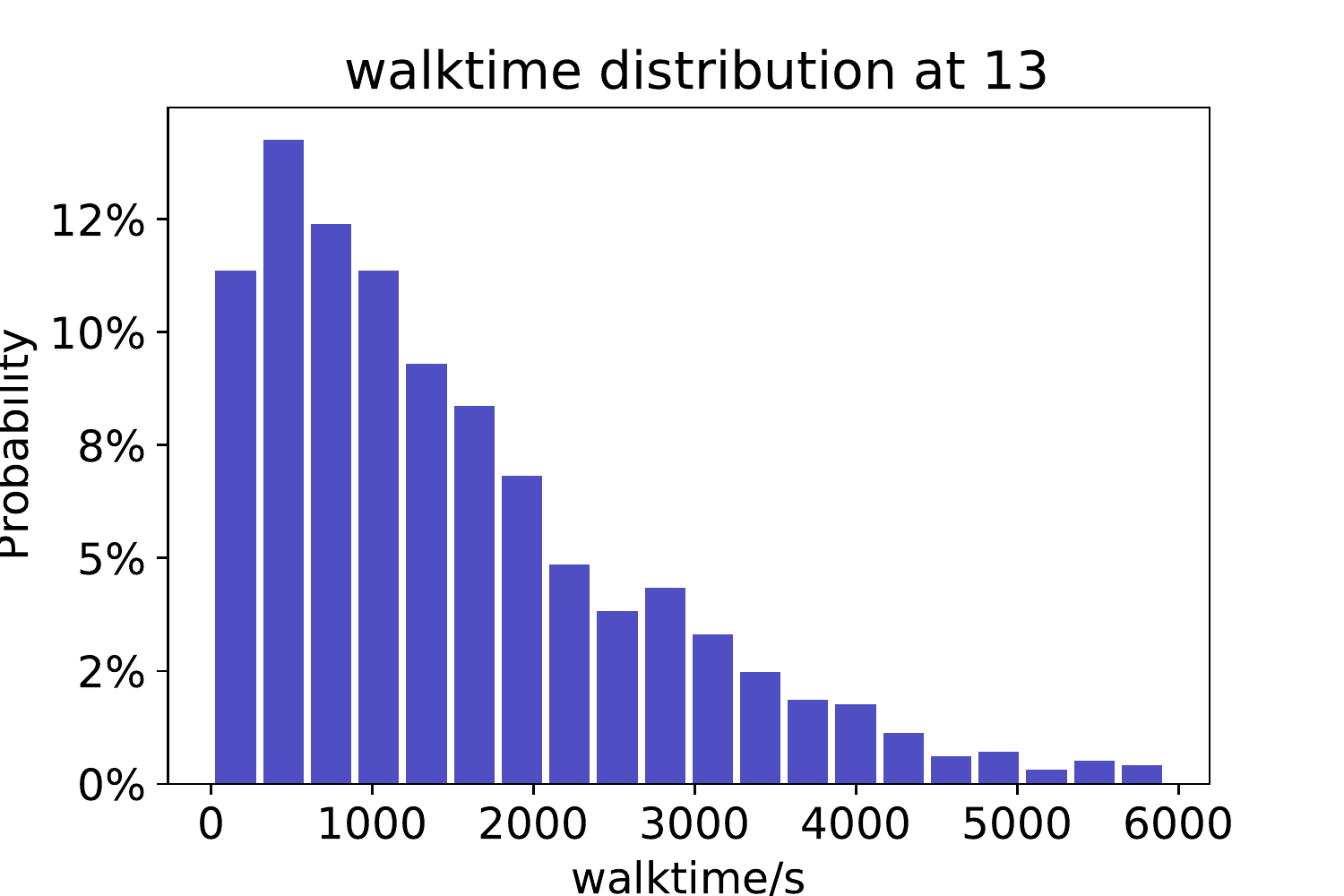}
	\includegraphics[width=0.3\textwidth]{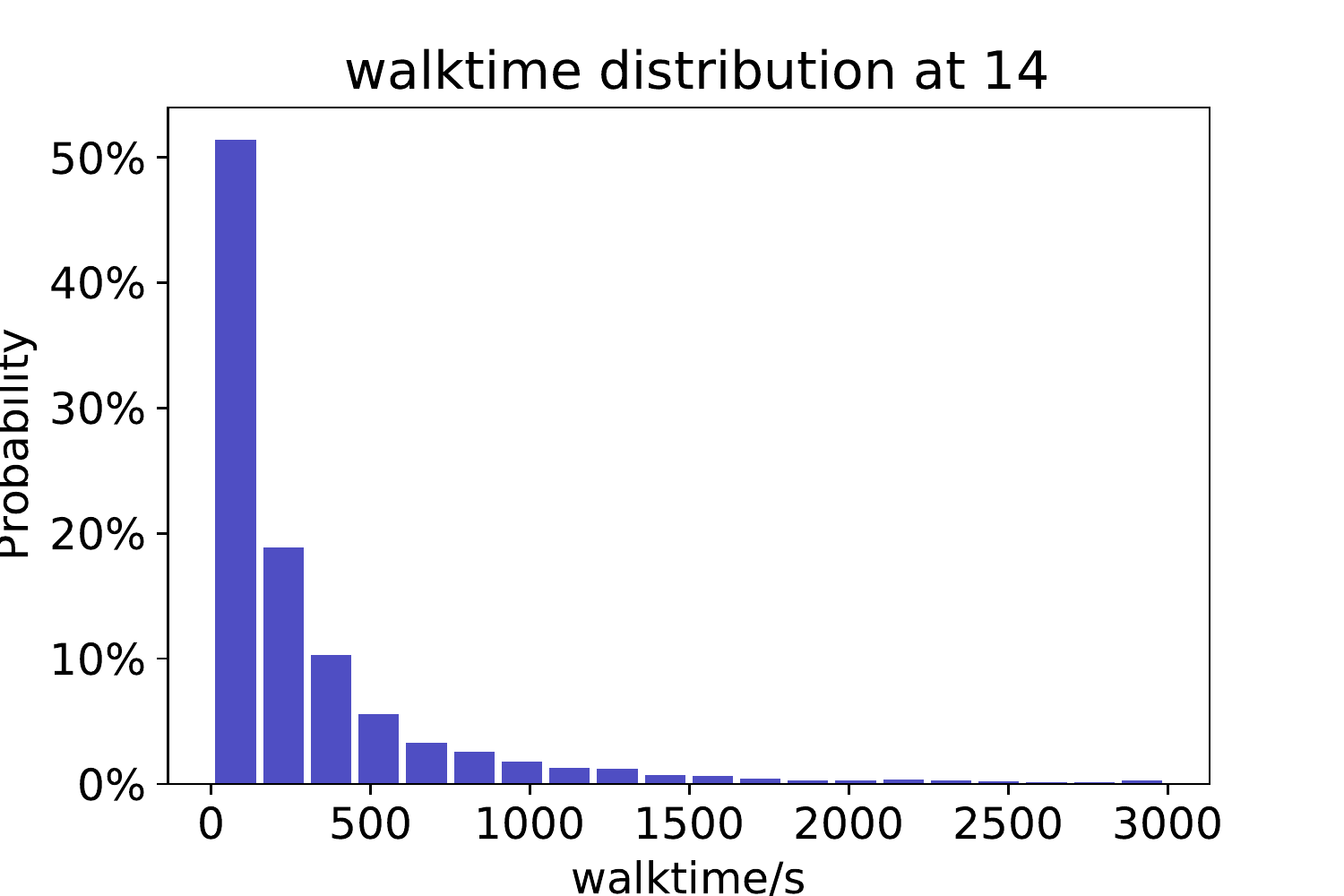}
	\includegraphics[width=0.3\textwidth]{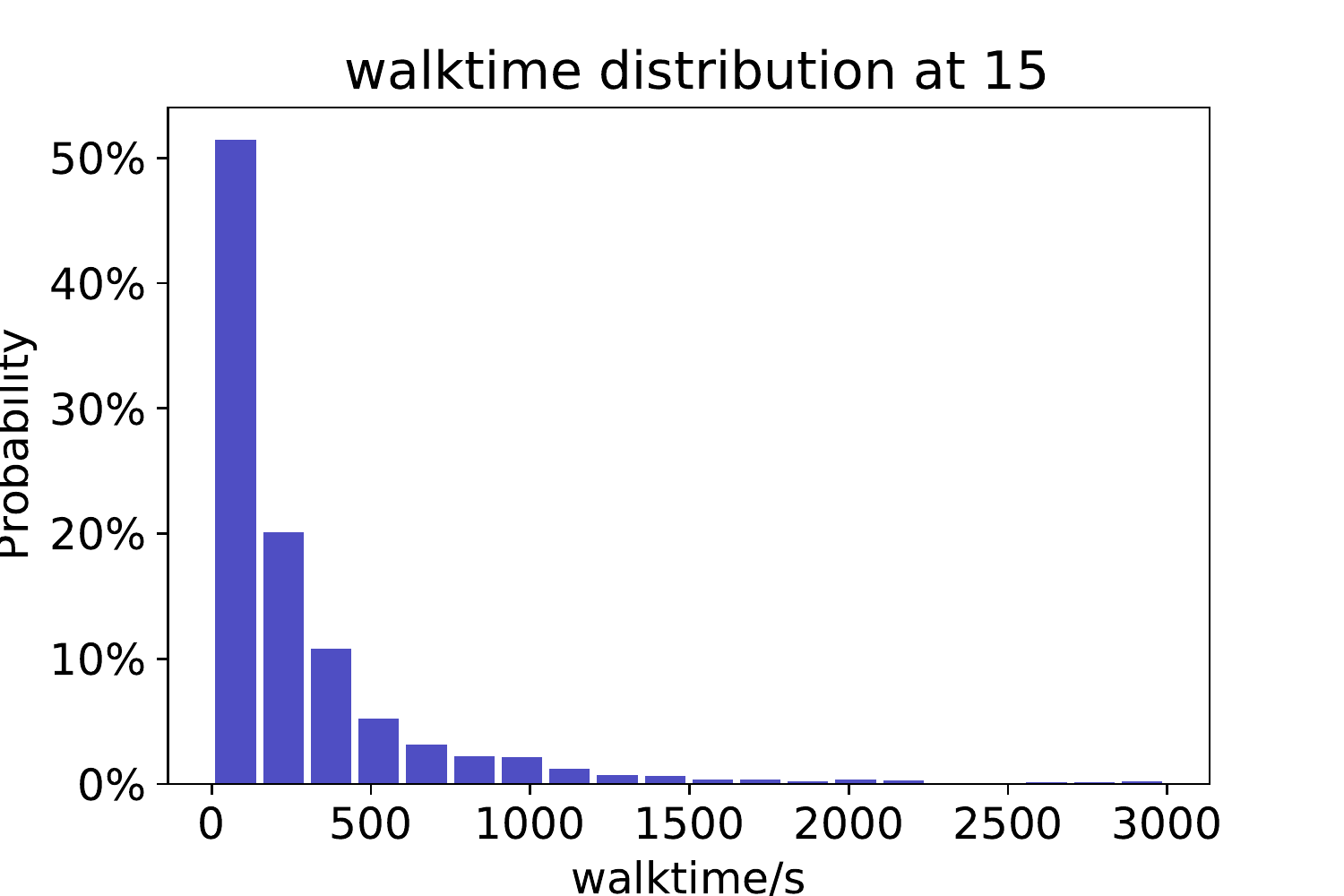}
	\includegraphics[width=0.3\textwidth]{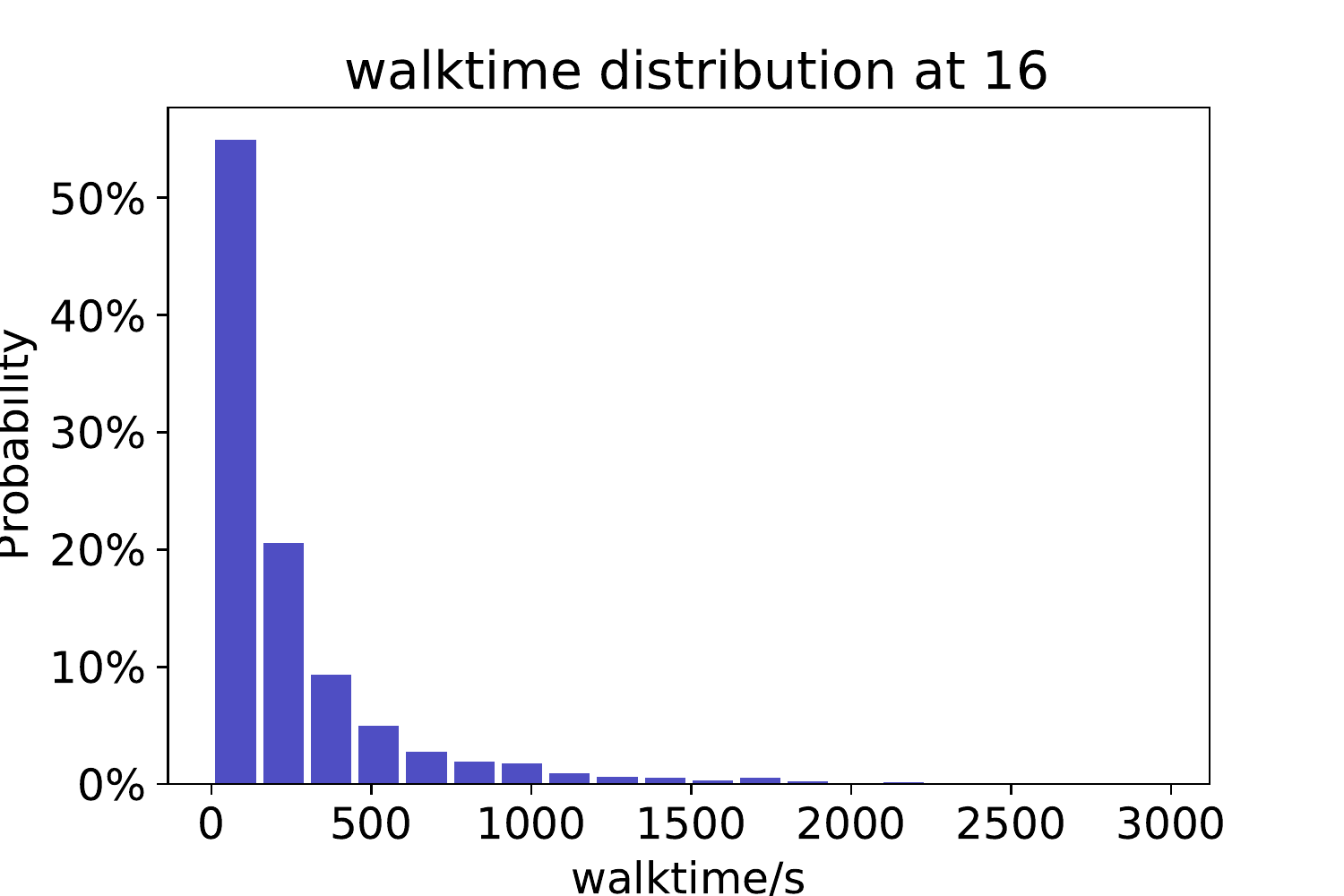}
	\includegraphics[width=0.3\textwidth]{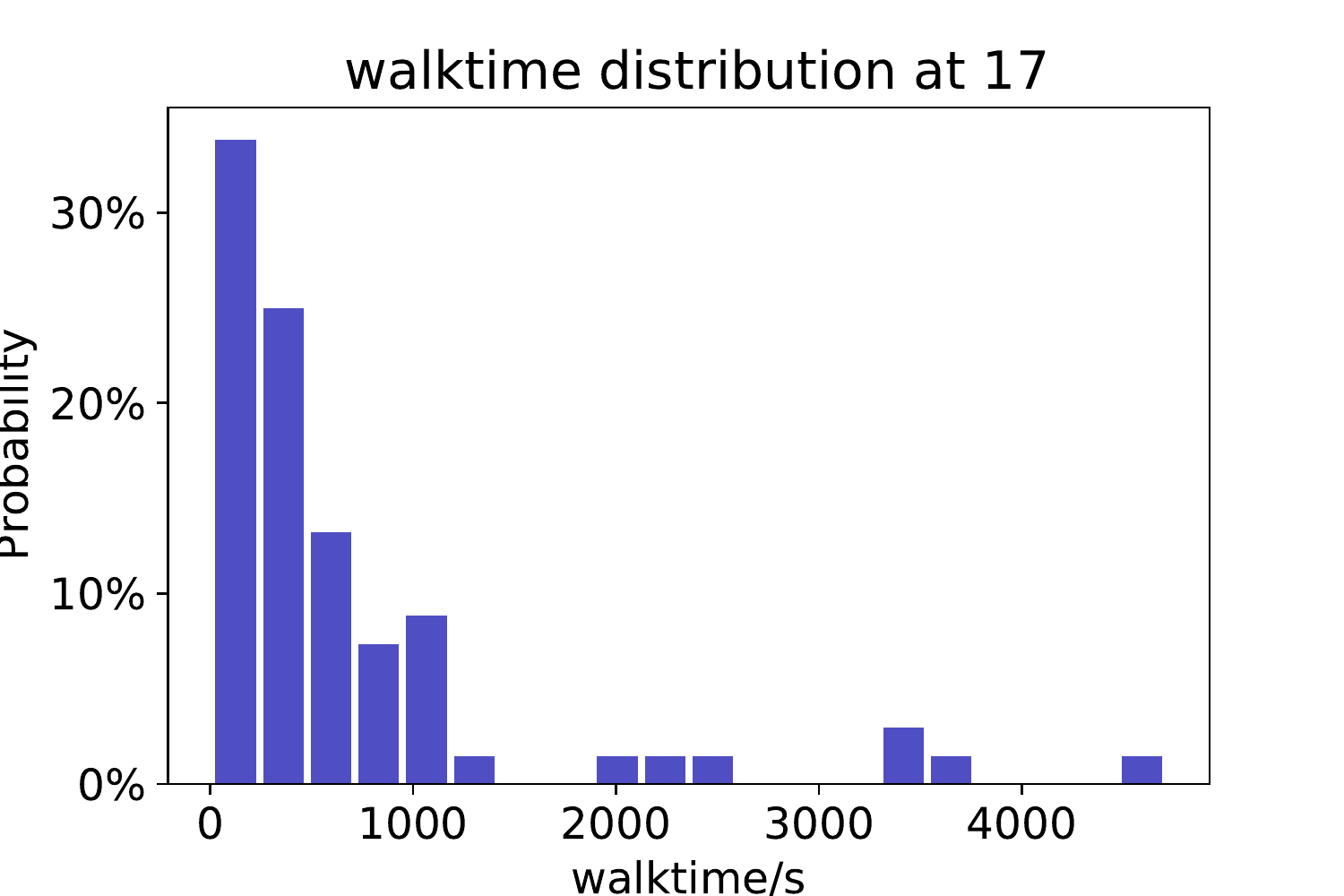}
	\caption{The walk time distribution by each hour.}
\end{figure}

\end{itemize}

\end{document}